# Some Problems of Deployment and Navigation of Civilian Aerial Drones

**Xiaohui Li**

School of Electrical Engineering and Telecommunications
University of New South Wales

# Contents

# List of Figures

# List of Tables

# Chapter 1

# Introduction

## 1.1 Civilian Drone Applications

In the past decades, Unmanned Aerial Vehicles (UAVs), often referred to as UASs (Unmanned Aerial Systems), flying robots, or simply drones, have received much attention from both the industrial and scientific community due to their potential transformative effect for a large number of application scenarios. Drones are initially developed for military applications. Thanks to the development of mechanics, materials science, electronic technology, automatic control and wireless communication, the key components of drones are rapidly miniaturized with lower cost and higher reliability, which plays a decisive role in promoting the birth of civilian drones. Recent years have been a golden period for the development of civilian drones. Moreover, drones are expected to usher in new opportunities in the coming wave of 5G and artificial intelligence technology. The latest figures show civilian drone demand will increase sharply over the next few years, reaching 43.1 billion US dollars global markets by 2024 [1].

Fixed-wing (FW) and rotary wing (RW) platforms are the two main types of drones. Unlike FW drones that need to continually move forward to stay in the air, RW drones have higher maneuverability in terms of vertical take-off and landing (VTOL) capabilities, steady hovering and slow cruising. With such superior maneuverability and relatively lower cost, RW drones have become increasingly popular among both amateur and professional societies for numerous applications. This report studies RF drone systems (or drones, for short) for some novel application scenarios such as animal herding and shark repelling, with a particular focus on the deployment and navigation of drones to achieve desired functions and improve system performance. Currently, the widespread implementation of drones is limited by battery life and regulatory frameworks. Nevertheless, with the development of drone technologies, the application scenarios of drones continue to expand.

A drone is generally composed of aircraft platform system, payload system and ground control system. For different load capacities and missions, one aircraft platform can carry multiple sets of payload systems to achieve complex functions. The success of drones can be explained in part by their great flexibility to carry different devices and sensors as payload. Specifically, drones can carry both sensing and interacting payloads (such as cameras and end-effectors) to collect information from the environment and interact with it. Initially, civilian drones were used primarily for data collection and image transmission (i.e. passive tasks). A new trend is emerging from using drones to interact physically with the environment (i.e. active tasks). In this work, we categorize civilian drone applications into two types, i.e., drone-enabled aerial sensing and drone-enabled aerial interacting. Figure 1.1 shows such a categorization and the corresponding examples of the applications of civilian drones.

## 1.1.1 Drone-enabled Aerial Sensing

Drones have been broadly employed in various aerial sensing applications. Being equipped with sensing devices such as cameras, LiDAR, multispectral, meteorological and chemical sensors, drones can collect diverse information from the physical world and measure many distinct physical quantities such as humidity, temperature or air pollution [2]. Typical sensing tasks include surveillance [3], monitoring [4], remote sensing [5], inspection [6] and so on. For example, drones are popular tools for the surveillance of static [7] and mobile targets [8]. Another application is to monitor some fast-changing environment such as bushfires [9] or road traffic [10,11]. The two main reasons for the popularity of drones in aerial sensing applications are improved mobility and reduced manufacturing costs due to the high-performance control of drones [12].

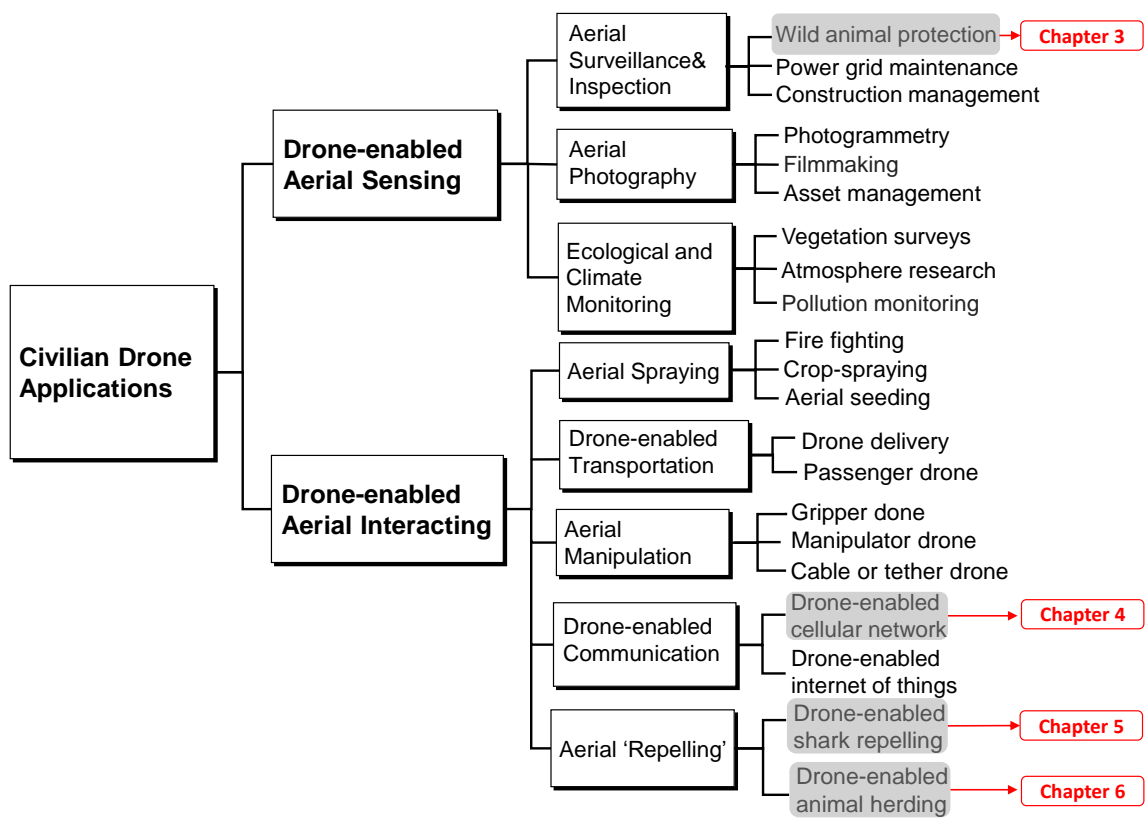

Figure 1.1: Civilian drone applications categorization and the positions of our contributions.

## 1.1.2   Drone-enabled Aerial Interacting

Recently, more is expected from drone systems. In particular, there is a growing interest in a drone to interact with the physical environment. An interacting task may consist of acting upon, exerting power or influencing on the physical world in a variety of ways, depending on the type of payloads [13]. For example, some interacting tasks include object manipulation [14], such as grasping and release [15], delivery [16, 17]. These tasks often involve physical contact between the drone and the targets. On the other hand, some drone-enabled tasks with indirect contact can also be considered as aerial interacting. One example is the drone-cell, which requires the drone to act as an aerial base station to assist wireless communication networks in a variety of scenarios [18]. Specifically, drone-cells are actively interacting with the users through wireless network, and the quality of service (QoS) from the user end is directly influenced by the drone-cells' performance in real-time. Other examples of aerial interacting with indirect contact can be seen in papers [19, 20], where drones are employed as animal repeller through electrical field and approaching motions.

## 1.2 Research Questions

For many drone applications, the efficient deployment and motion control (navigation) of the drones are critical issues. The main topic of this report is to make use of the mobility and fast deployment to improve drone systems' performance and functionality. In detail, we study the following questions:

1. Drones provide a relatively low-cost and risk-free way to quickly and systematically observe natural phenomena at high spatio-temporal resolution [21]. For these reasons, using drones have recently become a major trend in wildlife management and research [2]. However, a number of studies [22–24] have revealed that drones could cause significant disturbance to certain species, and many animals react negatively to drones' presence in biological field research. This phenomenon is being recognized by an increasing number of researchers. And many of them are calling for the development of the drone system that makes less interference to wildlife. Our first research question is how to navigate drone to induce less disturbance to the wild animals.

2. Using drone-cells is a promising solution to improve the area capacity and network coverage of cellular networks by moving supply towards demand when required [25]. However, the deployment of such drone-cells is facing some restrictions that have to be considered. One of the limitations is the availability of reliable wireless backhaul links [26, 27]. Considering the fact that the deployment of the drone-cells is limited by the physical distances/ranges of the available wireless backhaul links, our second research question is the optimized deployment/placement of multiple drone-cells with limited backhaul communication ranges, aiming at maximizing the number of served users. The considered problem is an NP-hard problem.

3. Shark attack is one of the major issues threatening beach visitors such as swimmers and surfers [28]. Despite its rarity, many people are still worried about being attacked by sharks after occasional serial attacks. Currently, the primary method for preventing shark attacks is placing shark nets near the beach area, which leads to the serious bycatch problem of endangered species like turtles and the death of a great number of sharks. In recent years, drones fitted with artificial intelligence (AI) algorithms have been employed as 'shark detector' in some beach areas. But spotting shark earlier is not the final solution for reducing the number of shark attacks. Our third research question is how to use drones to repel sharks and drive them away to protect beach visitors.

4. Animal herding as the vital step of livestock farming has long been the least automated. Dogs that have been used for centuries are still the dominant tools of animal herding. Study shows that herding dogs can only understand and execute 50% of human instructions after years-long training. Besides, herding dogs are suffering from some common issues such as overwork and poor housing conditions, and they cannot get rid of biological limitations, i.e., aging and illness. The applications of robots to animal herding started from the Robot Sheepdog Project in the 1990s [29,30]. But the existing robotic herding methods are mainly designed to deal with small group of animals (e.g., tens) [31–34], while a modern livestock farm usually has tens of thousands of cattle or sheep. Our last research question is how to use a group of drones to efficiently herd a large number of farm animals.

## 1.3 Contributions

Having the above research questions in mind, we conduct extensive research from various aspects, and with a particular focus on the deployment and navigation of the drones. The main contributions of this report are summarized in this section.

- Inspired by motion camouflage, the first contribution is the proposal of one of the first navigation methods for a drone to closely observe a group of animals with reduced visual disturbance. Unlike existing motion camouflage navigation approaches that deceive a single target, we introduce a sliding mode based method that reactively navigates the drone to induce less optical flow on multiple targets' visual system. Specifically, we design a metric to quantify the visual disturbance caused by the drone to a group of moving animals. With this metric, we formulate an optimization problem to minimize the maximum visual disturbance. We also propose the navigation method that guides the drone to minimize the proposed metric while conducting a close-up observation task. In detail, the proposed navigation laws can navigate the drone to approach the animals from an original location, perform a close observation, and fly back to the original location afterwards. We conduct extensive computer simulations to show the performance of the proposed method (see Chapter 3).

- The second contribution of this report lies in the study of backhaul-aware deployment problems of drone-cells. Specifically, we propose a computationally efficient genetic algorithm (GA) based method to solve the NP-hard optimal deployment problem of multiple drone-cells with limited backhaul communication ranges. In particular, GA is a popular method to cope with the complexity of NP-hard problems. But tests show it could easily trap in local optima for the considered deployment problem. To resolve this issue, we present a restart-strategy to enhance the searching efficiency and avoid local optima of GA. For comparison, we also introduce an exhaustive search algorithm that can find the quasi-optimal backhaul-aware deployment of the drone-cells. Simulations show that the proposed GA-based method can save the computing time up to 99.927% compared with the exhaustive search algorithm, and the restart-strategy helps the probability of finding the global optimum by the proposed GA-based method increased from 12% to 92% (see Chapter 4).

- We further introduce a novel shark defence system named as 'drone shark shield system', which uses communicating autonomous drones to intervene in shark attacks for protecting beach visitors. The third contribution is that we not only present the detailed design and the working mechanism of the drone shark shield system, but also propose an efficient interception algorithm that navigates the drone to predicted intersection points to deter the shark. A shark repelling strategy that can eventually drive the shark to leave the beach area by multiple interceptions is also introduced. In addition to protecting beach visitors, the proposed system can also save the life of a number of marine creatures from current shark defence methods. To the best of our knowledge, the proposed system is the first intelligent and non-lethal system that can proactively prevent shark attacks. The effectiveness of the proposed method is proved by computer simulation (see Chapter 5).

- The final contribution of this report is the proposal of a novel automated animal herding system based on a network of autonomous barking drones. The objective of such a system is to replace traditional herding methods (e.g., using dogs) so that a large number (e.g., thousands) of farm animals such as sheep can be quickly collected from a sparse status and then driven to a designated location. We present the detailed design, working mechanism and motion control algorithms of the system. Particularly, we develop a computationally efficient sliding mode based algorithm, which navigates the drones to track the moving boundary of the animal herd and drive the animals to the herd center with barks. The developed algorithm also enables the drones to avoid collisions with others by a dynamic allocation of the steering points. Simulations with an experimentally verified animal behavior model show the proposed system can efficiently herd a thousand animals (see Chapter 6). The proposed system has the potential to be one of the first practical automated herding solutions for a large number of farm animals.

## 1.4 Organization

The organization of the rest of this report is briefly outlined: Chapter 2 reviews the related work on the deployment and navigation of drone-enabled aerial sensing and interacting applications. Chapter 3 studies the autonomous navigation of an aerial drone to observe a group of wild animals with reduced visual disturbance. Chapter 4 presents the efficient optimal backhaul-aware deployment of multiple drone-cells based on genetic algorithm. Chapter 5 introduces a novel method for protecting swimmers and surfers from shark attacks using communicating autonomous drones, i.e. the 'drone shark shield system'. Chapter 6 studies the autonomous navigation of a network of barking drones for herding a large number of farm animals. Finally, in Chapter 7, we summarize the key results and highlight the main future research directions for the presented problems and solutions.

# Chapter 2

# Literature Review

## 2.1 Overview

A considerable amount of literature has been published on the deployment, navigation and control of drones for sensing and interacting applications. This chapter only presents a survey of work related to our studied problems, i.e., the deployment and navigation of drone for surveillance, drone for monitoring wildlife, drone-cell, and drone for repelling animals. Note, there are other hot topics on drone interacting applications like drone manipulation and drone delivery. Limiting by the breadth of this report, we refer readers to [35–42] and the references therein for more comprehensive reviews.

## 2.2 Drone-enabled Aerial Sensing

This section presents a summary of prior works in drone surveillance. We also highlight a brief review on drones for wildlife monitoring.

### 2.2.1 Drone Surveillance

Surveillance is the monitoring of a person, group of people, behaviours, activities, infrastructure, and building to collect, influence, manage, or guide information. Typical surveillance tasks include border patrol, construction management, power grid inspection, traffic monitoring, environmental monitoring, etc.

| The Pain Points of Traditional Artificial Surveillance |
| --- |
| **Low efficiency**: Because of the large scale and scattered environment, it is difficult for traditional manpower surveillance to efficiently locate and reach the concerned locations or fault facilities. |
| **High labour cost**: Traditional surveillance is a labour-intensive industry, many repetitive work scenes require a lot of manpower, and the labour cost is increasing year by year. |

With the development of drone technology, computer vision, and sensor technology, drone systems are becoming increasingly stable and mature to solve these pain points with lower cost, higher security and reliability. Drones can quickly cover large and difficult-to-reach areas, reducing labour costs, and do not require much space for the operators. It has become the best tool to replace human to complete surveillance, monitoring and inspection work efficiently and safely. Figure 2.3 shows some typical application scenarios of drone surveillance.

A large and growing body of literature has investigated drones for surveillance and monitoring, and presented a variety of technologies and methods. Searching for keywords in the Web of Science Core Collection, it can be seen that research related to drone or UAV surveillance has increased rapidly since 2011, as indicated in Figure 2.2.

Some researchers have tried to review the related works from different aspects and subsets, such as Uma *et al.* [43] for crops monitoring drones, Di *et al.* [44] for harmful algae blooms monitoring drones, Balmukund *et al.* [45] for search and rescue

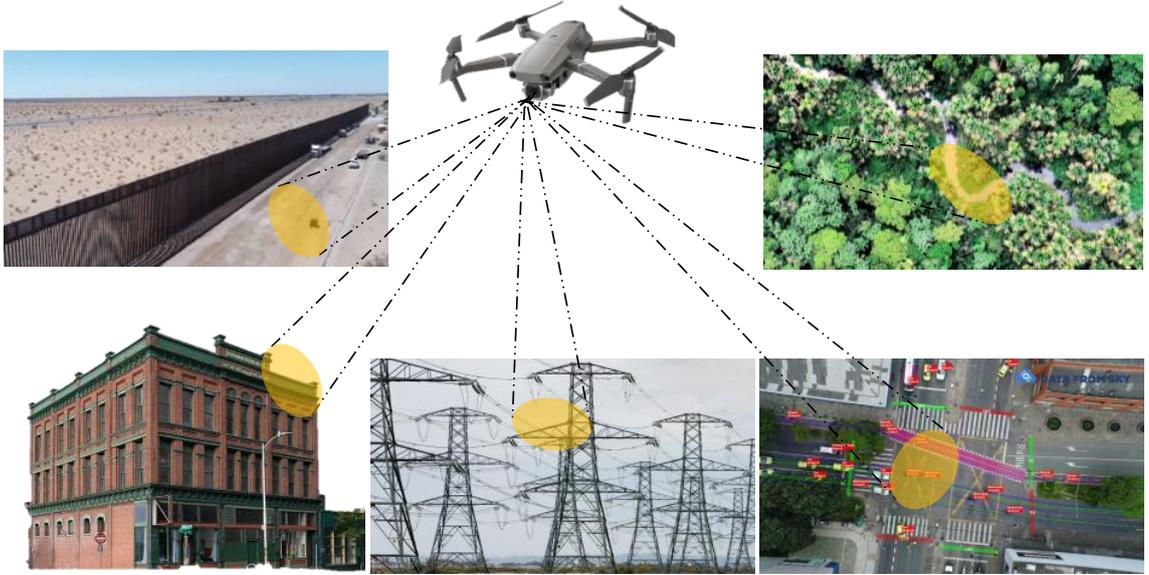

Figure 2.1: Typical application scenarios of drone surveillance, includes border patrol, construction management, power grid inspection, traffic and environmental monitoring.

drones, Francesco *et al.* [46] for railways surveillance drones.

Focusing on the the deployment, navigation and control of drones for surveillance, we select some representative approaches and classify them by research questions, number of drones, operating modes and target types, see Figure 2.3. Specifically, previous studies on the deployment of drones for surveillance include: [3,4,7,8], and [47]. References [6,9–12,48–50] investigated the optimized navigation of drones for surveillance of ground targets. From the perspective of targets types, the surveillance of static or stationary targets was studied in [3,4,7,47], whereas [6,9–12,48–50] discussed the scenario with mobile targets. The operating modes of the surveillance drones can be divided into two categories: proactive (also known as 'offline', see [4,7,47]) and reactive (i.e., 'online', see [6,8–10,12,48–50]). Obviously, the reactive approaches are more suitable for dynamic environment, and more research effort should be made on them. In addition, references [48,49] investigated the cases with single drone, and references [3,4,6,7,9,10,12,47,50] studied the surveillance with a group of drones.

Using only one drone for a specific mission may be risky because the drone may encounter technical or other problems. Generally, various tasks can be performed more efficiently by deploying a team/network/swarm of drones because they can collect more temporal-spatial data than using a single drone. Moreover, for a reconfigurable or robust network of drones, if one or part of the drones is lost in flight, the rest of the drones can still carry out the mission. Moreover, in group flight, a combination of various types of drones with different sizes and configurations can be used for a formation flight to conduct complex tasks. In fact, drone swarm has become one of the most important topics on drones' research [51].

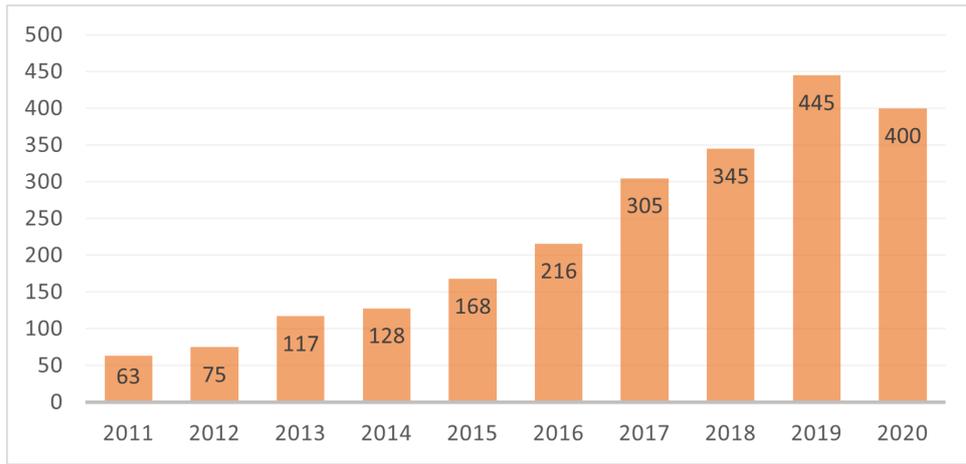

Figure 2.2: Number of Publications on Drone or UAV Surveillance From the Web of Science Core Collection.

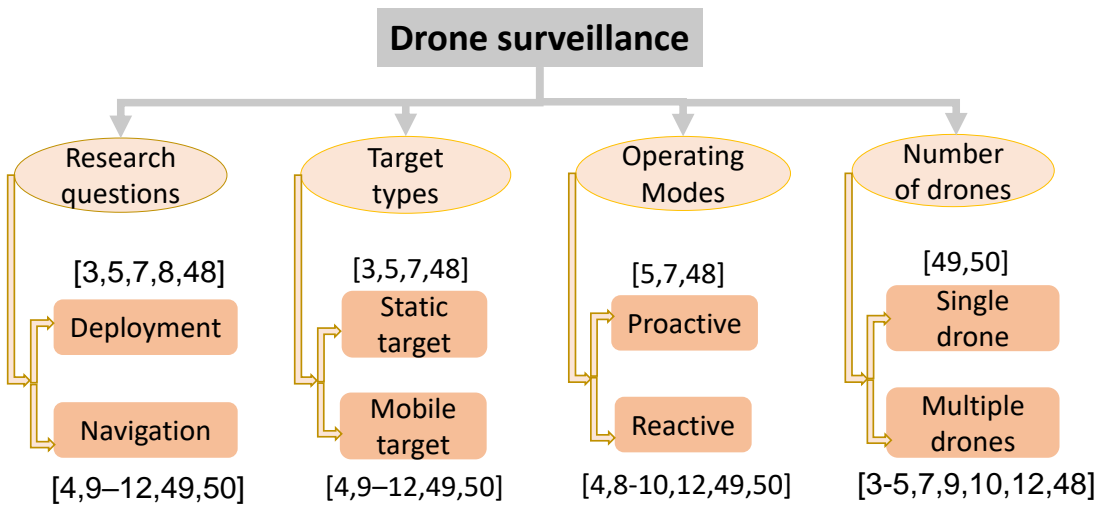

Figure 2.3: Summary and taxonomy of previous studies on drone surveillance.

### 2.2.1.1  Deployment Problem

From the optimization point of view, the literature on the deployment of surveillance drones can be divided into three categories according to the considered optimization objectives:

> **1**: To ***minimize the number of drones*** necessary to monitor a given area, see [7, 47];
>
> **2**: To ***maximize the quality of coverage*** of an area or targets for surveillance and monitoring, see [3, 4, 11];
>
> **3**: ***Joint objectives*** such as balancing drone's energy consumption and the number of covered targets, see [8, 48].

References [7] and [47] studied the problem of minimizing the number of drones for surveillance tasks. In detail, paper [47] investigated the problem of deploying a group of drones for surveillance and monitoring of a ground region, with the goal of minimizing the number of drones to observe every point of the region with ground-facing cameras. To solve this problem, the authors of [47] developed an easily implementable algorithm based on Kershner's theorem from combinatorial geometry. The results are proved to be asymptotically optimal in the sense that the number of drones deployed is close to the minimum number of drones for large ground regions. Different from [47], the paper [7] considered the scenario of monitoring a very uneven terrain. Specifically, drones are to be deployed over a very uneven terrain area with the goal to cover every point of the area. The authors in [7] viewed the problem as a drone version of the 3D Art Gallery Problem. A computationally simple algorithm was proposed to calculate an upper estimate of the minimal number of drones necessary. The proposed algorithm also gives the locations of the drones. The limitation of [7] is that the considered terrain model are simple polyhedrons, while real uneven terrains can be much more complicated with many irregular structures.

Reference [4] focusing on deploying a network of drones to surveillance and monitoring a set of static ground targets. To characterized the quality of coverage of targets by surveillance drones, authors in [4] proposed a novel coverage model and further presented a reactive collision-free three-dimensional deployment algorithm to maximize the overall quality of coverage of targets by a group of surveillance drones. In particular, the proposed algorithm consisted of two control laws for horizontal submovement and vertical submovement of the drones in real-time. The authors analyzed the computational complexity and proved the convergence of the algorithm. The letter [3] considered the deployment problem for a group of drones to maximize the quality of coverage of an area for surveillance. The authors proposed a distributed optimization model and coverage maximizing algorithm to find the locations of the drones. The proposed optimization model considered the constraint that a connected communication graph needs to be maintained between the drones and some ground nodes. The proposed algorithm does not require global information and can converge to a local maximum thin a finite number of steps. The authors conducted simulations with a real dataset to demonstrate the effectiveness of their method. For high-quality surveillance of groups of moving pedestrians or vehicles on given paths with unknown speeds, [11] proposed a computationally simple algorithm to determine the deployment of multiple drones. The concerned quality of coverage is calculated based on the distances between the drones and targets (the closer, the better). The proposed algorithm only requires local information with minimal involvement of the central station. The authors proved the local optimality of the proposed algorithm.

For the joint objectives considering both energy consumption and the number of covered targets of surveillance drones, [8] designed a control system containing a movement decision-maker and proposed a decentralized algorithm to reactively determine a fleet of drones' positions in 3D space that contributes more to the coverage. Simulations indicated that the proposed method achieves better network lifetime and target coverage. [8] only considered the targets on the 2D ground and did not consider possible blockage caused by some high-rise buildings.

### 2.2.1.2 Navigation Problem

The navigation of mobile robots (include drones) involves a number of hot research topics such as motion control [52, 53], path planning [12, 54, 55], trajectory tracking [56], collision avoidance [57–60], fault estimation [61] and so on. We now introduce some representative work on the navigation of the surveillance drone.

The problem of navigating surveillance drones to periodically monitor a set of moving targets is studied in [6, 12]. Specifically, to periodically monitor a group of moving ground targets on the 3D terrain, [12] proposed a reactive sliding mode control algorithm to navigate a team of communicating surveillance drones with ground-facing video cameras. Particularly, [12] adopted a Voronoi partitioning technique to decrease the movement range of the drones and reduce the revisit times of the targets. [6] studied the navigation of a group of solar-powered drones for periodical monitoring a set of scattered mobile ground targets in urban environments. In the considered scenario, the number of targets is larger than that of the drones, so that the drones need to carry out a periodical surveillance. In addition, the authors considered the existence of tall buildings in urban environments, which may block the Line-of-Sight (LoS) between a drone and a target. The tall buildings may also create some shadow region, so that the solar-powered drone may not be able to harvest energy from the sun, and the surveillance may become invalid. In [6], such a periodical surveillance problem is formulated as an optimization problem to minimize the target revisit time while considering the impact of the tall buildings. To solve this problem, the authors proposed an autonomous navigation algorithm based on rapidly exploring random tree (RRT) to guide the movements of the drones in real-time. To further narrow drones' moving space and reduce the target revisit time, a partitioning scheme is also adopted to group targets. The limitation of [12] and [6] is that the considered targets are all assumed to move on some given trajectories. When these trajectories are unavailable, the accuracy of the target position predictions may significantly decrease and the drones may lose some targets. To solve this problem, some searching operations can be adopted during the surveillance mission.

Different from the widely studied target tracking, [48] studied the problem of navigating a drone to carry out covert video surveillance to a single mobile target. In detail, [48] proposed an online trajectory planning method with a balanced consideration of the energy efficiency, covertness, and maneuverability of the surveillance drone. The authors in [48] first designed a new metric to quantify the covertness of the drone. Specifically, the drone disguises its intention by changing the relative drone-target angle and distance as drastically and frequently as possible. Then, they formulated a multiobjective trajectory planning problem to maximize the disguising performance and minimize the trajectory length of the drone, and presented a forward dynamic programming method to the problem. A similar study can be seen in [49], in which the authors proposed a bioinspired bearing only navigation law for a video surveillance drone to covertly monitor a moving target. In particular, the proposed method is based on sliding mode and inspired by motion camouflage stealth behavior observed in some attacking animals. It can navigate the drone to monitor a moving target while concealing its motion with respect to the target's visual system. Moreover, the proposed navigation law is based on bearing only measurements, i.e., directions from the drone's current position to the moving ground target. It does not require any information on the targets' velocity and the distance to the target.

With the increase in urban population and the rapid increase in the number of private vehicles, many roads have become more congested than ever. Road traffic monitoring plays a crucial role in traffic management. Currently traffic monitoring mainly rely on static road-side units, which passively collect the traffic information. For road traffic monitoring by a drone network, [10] proposed a decentralized autonomous navigation algorithm for the surveillance drone to detect traffic blockage and then effectively gather to the blocked area and monitor the majority of the targets. The proposed algorithm consists of stages: initial, searching, accumulating and monitoring, and the drones only need to share measured information and their positions with their neighbours. A limitation of [10] is that the proposed navigation laws are for a planar motion only, and 3D mobility of the drones wasn't utilized. Another study on road traffic monitoring by drones can be seen in [50], in which the authors proposed a distributed navigation algorithm based on Voronoi partition for a drone network to maximise the quality of surveillance of a group of targets moving along a curvy road with unknown time-varying speeds. The convergence of the drones positions to the local optimal locations was also proved in [50], but the global optimality is not guaranteed.

Another application of surveillance drone is on disaster relief. In the problem of monitoring a moving disaster area by drones equipped with ground-facing cameras, [9] proposed a sliding-mode control algorithm that navigates the drones to monitor the faster moving segment of the disaster area's frontier. The authors in [9] proved that the proposed method tracks the fastest spreading parts of the frontier of the moving disaster area, and converge to the global maximum in the considered optimization problem. The proposed method requires the initial positions of drones to be near the frontier of the moving disaster area.

## 2.2.2 Drone for Wildlife Monitoring

An increasing number of countries and organizations have adopted drones to conduct observation of wildlife in hard-to-reach places. From monitoring sandhill cranes in Colorado and counting waterbirds in Florida, to investigating orangutan dens in Indonesia and seals in Arctic waters, drones are flying at a safe distance, protecting endangered wildlife and enabling environmentalists to work more safely, accurately and economically. A summary of existing publications on drones for wildlife monitoring can be seen in Table 2.1. We now introduce some of the representative works on drones for wildlife monitoring:

For marine biology, 'SnotBot' project [62] uses a modified DJI Inspire 2 drone to conduct whale research. In particular, SnotBot collects blowing samples of whales as they surface and exhale. The collected blow samples contain whale's DNA, microbial communities, tissue particles, stress and pregnancy hormones, and viruses, all of which are important indicators of the whale's health. In addition to being a non-lethal and non-invasive approach to ocean research, protective drones such as SnotBot are also democratizing opportunities for ocean research. Whale research has long been confined to a privileged few because the chartering of expensive marine ships and equipment requires serious financial support. But now, drones in oceanography are making data collection affordable, replicable and scalable for researchers everywhere.

Table 2.1: Summary of Existing Publications on Drone for Wildlife Monitoring

| Applications | Species | References |
|---|---|---|
| Health Monitoring | Whale | [62, 63] |
| | Ungulates | [64] |
| | Forest | [65–68] |
| Population Survey | Feral horse | [69] |
| | Penguin | [70] |
| | White-tailed deer | [71] |
| | Sumatran orangutan | [72] |
| | Sea Lion | [73] |
| | Sea turtles | [74] |
| | Shark | [75] |
| | Koala | [76] |
| Behaviour Research | Sea turtles | [77] |
| | Whale | [78] |
| | Crocodiles | [79] |
| | Salmon | [80] |
| Habitats Investigation | Proboscis monkey | [81] |
| | Raptor | [82] |
| | Waterbirds | [83] |
| Anti-poaching | Rhinoceros | [84] |
| | Elephant | [85, 86] |

In 2019-20, an unprecedented bushfire swept more than 12.6 million hectares of land across Australia, caused the death of more than 61,000 koalas (one of the iconic marsupials in Australia). Sadly, even before the bushfire crisis, koalas were considered vulnerable to extinction due to threats posed by hunting, land development, food degradation, drought and disease. After the bushfires, the urgent problem is to seek surviving koalas in burned and unburned areas. To solve this problem, Australian ecologists demonstrate an infield protocol for wild koala surveil-lance drones, which provides real-time validation of high-resolution thermal signatures of koalas [76]. The authors also provide detectability considerations relative to wildlife–drone interactions, temperature, survey time, and detection of non-target species, which can be used to further inform drone survey protocols. Once a koala is located, a hyperzoom vision camera on the same drone helps first responders determine whether they need medical help or not. Mapping wildlife using thermal drones is much cheaper and more efficient than using traditional survey methods, because koalas have strong camouflage ability and hard to be spotted. Moreover, thermal imaging drones have also become a powerful tool in the areas of fire protection. From detecting invisible hot spots and preventing secondary fires to collecting wildfire data at night, thermal imaging drones have become an important technology to improve the safety of firefighters and civilians.

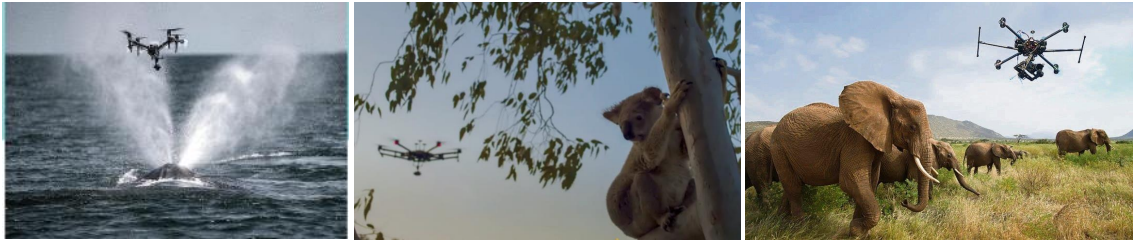

Figure 2.4: Examples of wildlife monitoring drones for protecting whales, koalas and elephants.

Elephants are important ecosystem engineers, helping to maintain the biodiversity of forests and grasslands. However, despite the international ban on the ivory trade, 20000 to 40000 elephants die from poaching each year. The situation is so bad that African elephants, previously classified as "vulnerable" by the International Union for Conservation of Nature (IUCN), were promoted to the "endangered" list last month, because studies have shown that the number of African forest elephants has declined in the past three decades. By more than 86%. Conservation efforts to protect elephants are challenging due to their vast habitat range [85]. To combat elephants poaching problem, anti-poaching drones have been used by conservation organization and researchers [86]. With thermal imaging anti-poaching drones, poachers can be found both during the day and at night. Besides anti-poaching, drones can record and forward real-time videos to ground teams that are usually miles away, and they can even record the videos for later analysis.

Any wildlife monitoring results will be shared with rangers and police. The use of anti-poaching drones has had a profound effect on detecting and stopping poachers [85]. Particularly, authors in [85] proposed methods for identifying spatial distribution patterns of elephant poaching incidents based on point pattern analyses in the Tsavo National Parks area in Kenya. The geospatial analyses on the physical environment were performed to create a risk map based on how land cover, water features, and roads correlate to poaching incidents. The drone flight paths were also modelled based on drones flight characteristics and the horizontal view angle for a selected thermal camera. Authors in [85] found that poaching incidents were geographically clustered and followed a predictive (deterministic) process, and were dominantly close to roads and water features. They conclude that a combination of GIS-based risk analysis and aerial surveillance will enable conservation teams to improve the efficiency of their anti-poaching efforts with limited budgets.

However, the negative impact of wildlife monitoring drones can not be ignored. Many studies show that wildlife monitoring drones can cause significant disturbance to different species of wild animals [22–24, 87]. For this problem, we will introduce one of the world's first navigation methods for a drone to closely observe a group of wild animals with reduced visual disturbance in Chapter 3.

## 2.3 Drone-enabled Aerial Interacting

### 2.3.1 Deployment of Drone-cells

Because of its mobility, fast deployment and corporation, drone-cells can assist wireless communication networks in a variety of scenarios, such as serving users in severe shadow or interference conditions. For drone-cells, a fundamental research problem is its optimal deployment. An increasing number of publications have investigated the optimal deployment of drone-cells for improving network performance [18, 25, 26, 26, 88–96]. The searching results from Web of Science Core Collection show that the research related to the deployment/placement of drone-cell/drone-BS/UAV-BS has increased rapidly since 2015, as indicated in Figure 2.5. We now present a brief review of some representative work on this topic.

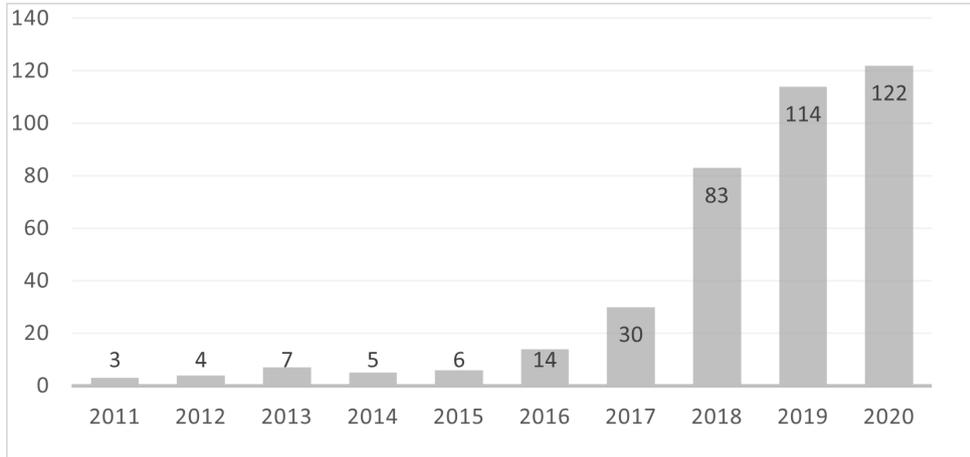

Figure 2.5: Number of Publications on Deployment/Placement of Drone-cell/Drone-BS/UAV-BS From the Web of Science Core Collection.

Existing publications have investigated various aspects regarding drone-cells' deployment for wireless coverage. For example, [97, 98] studied the optimal deployment of drone-cells to achieve energy-efficient wireless coverage. The authors in [97] formulated a problem to minimize the average transmit power of a drone-cell that serves a set of ground users with equal and non-equal transmit power to each user, respectively. Based on the decoupling method, the authors proposed an optimal drone-cell placement algorithm when considering the equal transmit power to each user. For non-equal transmit power case, the authors in [97] further proposed a drone-cell placement algorithm by using the successive convex approximation (SCA) technique. Simulation results verified that the power savings of the proposed algorithms. [98] proposed an optimal placement algorithm for multiple drone-cells that maximizes the number of covered users using the minimum transmit power. To simplifying the drone-cells' deployment problem, authors in [98] decoupled the problem in the vertical and horizontal dimensions. They modelled the drone-cell deployment in the horizontal dimension as a circle placement problem and a smallest enclosing circle problem. Simulations in [98] verified the transmit power savings and increase in the number of covered users as the user heterogeneity increases by applying the proposed method.

To investigate the 3D deployment of a drone-cell for maximizing the number of covered users with different quality-of-service (QoS) Requirements, [98] models the deployment problem as a multiple circles placement problem and proposed an exhaustive search (ES) algorithm over a 1-D parameter in a closed region. In addition to the ES algorithm, [98] also proposed a maximal weighted area (MWA) algorithm to solve the placement problem, which is computationally efficient. To investigate the drone-cell deployment for minimum-delay communications, [99] formulated a minimum-delay drone-cell placement problem, subject to practical constraints imposed on the drone-cell' battery life and velocity. Authors in [99] transformed the primal problem to the corresponding constrained Markov decision process (CMDP), and provided a reinforcement learning aided solution to the problems formulated under various assumptions concerning the wireless teletraffic dynamics. [96] studied the placement optimization of multiple drone-cells, subject to minimizing the number of drone-cells to cover every user in the considered area. To this end, authors in citelyu2016placement proposed a polynomial-time algorithm with successive drone-cells placement, where the drone-cells are placed sequentially starting on the area perimeter of the uncovered users along a spiral path toward the center, until all users are covered.

Providing wireless backhaul for drone-cells is another major challenge that must be considered, but the backhaul limitations of drone-cells' have not been studied in many details. [26] studied the optimal backhaul-aware 3D placement of a drone-cell over an urban area with users having different rate requirements was investigated. In detail, The authors in [26] considered both the wireless backhaul peak rate and the bandwidth of a drone-cell as the limiting factors in both the user-centric and network-centric methods in a typical Heterogeneous network. Specifically, the network-centric method maximizes the total number of served users, regardless of their required rates, while the user-centric method maximizes the sumrate of those users. [26] proposed a backhaul limited optimal drone-cell placement algorithm for various network design parameters, such as the sumrate of the served users and the number of the served users in a clustered user distribution. Simulations conducted by the authors in [26] demonstrates the robustness of the proposed algorithm by showing that only a small percentage of the total served users would experience an outage as they move. The proposed method is a centralized solution by assuming that the global view of the network is available at a central controller, which may not be a practical case.

In Chapter 4, we will introduce the efficient optimal backhaul-aware deployment of multiple drone-cells based on genetic algorithm. Some other related work can also be seen in Chapter 4.

### 2.3.2 Drone for Repelling Animals

Despite the ordinary applications of drones, they can be used in some non-ordinary tasks. As an example, animals can sometimes conflict with the human being, and drone is potentially a tool for repelling animals when necessary, a number of studies [20, 100–104] has proved its effectiveness. Specifically, In protected areas, human-wildlife conflicts in populated areas is a common problem [105]. For example, crop-raiding is one of the most common forms of conflict between people and wild animals, provoking both retaliatory killing of wild animals, and animosity towards wildlife among local communities [106].

Some studies described the use of drones in various management tasks, such as repelling monkeys and elephants away of human settlements or agriculture [100,104]. In detail, [104] introduces a case of using a drone to repel crop-feeding and fruit-raiding monkeys for protecting commercial fruit trees in the village of Tanoura (near the well-known Takasakiyama monkey park) in Japan. Modelled on a hawk, the drone has a beak and eyes. It also carries a toy monkey that emits (recorded) alarm cries to scare monkeys, see Figure 2.6. Besides, with mobile scare-chase capability, the drone can not only repel monkeys, but also pursue them. By maximize monkey fear with simulating hawk predation on a young monkey, the drone can drive the monkeys back into the interior of the mountain forest. It is also mentioned in [104] that drone-assisted scare-chasing through GPS-tracking of a troop of monkey (containing at least one collared individual), can continue into the forest. Authors in a brief report [100] presented a case study using drones to mitigate human–elephant conflict on the borders of Tanzanian Parks. Specifically, [100] reported on field trials in northern Tanzania that employed drones for wildlife managers to move elephants away from conflict zones from a hundred meters away. Thereby enhance the safety of the farmers, wildlife managers and elephants. 10 drones were deployed during crop-raiding events at the peak of the maize ripening period in 2015 and 2016 in the Tarangire–Manyara and Serengeti ecosystems. The results show that elephants responded to the presence of a drone by departing rapidly from crop fields in 51 out of 100 trials. The authors claim that the use of drone to solve the elephants-human conflicts is both efficient and less expensive.

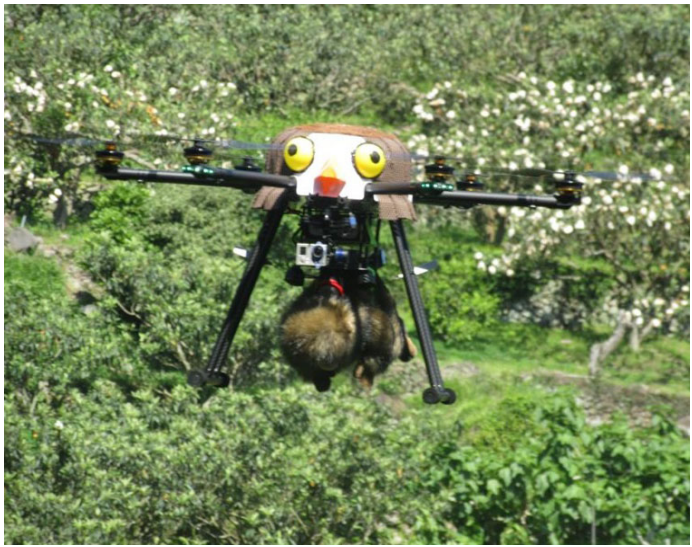

Figure 2.6: Use of drones for repelling fruit-raiding monkeys in the village of Tanoura in Japan.

Similar to the references we introduced in Section 2.2.2, drones for repelling animals can be the tool for protecting the animals themselves. For instance, to against poaching of rhinos, authors in [103] proposed a method using sirens and drones to elicit avoidance behaviour in white rhinoceros as an anti-poaching tactic. They claim that the use of drones to repel rhinos away from poaching hotspots could be a useful anti-poaching tactic. By experiments, authors in [103] found that Rhinos travelled significantly further in response to low-altitude drone flights than to higher altitude flights. And drones are superior at manipulating rhino movement than sirens.

Another important application of drones for repelling animals is on herding birds away from a prescribed area, such as an airport, developed by authors in [20] and [101]. In detail, references [20,101] developed a boundary control strategy, called the m-waypoint algorithm for enabling a single drone to herd a flock of birds away from the air space around an airport. The proposed algorithm is designed using a dynamic model of bird flocking based on Reynolds' rules. It allows a single pursuer drone to safely herd the bird flock without fragmenting it. The conditions under which bird flocks are exponentially stable to external perturbations are derived, and the performance of the proposed herding method was examined systematically by rigorous analysis in [20, 101]. The unique contribution of [20] is that the authors conducted real-world tests that demonstrated several facets of the proposed herding method on flocks of live birds reacting to a pursuer drone. The effectiveness of the proposed herding algorithm is also proved in tests for diverting a flock of birds approaching a prescribed area away from a protected zone around the area.

Based on existing literature on drones for repelling animals, we will further introduce the application of drones for repelling sharks and farm animals in Chapter 5 and 6, respectively.

## 2.4 Summary

In this chapter, we present a brief review of the existing literature related to our studied problems. For more in-depth reviews, readers are referred to the survey papers [36–38, 107–110].

# Chapter 3

# Autonomous Navigation of an Aerial Drone to Observe a Group of Wild Animals with Reduced Visual Disturbance

Protection of wild animals relies on understanding the interaction between the animals and their environment. With the ability to rapidly access rugged areas, aerial monitoring by drones is fast becoming a viable tool for ecologists to monitor wild animals. Unfortunately, this approach results in significant disturbance to different species of wild animals. Inspired by motion camouflage, this chapter explores a navigation method for a drone to covertly observe a group of animals and their habitat. Unlike existing motion camouflage navigation approaches that deceive a single target, we introduce a sliding mode based method that reactively navigates the drone to induce less optical flow on multiple targets' visual system. The proposed method is computationally simple and suitable for a drone to closely observe a group of moving animals with reduced visual disturbance. Computer simulations are conducted to demonstrate the performance of the proposed method.

# 3.1 Motivation

Global climate change is worsening the living conditions of wildlife. To preserve endangered wildlife and formulate better wildlife management strategies, ecologists and policymakers rely on wildlife monitoring to gather necessary information on the health of wildlife populations and their habitats [111]. A major problem in wildlife monitoring is the inability to arrive at wildlife habitats because of the complex terrain or the high risk of arrival. With superior maneuverability and flexible operation, drones, also known as unmanned aerial vehicles (UAVs), can help with solving this problem. Recent years have witnessed the increasing use of drones in different areas, including disaster relief [9], surveillance [3, 49], and many others. Particularly, drone technology has opened up a new way for ecologists to monitor wild animals, such as count animal groups and determine their gender by visible or thermal imaging [77]. Compared with previous methods, drones promise to revolutionize ecological research paradigms with their ability to accurately estimate wild animals' numbers and distribution at previously inaccessible areas and spatial resolutions. Moreover, aerial drone monitoring can provide significant ecological insights by offering a wide aerial viewpoint, enabling a clear overview of the animals' interaction with their habitat. Such insights are not previously possible using traditional land-based survey techniques.However, the use of drones is of biological and ethical concern. When drones are used to monitor wild animals, a major problem is the consequential disturbances to the animals. Recent studies [22–24] have revealed that drones could cause major interference to certain species, and many animals react negatively to drones' presence in biological field research. A recent systematic literature review [112] concludes that unmanned aircraft systems have become a new source of disturbance for wildlife. For example, studies have shown that drones have negative impacts on different species of birds [22]. Typical adverse reactions of birds to drones include severe panic responses, delayed return times to the nest, and nest abandonment [87] . Studies also evaluate the effect of drones on terrestrial animals. Ditmer *et al.* [23] show that black bears experienced consistently strong physiological responses such as raised heart rates in response to drones flying overhead. Bennitt *et al.* [24] deploy two drones to approach seven species of terrestrial mammalian wildlife (e.g., elephant, giraffe, zebra, etc.). The results show that drones can trigger behavioral responses in most species. Bennitt *et al.* [24] therefore conclude that enforced regulations on the use of drones in wildlife areas are necessary to minimize the disturbance.

To solve this problem, a study by Jarrod *et al.* [113] proposes the best practice for minimizing drone disturbance to wildlife in biological field research. They emphasize that minimum wildlife disturbance flight practices need to be exercised and drone trajectories that are potentially threatening should be avoided. In fact, there are growing appeals for researchers to develop effective strategies to safely apply drones to monitor wild animals to minimize the negative impacts [114,115]. The research to date has only proposed some basic disturbance-reducing solutions such as increasing the drones' altitude or keeping maximum useful distance to reduce the disturbance [115,116]. But other than these, more specific methods for drones to perform covert monitoring of wildlife are still under investigation.

Wildlife monitoring by drones can be broadly divided into two categories based on the mission types. The first typically involves flying overhead of the animals to monitor the distribution of animals in a given area [114], referred to as the "overflight". The second category is usually called "close-up" to closely inspect or observe a single animal or a small group of animals whose locations are known ahead of launching. In general, close-up monitoring will tend to induce stronger disturbance to subject animals. Several studies have emphasized that great caution should always be exercised when conducting close-up animal monitoring [113,114]. Particularly, wild animals can respond to both visual and auditory cues from drones in negative ways [112]. Studies show that for animals in a noisy colony, any sign of discomfort can be attributed to visual rather than auditory contact with the drones [87], and vice versa for animals in ecosystems with little environmental noise [117]. There have been some commercialised quiet alternatives to standard drones. For example, DJI Mavic Pro is claimed to be 60% quieter than its other models [118]. Adopting such drones to monitor animals can cause less noise disturbance.

This chapter concerns a specific problem of reducing the visual disturbance caused by the close-up wildlife observing drone. To address this problem, we take a group of terrestrial moving animals as an example and research on the drone's navigation laws to induce less visual disturbance. In particular, we design a metric to quantify the visual disturbance caused by the drone to the animals. With this metric, we formulate an optimization problem to minimize the maximum visual disturbance to the animals. We also propose the navigation laws that guide the drone to minimize the proposed visual disturbance metric while conducting a close-up observation task. In detail, the proposed navigation laws can navigate the drone to approach the animals from an original position, perform the observation, and fly back to the original position afterward. We conduct extensive computer simulations to show the performance of the proposed method.

The outline of this chapter is as follows: Section 3.2 Section discusses some relevant publications in the literature and explains the main contribution of this chapter. Section 3.3 presents the system model and states the studied optimization problem. Section 3.4 presents the proposed method. Section 3.5 gives some simulation results to demonstrate the performance of the proposed method. Finally, Section 3.6 concludes the chapter together with some future research directions.

## 3.2 Preliminaries

Camouflage is a widely used deception mechanism in nature. First discovered in hoverflies in 1995 by Srinivasan and Davey [119], motion camouflage as a stealth behaviour has been observed in different species of insects and animals, such as dragon-flies [120], bats [121], and falcons [122]. It allows a moving object to induce no optical flow on a target animal's visual system, and enables a predator or pursuer to have a degree of concealment to avoid attracting the target's attention.

**_Basic mechanism of motion camouflage:_** when a pursuer moves towards a moving prey, the former chooses its path so that it remains on the camouflage constraint lines (the straight line segment connecting the instantaneous position of the target and a fixed reference point), see Figure 3.1.

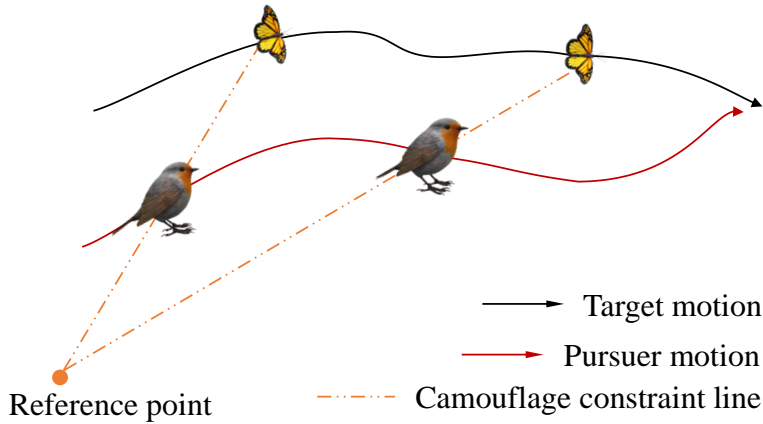

Figure 3.1: Illustration of motion camouflage: a pursuer flies towards a moving target, camouflage its motion by staying on the continually changing camouflage constraint lines.

Since with motion camouflage the pursuer induces no optical flow on the target's visual system, the target is unable to distinguish the moving pursuer from a steady object. This motion strategy allows a pursuer to successfully follow a target while concealing its motion even without any advantage in speed [119]. Srinivasan *et al.* [119] also demonstrate that motion camouflage could allow successful concealment against both homogenous and structured backgrounds. Although motion camouflage is simple in concept, experiments show that even humans, with our advanced visual system, can be fooled by this technique [123].

Motion camouflage has also been studied for robot systems. For example, various studies have proposed control algorithms to perform motion camouflage on robotic platforms (e.g., [48, 49, 124–126]). Specifically, Rañó *et al.* [124] present the first implementation of motion camouflage in real wheeled robots through a non-linear polynomial controller. It was found, however, this solution may produce undesirable trajectories, as the proposed controller learns from a number of data sets constructed using a computationally expensive heuristic method to define the motion camouflage trajectory. Strydom *et al.* [125] derive a motion camouflage guidance law by which a drone can pursue a moving target at a constant distance.

Experiments with a drone in a realistic virtual environment demonstrate that it is possible to remain well camouflaged using their proposed method, even with noisy state information. Savkin *et al.* [49] present a motion camouflage sliding mode based navigation law for a drone to survey a moving target based on bearing only measurements. Particularly, the authors do not assume that the drone knows the distance to the target or to the reference point. So the proposed method cannot achieve the surveillance from a specific standoff distance. Prasad *et al.* [126] introduce a Lyapunov-based control scheme for point-mass robots to perform motion camouflage. The authors use the direct method of Lyapunov to carry out the stability analysis and theoretically show that the equilibrium point is stable. Although extensive research has been carried out on motion camouflage navigation laws with one pursuer and one target, there is no study investigating the navigation law for a pursuer to deceive multiple targets.

The main contribution of this chapter is as follows:

- The originality of this work is that we explore a motion camouflage method for a drone to induce less visual attraction to a group of moving targets. The proposed method is computationally simple and belongs to the class of sliding mode control method.

- The importance of the proposed method is that it can be implemented on wildlife observing drones to solve a problem under investigation: closely observing wildlife and their habitat with reduced visual disturbance.

## 3.3 System Model and Problem Statement

In this section, we first present the models used in this work and then formally state the considered problem. We present a list of the main notations in Table 3.1.

We considered a wildlife observing drone flying in a three-dimensional (3D)

Table 3.1: Notations and Descriptions

| Notation | Description |
|---|---|
| $D(t)$ | Position of the drone |
| $\boldsymbol{h}(t)$ | Heading of the drone |
| $\boldsymbol{u}(t)$ | The control input to change $\boldsymbol{h}(t)$ |
| $v(t)$ | Speed of the drone (another control input) |
| $U_{\max}$ | The maximum of $\|\boldsymbol{u}(t)\|$ |
| $V_{\max}$ | The maximum of $v(t)$ |
| $F$ | Position of the farthest point on the habitat |
| $R$ | Position of the reference point |
| $L_u$ | The drone's maximum useful observing distance |
| $n(t)$ | Number of observable targets at time $t$ |
| $A_j$ | Position of target $j$ |
| $V_T$ | The maximum speed of target |
| $\boldsymbol{r}(t)$ | The vector from $D(t)$ to $R$ |
| $\boldsymbol{f}(t)$ | The vector from $D(t)$ to $F$ |
| $\boldsymbol{a_j}(t)$ | The vector from $D(t)$ to $A_j$ |
| $\beta_j$ | The bearing change at target $j$ |
| $\boldsymbol{b_j}(t)$ | The "steering" vector for target $j$ |
| $t_e$ | The time when the drone enters any target' visual field |
| $t_l$ | The time when the drone leaves all the targets' visual field |
| $D_s$ | The start of the *To-sphere* path |
| $D_o$ | The start of the *On-sphere* path |
| $\boldsymbol{s}(t)$ | The vector from $D(t)$ to $D_s$ |

space. Let

$$D(t) := [x(t), y(t), z(t)] \tag{3.1}$$

denote the drone's Cartesian coordinates (position) at time $t$. The motion of the
drone is described by the kinematic equations:

$$\dot{D}(t) = v(t)\boldsymbol{h}(t), \tag{3.2}$$

$$\dot{\boldsymbol{h}}(t) = \boldsymbol{u}(t), \tag{3.3}$$

where $\dot{D}(t)$ is the velocity vector of the drone. $\boldsymbol{h}(t) \in \mathbf{R}^3$ is the motion direction or
heading of the drone. $\|\boldsymbol{h}(t)\| = 1$ for all $t$. $v(t) \in \mathbf{R}$ is the linear velocity or speed
of the drone. $\boldsymbol{u}(t) \in \mathbf{R}^3$ is the vector applied to change the direction of the drone's
motion. The scalar variable $v(t)$ and the vector variable $\boldsymbol{u}(t)$ are the control inputs,
and the following constraints hold:

$$\|\boldsymbol{u}(t)\| \leq U_{\max}, \quad v(t) \in [-V_{\max}, V_{\max}] \tag{3.4}$$

$$\boldsymbol{h}(t) \cdot \boldsymbol{u}(t) = 0 \tag{3.5}$$

Here $\|.\|$ denotes the standard Euclidean vector norm, and "$\cdot$" denotes the scalar product of two vectors. $U_{\max}$ and $V_{\max}$ are given constants. The condition (3.5) guarantees that the vectors $\boldsymbol{h}(t)$ and $\boldsymbol{u}(t)$ are always orthogonal. The non-holonomic model (3.2), (3.3), (3.4), (3.5) have been used to describe the kinematics of many unmanned aerial vehicles; see, e.g., [127] and references therein. In real life application this kinematic model is often supplemented by a dynamic model of an aerial drone with advanced controllers and state estimators such as H-infinity controllers [128–131] and robust state estimators [132–138].

We aim at navigating the drone to observe a group of moving target animals (targets) and their habitat[1]. For better investigating the resources and changes on the habitat, as well as the interaction between the targets and their habitat, this work considers a particular scenario that the drone needs to has a clear view of the entire targeted habitat, while keeping the maximum useful observing distance to the habitat for reducing the disturbance. Let $F$ be the position of a pre-defined farthest point to the drone on the habitat. Let $L_u$ be the maximum useful observing distance of the drone. When performing the close-up observation, the drone's motion should be constrained on a sphere surface (**observation sphere**) with $F$ as the center and $L_u$ as the radius. $L_u$ and $F$ are assumed to be known and acquired by some previous measurements and mapping. Furthermore, we select a pre-defined fixed reference point over the ground. It may be some tall landmark such as the tip of a tree. Let $R$ be the position of the reference point, as shown in Figure 3.2.

Let $\mathcal{A} = \{A_j\}$, $j = 1, ..., n(t)$ denote the set of $n(t)$ observable targets' positions at time $t$. The observable targets are the targets that have line-of-sight (LoS) with the drone. During the observation, some targets may be blocked by obstacles such as trees and become unobservable. Thus, $n(t)$ may vary with time. We assume that the targets are moving on the ground that is not necessarily flat but might be a quite uneven terrain. Let $V_T$ be the maximum speed of the targets.

---

[1]The targeted habitat can be a pre-defined part of a large habitat and should be relatively small for the drone to complete a close-up observation.

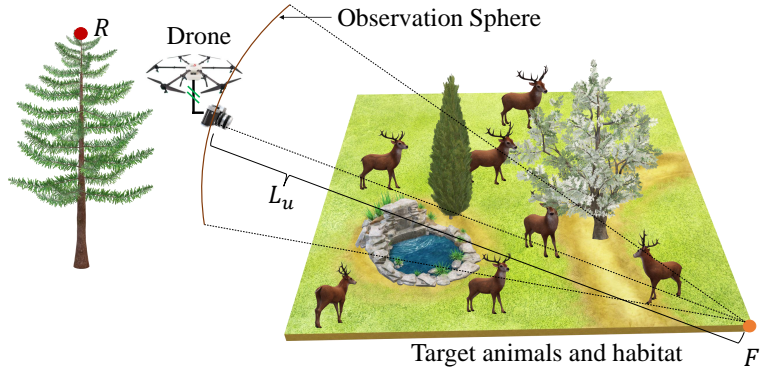

Figure 3.2: Illustration of the close-up observation of the target animals and their habitat. Where $F$ is a pre-defined farthest point to the drone on the habitat and $L_u$ is the maximum useful observing distance.

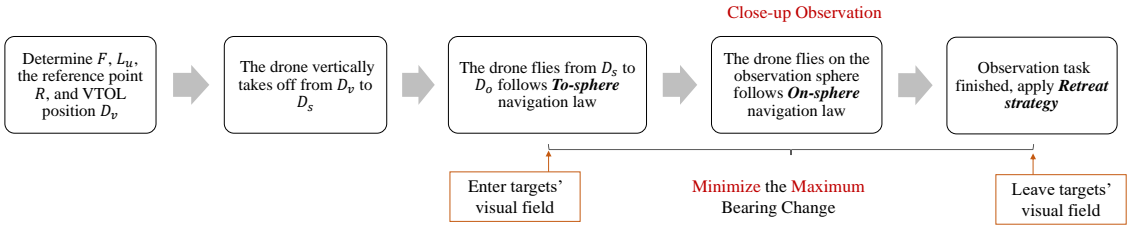

Figure 3.3: Overview of the proposed method.

**Available measurement:** we assume that at any time $t$, the drone can measure its current position $D(t)$ by onboard GPS chip. The drone also has measurements of the vector $\boldsymbol{r}(t)$ from $D(t)$ to the reference point $R$, and the vector $\boldsymbol{f}(t)$ from $D(t)$ to the farthest point $F$. Since $R$ and $F$ are pre-defined, $\boldsymbol{r}(t)$ and $\boldsymbol{f}(t)$ can be directly derived from $D(t)$. In addition, we assume the drone has measurements of the vector $\boldsymbol{a_j}(t)$, $j = 1, ..., n(t)$ from $D(t)$ to $A_j$, as shown in Figure 3.4. The measurements of $\boldsymbol{a_j}(t)$ can be acquired by a mounted thermal imaging camera and some image processing techniques.

The goal of this work is to develop a solution for a drone to observe a group of moving targets with reduced visual disturbance. Specifically, the drone starts from an initial position, flies to the observation sphere, conducts a close-up observation of the moving targets and their habitat, then flies back to the start position. During this process, the drone will enter and leave the targets' visual field. The navigation laws should reduce the visual disturbance to the targets when drone is within their visual field. Considering that the drone induces a minimized visual attraction or disturbance (no optical flow) on a target's retina if the drone remains on the camouflage constraint line connecting the current target position and the reference point $R$. We now introduce a variable $\beta$ as the *bearing change* to quantify the visual disturbance caused by the drone to a target.

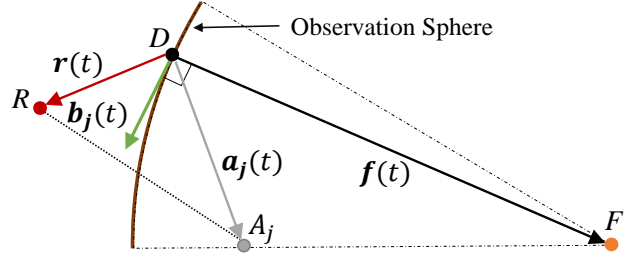

Figure 3.4: The measurements available at the drone, where $D$ is the position of the drone; $A_j$ is the position of target $j$; $R$ is the reference point; $F$ is the farthest point.

**Definition 3.3.1.** The bearing change at a target stands for the angle difference between the direction from its current position to $R$ and the direction from its current position to the drone, as shown in Figure 3.5. Clearly, $\beta_j$ equals to the angle between $\boldsymbol{a_j}(t)$ and $(\boldsymbol{a_j}(t) - \boldsymbol{r}(t))$.

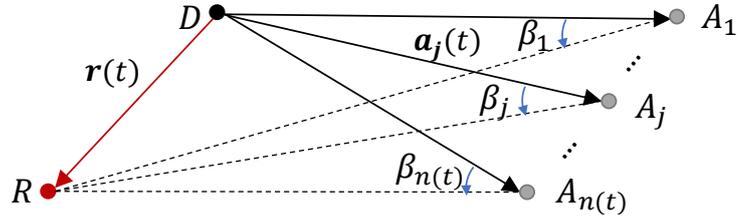

Figure 3.5: Illustration of the bearing changes $\beta$.

With $\boldsymbol{r}(t)$ and $\boldsymbol{a_j}(t)$, $\beta_j$ is obtained by:

$$\beta_j = \arccos(\frac{\boldsymbol{a_j} \cdot (\boldsymbol{a_j} - \boldsymbol{r})}{||\boldsymbol{a_j}||||\boldsymbol{a_j} - \boldsymbol{r}||}), \quad j = 1, ..., n(t). \tag{3.6}$$

Since the drone could induce different level of visual disturbance to the targets at different position. For a group of targets, the drone should minimize the maximum bearing changes it may induce. Thus, our objective function is formulated as follows:

$$\max_{j=1,\dots,n(t)} \beta_j \longrightarrow \min, \quad \forall t \in [t_e, t_l], \tag{3.7}$$

where $t_e$ and $t_l$ stand for the time when the drone enters and leaves the targets' visual field, respectively.

*Problem statement:* The optimization problem under consideration is stated as follows: when $t \in [t_e, t_l]$, for the given $F$, $L_u$, $R$, the measured vectors $\boldsymbol{a_j}(t)$, $j = 1, \dots, n(t)$, find the control inputs $v(t)$ and $\boldsymbol{u}(t)$ that navigate the drone to minimize the function (3.7).

## 3.4 Proposed Solution

This section introduces the navigation laws that guide the drone to minimize the function (3.7) while conducting a close-up observation of multiple moving targets.

Let $D_v$ be the initial position of the drone on the ground. $D_v$ is also called the vertical take-off and landing (VTOL) position. Our covert close-up observing solution is first letting the drone take-off vertically to arrive a position $D_s$ that is close to the reference point $R$. Then, the drone flies to the observation sphere to conduct the close-up observation. Once finishing the observation, the drone applies a retreat strategy to fly back to the VTOL position $D_v$. We assume that the close-up observation starts when the drone arrives at the observation sphere. Let $D_o$ be the arriving position.

We first introduce *On-sphere* navigation law for guiding the drone to minimize the function (3.7) while flying on the observation sphere after arriving at $D_o$. *To-sphere* navigation law that guides the drone flies to $D_o$ and the *Retreat Strategy* will be introduced afterward. An overview of the proposed method can be seen in Figure 3.3. The illustration of the *Take-off* path, *To-sphere* path, and *On-sphere* path of the drone is as shown in Figure 3.6.

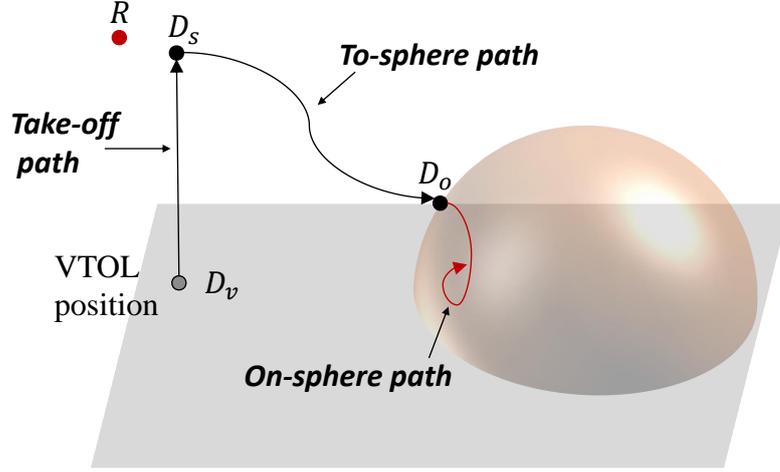

Figure 3.6: Illustration of the *Take-off* path, *To-sphere* path, and *On-sphere* path. $D_v$ is the VTOL position. $D_s$ and $D_o$ are the start and end of the *To-sphere* path, respectively.

### 3.4.1 On-sphere Navigation Law

Let $\boldsymbol{\mu}_1$ and $\boldsymbol{\mu}_2$ be non-zero 3D vectors. We now introduce a function $W(,)$ mapping from $\mathbb{R}^3 \times \mathbb{R}^3$ to $\mathbb{R}^3$ as

$$W(\boldsymbol{\mu}_1, \boldsymbol{\mu}_2) := \begin{cases} 0, & w(\boldsymbol{\mu}_1, \boldsymbol{\mu}_2) = 0, \\ ||w(\boldsymbol{\mu}_1, \boldsymbol{\mu}_2)||^{-1} w(\boldsymbol{\mu}_1, \boldsymbol{\mu}_2), & w(\boldsymbol{\mu}_1, \boldsymbol{\mu}_2) \neq 0, \end{cases} \tag{3.8}$$

where

$$w(\boldsymbol{\mu}_1, \boldsymbol{\mu}_2) := \boldsymbol{\mu}_2 - (\boldsymbol{\mu}_1 \cdot \boldsymbol{\mu}_2) \boldsymbol{\mu}_1. \tag{3.9}$$

**Remark 3.4.1.** The vector $W(\boldsymbol{\mu}_1, \boldsymbol{\mu}_2)$ defined by rules (3.8), (3.9) is a vector in the plane of vectors $\boldsymbol{\mu}_1$ and $\boldsymbol{\mu}_2$ that is orthogonal to $\boldsymbol{\mu}_1$ and directed "towards" $\boldsymbol{\mu}_2$, as shown in Figure 3.7. Moreover, $W(\boldsymbol{\mu}_1, \boldsymbol{\mu}_2) = 0$ if $\boldsymbol{\mu}_1$ and $\boldsymbol{\mu}_2$ are co-linear, and $||W(\boldsymbol{\mu}_1, \boldsymbol{\mu}_2)|| = 1$ otherwise.

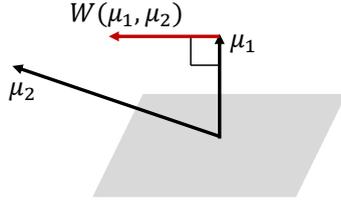

Figure 3.7: Illustration of the vector $W(\boldsymbol{\mu}_1, \boldsymbol{\mu}_2)$.

In addition, we introduce the function $g\left(\boldsymbol{\mu}_1, \boldsymbol{\mu}_2\right)$ as follows

$$g\left(\boldsymbol{\mu}_1, \boldsymbol{\mu}_2\right) := \begin{cases} 1, & (\boldsymbol{\mu}_1 \cdot \boldsymbol{\mu}_2) > 0, \\ -1, & (\boldsymbol{\mu}_1 \cdot \boldsymbol{\mu}_2) \le 0, \end{cases} \tag{3.10}$$

Let $\boldsymbol{p_j}(t)$ be the vector from $D$ to the nearest point on the camouflage constraint line connecting $A_j$ and $R$, as shown in Figure 3.8. With $\boldsymbol{r}(t)$ and $\boldsymbol{a_j}(t)$, $\boldsymbol{p_j}(t)$ is given by:

$$\boldsymbol{p_j} = \frac{(\boldsymbol{a_j}^2 - \boldsymbol{r} \cdot \boldsymbol{a_j})\boldsymbol{r} + (\boldsymbol{r}^2 - \boldsymbol{r} \cdot \boldsymbol{a_j})\boldsymbol{a_j}}{\boldsymbol{r}^2 + \boldsymbol{a_j}^2 - 2\boldsymbol{r} \cdot \boldsymbol{a_j}}. \tag{3.11}$$

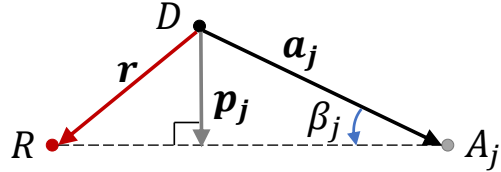

Figure 3.8: Illustration of the vector $\boldsymbol{p_j}(t)$ as the shortest path from $D$ to the camouflage constraint line connecting $A_j$ and $R$.

Furthermore, we introduce a 'steering' vector $\boldsymbol{b_j}(t)$ as follows:

$$\boldsymbol{b_j}(t) = W(\boldsymbol{f}(t), \boldsymbol{p_j}(t)). \tag{3.12}$$

According to Remark 3.4.1, $\boldsymbol{b_j}(t)$ is always orthogonal to $\boldsymbol{f}(t)$, as shown in Figure 3.4.

We are now in a position to present the following *On-sphere* navigation law:

**A1.** Find the target $m$ with the maximum bearing change

$$\beta_m = \max_{j=1,...,n(t)} \beta_j, \quad m \in 1,...,n(t). \tag{3.13}$$

**A2.** Calculate the "steering" vector

$$\boldsymbol{b_m}(t) = W(\boldsymbol{f}(t), \boldsymbol{p_m}(t)). \tag{3.14}$$

**A3.** Apply the control inputs

$$\boldsymbol{u}(t) = U_{\max}g(\boldsymbol{h}(t), \boldsymbol{b_m}(t))W(\boldsymbol{h}(t), \boldsymbol{b_m}(t)),$$
$$v(t) = V_{\max}g(\boldsymbol{h}(t), \boldsymbol{b_m}(t)). \tag{3.15}$$

**A4.** Repeat **A1**-**A3** until finishing the observation task.

**Theorem 3.4.1.** The proposed *On-sphere* navigation law guarantees that the drone stays on the observation sphere while navigating the drone to minimize the function (3.7) in the most efficient way.

**Proof 1.** As shown in Figure 3.8, at any time $t$, $\boldsymbol{p_m}(t)$ is the fastest path for the drone to minimize the maximum bearing change $\beta_m$. By definition, $\boldsymbol{b_m}(t)$ is always orthogonal to $\boldsymbol{f}(t)$ and directed "towards" $\boldsymbol{p_m}(t)$. Therefore, $\boldsymbol{b_m}(t)$ is a tangent vector of the observation sphere and directed "towards" the fastest path to minimize the function (3.7). Moreover, from the definitions of the functions $W(,)$ and $g(,)$, (3.15) adjusts the heading of the drone towards $\boldsymbol{b_m}(t)$ with the maximum angular speed and maximum linear speed. Thus, the proposed *On-sphere* navigation law navigates the drone to to minimize the function (3.7) in the most efficient way, while keeping the drone on the observation sphere. This completes the proof of Theorem 3.4.1.

### 3.4.2 To-sphere Navigation Law

Once the drone arrives at position $D_s$ by vertical take-off from $D_v$, it should then fly to the observation sphere to conduct the close-up observation task.

**Remark 3.4.2.** We assume that the range of the target's visual field is pre-known or can be estimated. We also assume that the observation distance of the camera mounted on the drone is larger than the range of the target's visual field. $D_v$ and $D_s$ should be carefully selected so that $D_s$ is close to the reference point $R$, and the VTOL process is outside any target's visual field.

We now introduce the *To-sphere* navigation law that navigates the drone to minimize the function (3.7) while flying from $D_s$ to $D_o$. We first introduce a vector $\boldsymbol{b}^*(t)$ as follows:

$$\boldsymbol{b}^*(t) = W(\boldsymbol{p_m}(t), \boldsymbol{f}(t)). \tag{3.16}$$

Moreover, we introduce a new "steering" vector

$$\boldsymbol{b_m}(t) = \begin{cases} \boldsymbol{f}(t), & n(t) = 0, \\ \boldsymbol{p_m}(t) + \boldsymbol{b}^*(t), & n(t) > 0. \end{cases} \tag{3.17}$$

Then, the *To-sphere* navigation law repeats (3.6), (3.13), (3.16), (3.17) and (3.15) until the drone arrives at a point on the observation sphere (i.e., $D_o$). The proposed *To-sphere* and *On-sphere* navigation laws belong to the class of sliding-mode control laws (see, e.g., [139]).

**Theorem 3.4.2.** The proposed *To-sphere* navigation law navigates the drone to minimize the function (3.7) from $D_s$ to $D_o$.

**Proof 2.** Since $\boldsymbol{f}(t)$ is the vector from $D(t)$ to $F$, the new "steering" vector $\boldsymbol{b_m}(t) = \boldsymbol{f}(t)$ will navigate the drone to approach the observation sphere before it detected any target (i.e., $n(t) = 0$). When $n(t) > 0$, $\boldsymbol{b}^*(t)$ is always orthogonal to $\boldsymbol{p_m}(t)$ and directed "towards" $\boldsymbol{f}(t)$ (i.e., "towards" $F$). Therefore, the new "steering" vector $\boldsymbol{b_m}(t)$ consists of two orthogonal components: $\boldsymbol{p_m}(t)$ as the fastest path to minimize the function (3.7) and $\boldsymbol{b}^*(t)$ for navigating the drone to approach the observation sphere without affecting minimizing the function (3.7). Therefore, the proposed *To-sphere* navigation law navigates the drone to minimize the function (3.7) from $D_s$ to $D_o$. This completes the proof of Theorem 3.4.2.

### 3.4.3 Retreat Strategy

We now introduce the *Retreat Strategy* for the drone to retrace the paths back to the VTOL position $D_v$ after finishing the observation task. Firstly, the drone needs to fly back to $D_s$ from the observation sphere while minimizing the function (3.7). Let $s(t)$ be the vector from the drone's current position $D(t)$ to $D_s$, Similar to the *To-sphere* navigation law, we define the new $b^*(t)$ and $b_m(t)$ as

$$b^*(t) = W(p_m(t), s(t)). \tag{3.18}$$

$$b_m(t) = \begin{cases} p_m(t) + b^*(t), & n(t) > 0, \\ s(t), & n(t) = 0, \end{cases} \tag{3.19}$$

Then, the proposed *Retreat Strategy* is as follows:

---
**Algorithm 1** Retreat Strategy
---
**Input:** $a_j(t), j = 1, ..., n(t), r(t), s(t), D_s, D_v, D(t)$
1: Repeat (3.6), (3.13), (3.18), (3.19) and (3.15) until the drone arrives at $D_s$.
2: Once the drone arrives at $D_s$, it flies back to $D_v$ by vertical landing. =0

---

**Theorem 3.4.3.** The proposed method navigates the drone to minimize the function (3.7) when the drone is within the targets' visual field (i.e., $t \in [t_e, t_l]$)

**Proof 3.** Similar to the *To-sphere* navigation law, when $n(t) > 0$, the new $b^*(t)$ is always orthogonal to $p_m(t)$ and directed "towards" $D_s$. Thus, the new "steering" vector $b_m(t)$ will navigate the drone to approach $D_s$ while minimizing the function (3.7). Moreover, $b_m(t) = s(t)$ will navigate the drone to approach $D_s$ after it lost the detection of all the targets (i.e., $n(t) = 0$). Since we assume that the observing distance of the drone is larger than the range of the target's visual field, the drone will leave all the targets' visual field before $n(t) = 0$. Thus, the proposed method navigates the drone to minimize the function (3.7) when the drone is within the targets' visual field (i.e., $t \in [t_e, t_l]$). This completes the proof of Theorem 3.4.3.

## 3.5   Simulations

In this section, we present computer simulation results conducted in MATLAB to confirm the performance of the proposed method. The parameters used in the simulations are shown in Table 3.2. The targets move on a simulated uneven square zone (200 m by 200 m). For simplicity, we ignore the VTOL process in the simulations and assume $D_s$ is the drone's start and end position. Same as the paper [49], to avoid fast switching (chattering) that is typical for sliding mode controllers in our simulations, we use the standard sliding mode control trick of approximating the sign type functions in (3.15) by piecewise linear continuous saturation type functions. Alternatively, the throttle control can be used to avoid fast switching and reduce the control effort.

Table 3.2: Simulation Parameter Values

| Parameters | Values | Parameters | Values |
|---|---|---|---|
| $V_{\max}$ | $10 \ m/s$ | $U_{\max}$ | $5 \ m/s^2$ |
| $F$ | $[200,200,0]$ | $R$ | $[10,10,40]$ |
| $D_s$ | $[20,20,38]$ | $L_u$ | $200 \ m$ |
| $V_T$ | $2 \ m/s$ | Simulation period | $120 \ s$ |

**Benchmark method:** for comparison, the benchmark method we adopt is that the drone first flies towards $F$ to reach the observation sphere, then it stays at the arrival position until the task is completed. Afterwards, the drone flies directly towards $D_s$.

We first present a simulation of observing two moving targets, i.e., Case 1, as shown in Figure 3.9. An overview of the observation process is as shown in Figure 3.9a, where the translucent surface stands for the observation sphere. The trajectories of the drone and the targets during the simulation of 120 seconds are shown in Figures 3.9b and 3.9c, respectively. We assume that the targets' positions can be detected by the drone during the entire simulation. The *To-sphere*, *On-sphere* and *Retreat* trajectories of the drone are as shown in Figure 3.9b, with the moving directions marked by the black, red, and blue arrows, respectively. We can see the process of the drone flying from $D_s$ to the observation sphere, flying on the observation sphere, and flying back to $D_s$. In Figure 3.9c, the initial positions of the target are represented by the cyan balls, and the moving directions of the targets are marked by arrows. Target 1 moves along an S-shape trajectory. Target 2 moves along a straight trajectory. During the simulation, our method reactively navigates the drone to minimize the maximum bearing changes according to the moving targets' positions. For example, from Figures 3.9a and 3.9c, we can see that target 1 is climbing a hill at the second half of its trajectory, and the altitude difference between the two targets is increasing. In this case, our method navigates the drone flying upward to minimize the maximum bearing changes, which results
in the upward part of the drone's *On-sphere* trajectory, as shown in Figure 3.9b.

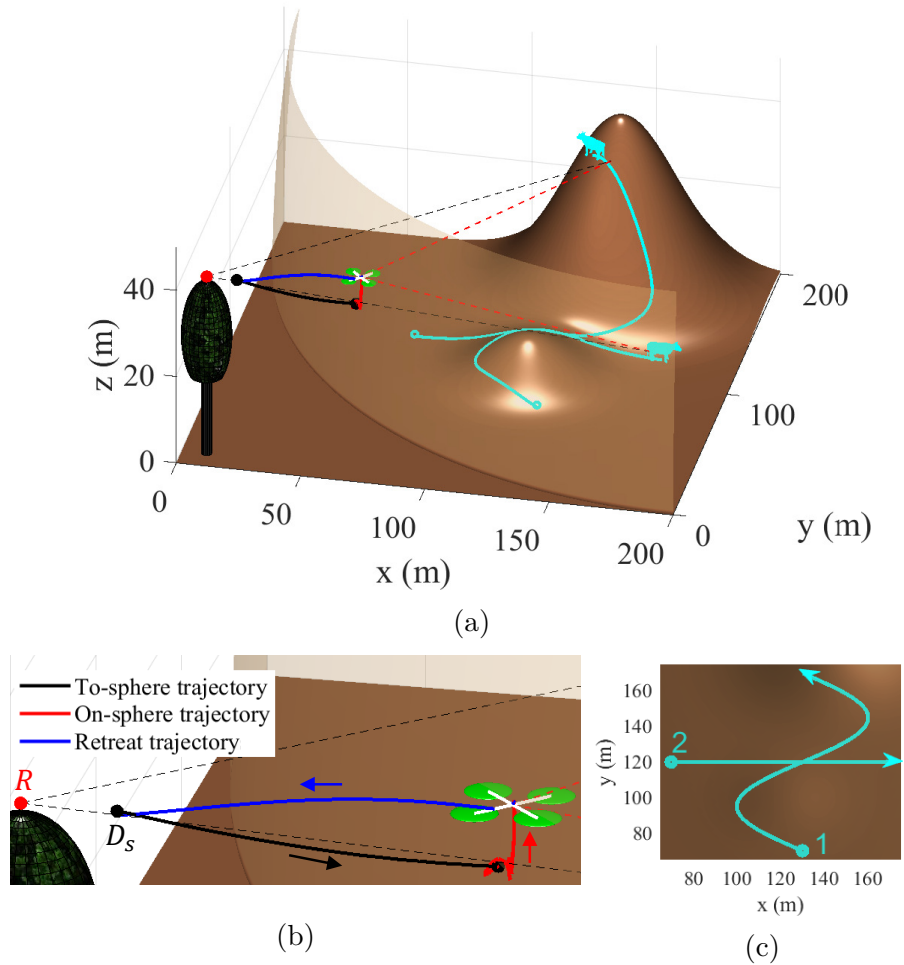

(a)

(b)                                    (c)

Figure 3.9: Case 1: observation of two moving targets (a video recording the movements is available online at: https://youtu.be/iW7fIwgSeTw). (a) Overview of the observation process. (b) *To-sphere*, *On-sphere* and *Retreat* trajectories of the drone. (c) The trajectories of two targets.

The comparison between the proposed method and the benchmark method in
terms of the bearing changes is as shown in Figure 3.10.  We can see that the
bearing changes of the proposed method are always lower than that of the benchmark
method, which means the proposed method tends to attract less attention from the
targets.  Specifically, Figure 3.10 shows that the maximum recorded bearing change
(i.e., the maximum recorded $\beta_m$) of the proposed method is 15.5°, which is 28.9%
lower than the maximum recorded $\beta_m$ of the benchmark method (i.e., 21.8°).

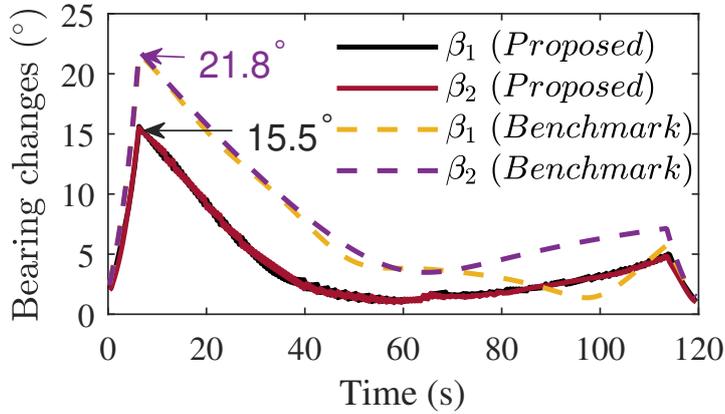

Figure 3.10: Case 1: comparison between the proposed method and the benchmark
method.

We show the bearing changes, the drone-to-targets distance, and the drone-to-$F$
distance of Case 1 in Figure 3.11.  The drone arrives at the observation sphere at
time $t_a = 7.4\ s$, and leaves the observation sphere at $t_f = 113.2\ s$.  As shown in
Figure 3.11, the two targets' bearing changes $\beta_1$ and $\beta_2$ are almost the same during
the entire simulation.  Note that only in this way the maximum bearing change $\beta_m$
can be minimized.  Otherwise, the drone can always fly closer to the camouflage
constraint line of the target with the larger bearing change to decrease $\beta_m$.  Thus,
Figure 3.11 shows that the proposed method can effectively minimize the maximum
bearing change in Case 1.  Moreover, the distance between the drone and $F$ keeps
at 200 $m$ (i.e., $L_u$) when $t_a \leq t \leq t_f$, which verifies that our proposed *To-sphere*
navigation law guarantees that the drone stays on the observation sphere.  Usually,

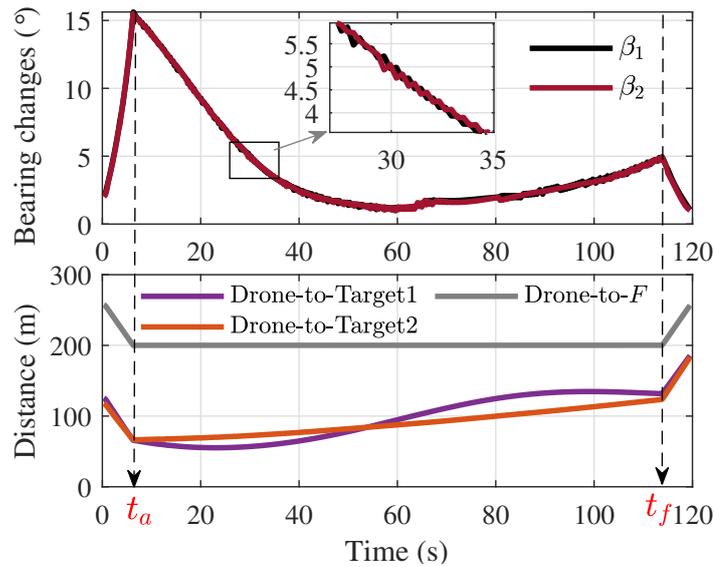

Figure 3.11: Case 1: bearing changes and the distance from the drone to $F$ and to the targets.

a larger drone-to-targets distance tends to induce smaller bearing changes. From Figure 3.11, the bearing changes increase sharply when $t < t_a$. This is because of the rapid decrease of drone-to-targets distance when the drone flies to the observation sphere. The reduction of the bearing changes at $t > t_f$ is because of the same reason.

In addition, a smaller target-to-target distance tends to induce smaller bearing changes. Figure 3.9c shows that the two targets are getting closer to each other in the first half of their trajectories, and are drifting away in the second half of their trajectories, which can explain the decrease of bearing changes at $t_a \leq t < 60\ s$, and the increase of the bearing changes at $60\ s \leq t \leq t_f$. The rate of change of the bearing changes at $t \geq 60\ s$, however, is lower than that of $t < 60\ s$. The reason is that the targets are moving away from the drone when $t > t_a$, and a larger drone-to-targets distance results in a slower change of the bearing changes.

To investigate the impacts of the maximum speed of the targets, i.e., $V_T$, on our method, we also simulate Case 1 with $V_T = 5\ m/s$. The comparison of the bearing changes with $V_T = 2\ m/s$ and $V_T = 5\ m/s$ can be seen in Figure 3.12. The targets' trajectories remain the same, so the simulated period of $V_T = 5\ m/s$ is shorter than $V_T = 2\ m/s$. Figure 3.12 shows that our method can minimize the maximum bearing change when $V_T = 5\ m/s$, because $\beta_1$ and $\beta_2$ are still overlapping during the simulation period.

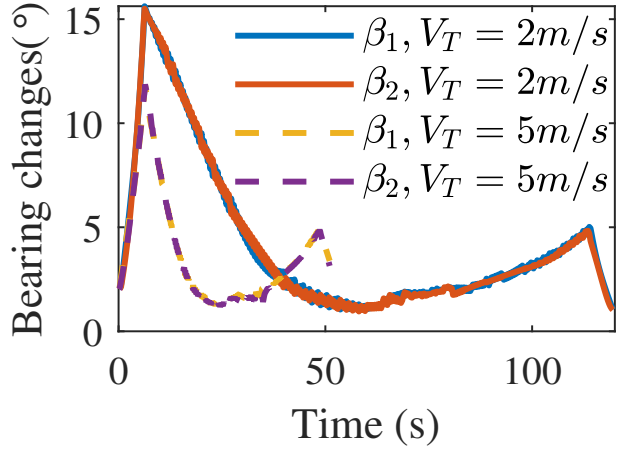

Figure 3.12: Case 1: bearing changes with different target speeds.

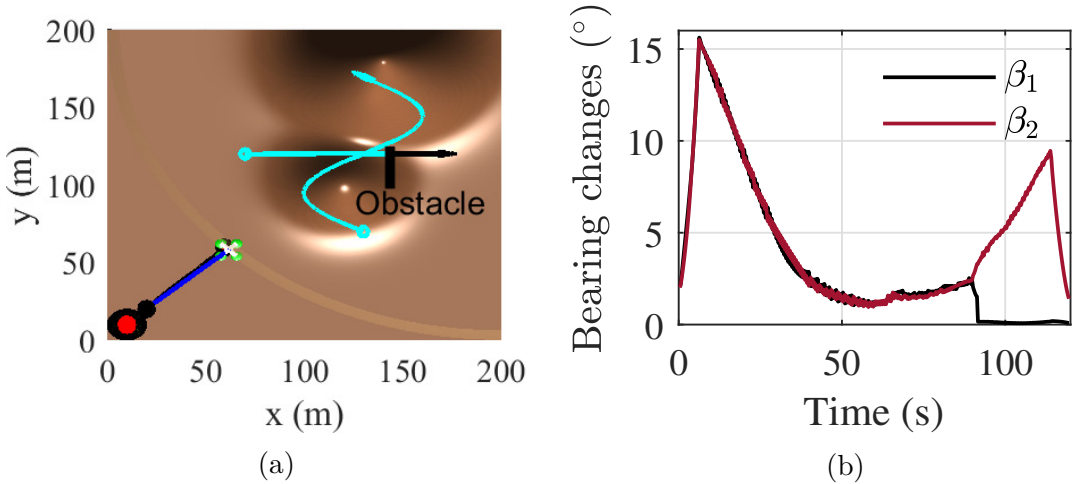

(a)

(b)

Figure 3.13: Case 1: impact of the obstacle. (a) Overview of the observation process. (b) Bearing changes

In addition, the proposed method is suitable for the situation that some targets been blocked by obstacles during the observation. Figure 3.13 reports this situation using Case 1 as an example, where target 2 is blocked by an obstacle and become undetectable by the drone at $t \geq 90\ s$. From Figure 3.13(b), we can see that after $t = 90\ s$, $\beta_1$ is close to zero, which means the drone stays close to the camouflage constraint line of target 1 to minimize the bearing change of the only detectable target (i.e., target 1).

We conduct more simulations with ten moving targets, i.e., Case 2, as shown in Figure 3.14. The targets move along some random curvy trajectories. Figure 3.14a presents the overview of the observation process. Figure 3.14b presents the comparison between the proposed method and the benchmark method in terms of the maximum bearing change $\beta_m$. We can clearly see that the proposed method out-performances the benchmark method because $\beta_m$ of the proposed method is significantly lower during the simulation period. Moreover, the maximum recorded $\beta_m$ of the proposed method is 22.2°, which is 33.1% lower than the maximum recorded $\beta_m$ of the benchmark method (i.e., 32.2°). Compared with the 28.9% reduction of the maximum recorded $\beta_m$ in Case 1, the proposed method performs better with more targets.

Figure 3.14c shows the bearing changes and the distances from the drone to $F$. It can be observed from Figure 3.14c that the envelope of the bearing change lines always consists of two or more bearing change lines. This means the maximum bearing change is always the same as the second maximum bearing change. Thus, Figure 3.14c shows that the proposed method can effectively minimize the maximum bearing change with ten moving targets. Same as Case 1, the distance between the drone and $F$ keeps at $L_u = 200 \ m$ when $t_a \leq t \leq t_f$, which verifies that the drone stays on the observation sphere during this period.

We are also interested in the impacts of the measurement errors on our method. We add random noise to the measured positions of the targets, and the amplitude of the noise is from 2% to 10% of the real measurement. We conduct 20 simulations with two targets and ten targets independently for each value. The results are as shown in Figure 3.15. Under the measurement error, the average impact is relatively small. For example, with 2% measurement error, the maximum error of $\beta_m$ is less than 0.3° for both cases. Here the maximum error of $\beta_m$ means the difference between the maximum recorded bearing change $\beta_m$ under the measurement error and the $\beta_m$ without measurement error during the simulation period. Moreover, even under 10% measurement error, the maximum error of $\beta_m$ is less than 2° for both cases. Compared with the 11° reduction of the maximum recorded $\beta_m$ with the proposed method in Case 2 (see, Figure 3.14b), the impact of measurement errors on the disguising intention is not significant.

## 3.6 Conclusions

In this chapter, we considered the problem of reducing the visual disturbance caused by the close-up wildlife observing drone. An optimization problem was formulated with the objective of minimizing the maximum visual disturbance (indicated by bearing changes) of multiple moving targets. To solve this problem, we proposed the sliding mode based navigation laws that guide the drone to minimize the maximum bearing changes while conducting a close-up observation task. The present study provides one of the first investigations into how to reduce the negative impacts of wildlife observing drones by motion control. The effectiveness of the proposed method was tested via computer simulations. One of our future efforts is to extend the current work to dynamic environments with either moving or static obstacles that the drone has to avoid collisions with. Another research direction is to explore the methods for reducing the auditory disturbance of wildlife observing drones.

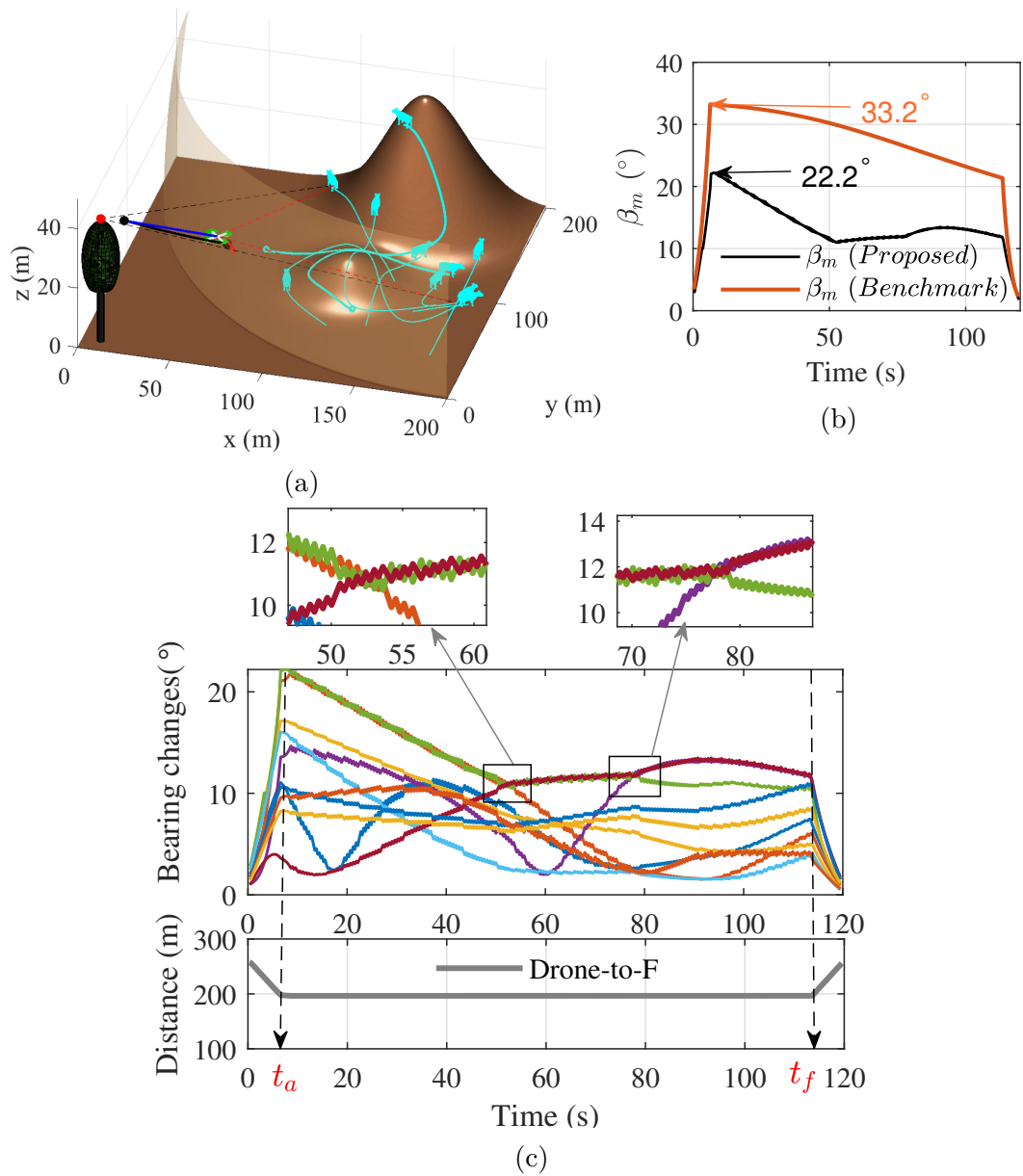

(a)

(b)

(c)

Figure 3.14: Case 2: Observation of ten moving targets (a video recording the movements is available online at: https://youtu.be/BdfzDrIfPPw). (a) Overview of the observation process. (b) Comparison between the proposed method and the benchmark method in terms of $\beta_m$. (c) The bearing changes and the distances from the drone to $F$.

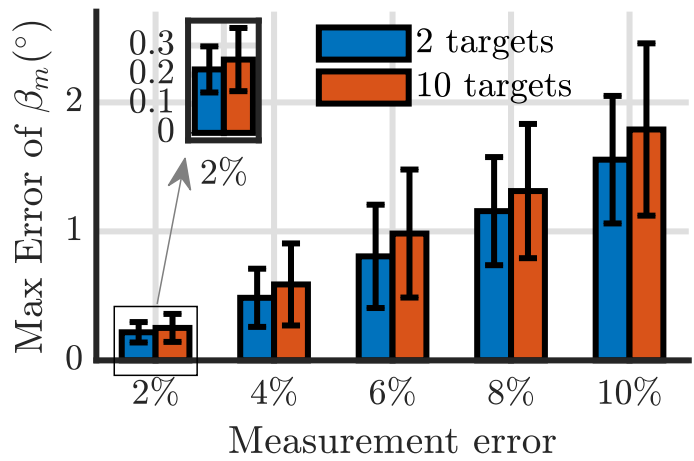

Figure 3.15: Impact of measurement errors.

# Chapter 4

# Efficient Optimal Backhaul-aware Deployment of Multiple Drone-Cells Based on Genetic Algorithm

Drone-cell is a promising solution for providing GSM/3G/LTE/5G mobile networks to victims and rescue teams in disaster-affected areas. One of the challenges that drone-cells are facing is the limitation of reliable backhaul links. In this chapter, we study the optimal deployment problem of a group of drone-cells deployed in a disaster area with limited backhaul communication ranges, aiming at maximizing the number of served users. Two approaches, exhaustive search algorithm and a computationally efficient genetic algorithm are proposed, and the optimal 2D backhaul-aware deployments of multiple drone-cells are found for each approach. We also introduce a restart-strategy to enhance searching efficiency and avoid local optima for the proposed genetic algorithm. Simulations show that the proposed genetic algorithm can save the computing time up to 99.927% compared with the exhaustive search algorithm, and the restart-strategy helps the probability of finding the global optimum by the proposed genetic algorithm increased from 12% to 92%.

## 4.1 Motivation

Establishing emergency communication networks has long been a difficult problem in disaster areas, where communication can save lives. With the capability of rapidly moving communication supply towards demand when required, low-altitude unmanned aerial vehicles equipped with base stations, i.e., drone-cells, have recently attracted a lot of attention [26]. As a promising solution to temporarily provide cellular networks in an area that has lost coverage, drone-cells can serve as aerial base stations with a quick deployment opportunity [88]. One of the biggest challenges, however, is to determine the optimal deployment of drone-cells so that users can benefit the most.

A considerable amount of literature has been published on drone-cells' placement problem. Reference [88] formulated the 3D placement problem into a mixed-integer non-linear problem (MINLP) and solved the problem by bisection search and the interior-point optimizer of the MOSEK solver. Reference [18, 25, 89] studied the problem of deploying unmanned aerial base stations to serve mobile users based on apriori user density function, which reflects the traffic demand at a certain position in the operating area. Reference [90, 91] considered the drone-cells' deployment problem without apriori user locations. Besides the mobility, another major difference between a ground cell and a drone-cell is that the latter one relies on wireless links for the backhaul connection, while a ground-cell usually has a fixed wired backhaul link. Therefore, one of the major limitations in drone-cells' deployment is the availability of reliable wireless backhaul links. While researchers have considered various aspects regarding the deployment of drone-cell for wireless coverage, the backhaul limitations of drone-cell have not been treated in many details. The backhaul link of drone-cell may be satellite-based, dedicated, or in-band. Satellite-based backhauling requires the drone to be equipped with a satellite transceiver for establishing the connection via a backhaul satellite. Dedicated backhauling could be free-space optical communication (FSO) or mmWave link between the drone and core networks. As for in-band backhauling, the main technology currently used in the wireless backhaul links of LTE or Wi-Fi is based on RF microwave [94]. Considering that the satellite transceiver is both expensive and energy inefficient, the drone-cells with dedicated or in-band backhauling are more practical.

Existing research recognizes the critical role played by the backhaul limitation while design and deploying the drone-cells include [26, 94, 95]. Reference [95] proposed a framework that utilizes the flying capabilities of the UAV-BSs as the main degree of freedom to find the optimal precoder design for the backhaul links, user-base station association, UAV-BS 3D hovering locations, and power allocations by exhaustive search. Reference [26] investigated the optimal 3D placement of a drone-cell over an urban area with users having different rate requirements, considering both the wireless backhaul peak rate and the bandwidth of a drone-cell as the limiting factors. Particularly, Reference [26] adopt the branch-and-bound method and exhaustive search in the step size of 100 meters to search drone-cell's 3D location for maximizing the total number of served users and sum-rates, which has the same problem as [95] that the solutions are inefficient and cannot guarantee accurately optimal results. Moreover, these two approaches are simply assuming that the drone-cell are connected to remote ground-cells, which is not practical for ignoring the communication range constraint of the backhaul link. In [89], the authors introduces using the robust extended Kalman filter to estimate users' locations based on the received signal strength indication, and proposes a decentralized algorithm to find a locally optimal solution for coverage maximizing while considering both the collision avoidance and backhaul limitation of drone-cells, the results, however, may not be globally optimal.

In this chapter, we study the optimal deployment problem of a group of drone-cells deployed in a disaster-affected area with limited backhaul communication ranges, aims to maximize the number of served users. Particularly, the backhaul communication of drone-cells can be achieved by wireless link direct to a ground base station (GBS) or through the wireless links to the other drone-cells as the bridge to the GBS, which is connecting to the core network. The problem is an NP-Hard problem. We first proposed an exhaustive search algorithm that can find the quasi-optimal de-ployment for the drone-cells. After that, a computationally efficient algorithm based on GA is proposed to cope with the complexity of the problem; however, it could potentially produce, on some occasions, poor solutions. A subsequent restart-strategy for the GA to enhance searching efficiency and avoid local optima is introduced. The rest of this chapter is organized as follows. In Section 4.2, the system model is presented. Two algorithms, including exhaustive search algorithm and GA to find the optimal backhaul-aware deployment of multiple drone-cells are described in Section 4.3, followed by a detailed presentation of the results and related discussions in Section 4.4.

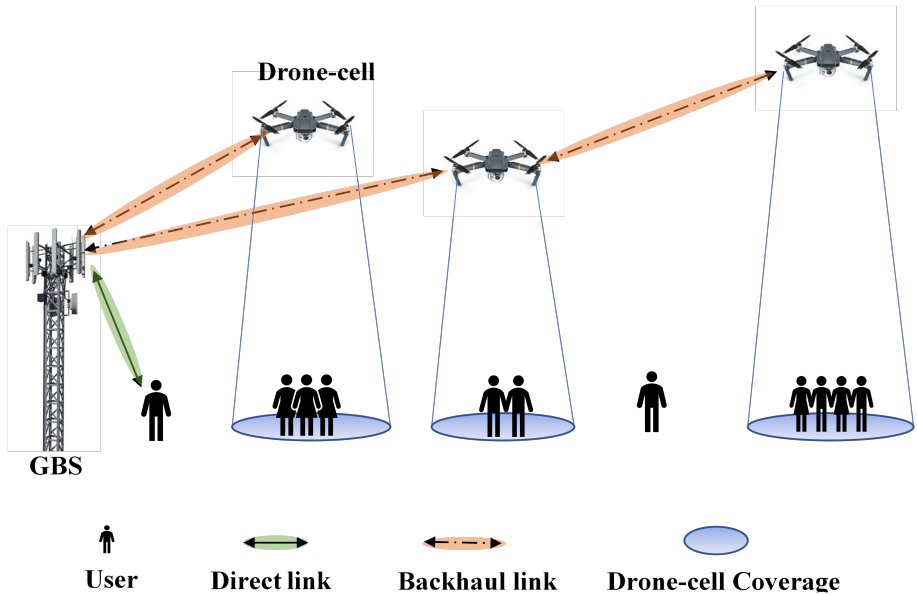

Figure 4.1: An illustration of the considered scenario.

## 4.2 System Model and Problem Statement

Consider a relative lager disaster-affected where most of the existing g cellular GBSs were manufactured because of the disaster, resulting in that the users in such an area have no access to the cellular network. In this case, a group of drone-cells is deployed to build a multi-tier drone cellular network. The first tier includes GBSs that, on the one hand, provide direct access links to users within its coverage and on the other hand, provides backhaul links to drone-cells. The second tier represents the drone cells and the users associated with them. An illustration of the considered scenario is shown in Figure 4.1. With this regard, we considered a transmission system consisting of $m$ drone-cells labeled $i = 1, 2, \ldots, m$ with $k$ surviving GBS. Some drone-cells should be able to directly communicate to any of the GBS through the backhaul link. These drone-cells can also bridge the other remote drone-cells to GBSs through backhaul links so that all the drone-cells are connecting with the core network. We assume that some very remote drone-cell may rely on a long backhauling bridge consists of two or more drone-cells to connect to the GBS. Since the network is built for emergency use, we assume the drone-cells try to serve as many users as possible, regardless of their rate requirements. Practically, the number of available drone-cells is often limited, while the disaster-affected areas tend to be relatively large. Hence we deploy the drone-cells to try to cover a maximum number of users, and some of the isolated users may remain uncovered.

According to [140], the optimal altitude and the corresponding coverage radius of the drone-cell can be numerically solved for a certain environment (urban, suburban) and a given path loss threshold. Therefore, we deploy all the drone-cells at the same altitude in one horizontal plane. Hence, we consider the drone-cells' coordinates at that plane as coordinates on the surface. The coverage of the drone-cell is disk-like with the drone-cell' projection on the ground as the center and a coverage radius of $R_d$. Let the set $G = \{G_1, G_2, \cdots, G_k\}$ denotes the coordinates of the GBSs' locations, $D = \{D_1, D_2, \cdots, D_m\}$ denotes the coordinates of the drone-cells' deployment locations , where $G, D \in \mathbb{R}^2$. Considering the limitation of the communication range of such backhaul links, we require that the drone-cells and GBSs, which should be able to communicate by direct backhaul link, are within certain ranges ensure the backhaul links are reliable. We will consider the deployment of drone-cells at any time satisfying the following constraints:

$$|D_i, G_h| \leq r_1 \qquad (4.1)$$

If drone-cell $D_i$ and the corresponding GBS $G_h$ are connected by a direct backhaul

link; and

$$|D_i, D_h| \leq r_2 \tag{4.2}$$

If drone-cells $D_i, D_h$ are connected by a direct backhaul link. Here $|,|$ denotes the standard Euclidean distance between two points. $r_1, r_2$ are some given constants describing the communication ranges between drone-cells and a GBS and another drone-cell, beyond which the wireless backhaul links may be unreliable. Practically, $r_1, r_2$ should always hold that:

$$r_1 \geq R_g + R_d \tag{4.3}$$

$$r_2 \geq 2R_d \tag{4.4}$$

Where $R_g$ is the coverage radius of the GB. The values of $r_1, r_2, R_g, R_d$ are influenced by actual signal effects in the actual environment.

To obtain heterogeneity in spatial user distribution, we adopt a doubly Poisson cluster process: *matérn* cluster process [140]. Where the centers of user clusters are created by a homogeneous Poisson process. The users within each cluster are uniformly scattered in circles with radius $R_c$ around parent points by using another homogeneous spatial Poisson process [26]. The density function of clustered users in location $z$ is

$$f(z) = \begin{cases} \frac{1}{\pi R_c^2}, & \text{if } \|z\| \leq R_c \\ 0, & \text{otherwise} \end{cases} \tag{4.5}$$

Suppose there are uncovered users labelled $i = 1, 2, \ldots, n$. Let the set $U = \{U_1, U_2, \cdots, U_n\}$ denotes the coordinates of the locations of all uncovered uses within the operating area, exclude the users that already been covered by any of the GBSs through direct link. Our target is to find the optimal 2D deployment $D = \{D_1, D_2, \cdots, D_m\}$ so that the number of served users is maximized. Our deployment optimization problem is formulated as follows:

$$\max_{D, \{I_j\}} \sum_{j=1}^n W_j \tag{4.6}$$

Subject to:

$$R_g + R_d \leq |D_i, G_h| \leq r_1 \tag{4.7}$$

$$2R_d \leq |D_i, D_h| \leq r_2 \tag{4.8}$$

$$W_j = \begin{cases} 1, & \text{if } |D_i, U_j| \leq R_d, \forall i, j \\ 0, & \text{otherwise} \end{cases} \tag{4.9}$$

Where (4.7),(4.8) indicate that: besides the backhaul link constraints, the coverage area of a drone-cell is not allowed to overlap with the coverage area of any GBS or another drone-cell. Moreover, $W_j = 1$ if user $j$ is covered by any drone-cell and $W_j = 0$ otherwise. As one of the important features of a drone-cell is its fast mobility, drone-cells can change their deployment rapidly to follow the movement of users if needed. Moreover, for battery-life saving purpose, if a drone-cell flies to a suitable location, it can stay there for a while until the network reaches a particular pre-determined threshold of user-dropped out. Therefore, in this chapter, we find the deployment of the drone-cells for only one snapshot of the users' positions. The users' positions are assumed to be apriori, the methods of acquiring such positions information can be seen in [91, 93].

## 4.3 Backhaul-aware Deployment Algorithms

Generally, the optimal deployment of drone-cells is a NP-hard problem [92]. Adding a new dimension to the problem, which is the backhaul link constraints of the drone-cells, makes the problem even more complicated. To find the backhaul-aware deployment of a group of drone-cells ensuring that a maximum number of users can be served, we first propose an exhaustive search algorithm that solves the problem using brute force. After that, we introduce a much more efficient solution an algorithm based on GA with restart-strategy. We then compared the performance of the two algorithms through a bunch of simulations.

### 4.3.1 Exhaustive Search Algorithm

Before describing the algorithm, we first investigate the optimal deployment problem for a single drone-cell. Which is very similar to a classic circle placement problem: given a set of $N$ points $P = \{p_0, p_1, \ldots, p_N\}$, in a 2-D plane, and a fixed disk of radius $R$, find a location to place the disk such that the total number of the points covered by the disk is maximized.

One of the most efficient solutions for the circle placement problem is an $O(N^2)$ greedy algorithm based on finding the maximum *Clique* (graph theory) in an intersection graph. This is the approach developed by [141] in 1984. The key observation is that finding the maximum number of points covered by a single disk with fixed radius $R$ is the same as finding a single point covered by the maximum number of disks with radius $R$. Moreover, if a point is covered by $k$ disks, then those disks must have their mutual distances all less than $2R$. Therefore, an intersection graph consists of $N$ vertices can be created. Each of the vertices is corresponding to one disk, and two vertices are connected by an edge if the centers of those disks are less than $2R$ distance apart. If a single point is covered by $k$ disks, then these disks' corresponding vertices are all connected to each other by edges and therefore form a *Clique*. Reference [141] then proposed a greedy algorithm to find the largest *Clique*, which solved the circle placement problem accordingly, more details can be found in [141].

When solving the single drone-cell's optimal deployment problem by the solution from [141], it must be noted that (4.7) (4.8) as the deployment constraints should always be satisfied. Which means the drone-cell can only be deployed in a bounded closed region $\mathcal{U}$ decided by the locations of the GBSs and the other drone-cells that have got their locations fixed. We now make use of the notion of *deployment region*.

**Definition 4.3.1.** Deployment region: A deployment region is the limited bounded closed region that a drone-cell can to be deployed with the locations of GBSs and the other drone-cells are fixed, and constraints (4.7) (4.8) are satisfied.

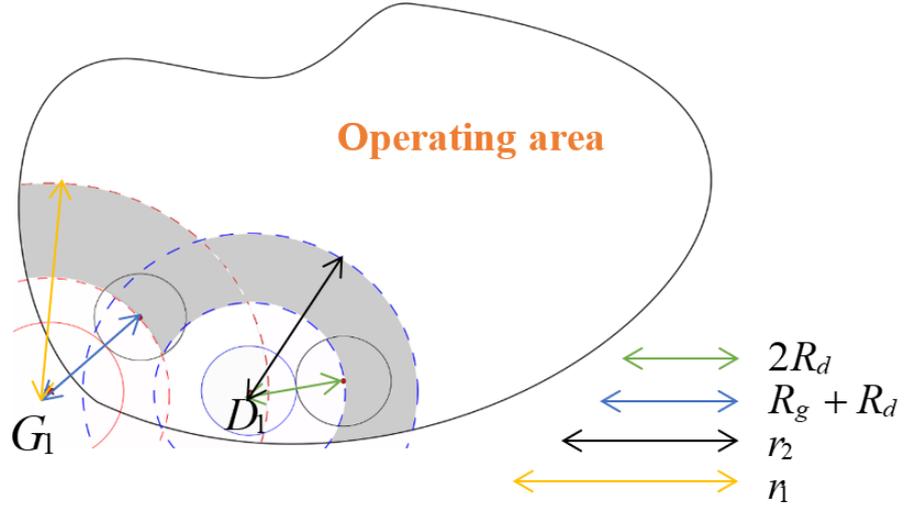

Figure 4.2: The deployment region $\mathcal{U}$ (grey area) of a drone-cell decided by a GBS $G_1$ and a deployed drone-cell $D_1$.

An example of the deployment regions $\mathcal{U}$ of a drone-cell is shown in Figure 4.2. We now describe the exhaustive search algorithm in its entirety in Algorithm 2. The variables with the subscript like $(j1, ..., jt)$ mean these variables are decided by the locations of the first drone-cell $D_1$ to the $t$th drone-cell $D_t$. To explain the proposed exhaustive search algorithm in more detail. We take the example of deploying a group of $m = 3$ drone-cells and introduce applying the exhaustive search algorithm to find their optimal deployment. We start with finding the deployment region $\mathcal{U}^1$ of the first drone-cell $D_1$. Let set $U^1 = \{U_1^1, U_2^1, \cdots, U_{N1}^1\}$ denotes the locations of all the $N1$ users within $\mathcal{U}^1$. Visit each $U_{j1}^1 \in \mathbb{U}^1$, where $j1 \in 1, 2, \cdots, N1$. Let $D_1 = U_{j1}^1$, find the number of users $n_{j1}$ that can be covered by $D_1$ and the corresponding deployment regions $\mathcal{U}_{j1}^2$ for the second drone-cell $D_2$. $n_{j1}$ is simply the number of users $U_{j1}^1 \in \mathbb{U}^1$ satisfy $\left| U_{j1}^1, D_1 \right| \leq R_d$. Let set $\mathcal{U}^2 = \{\mathcal{U}_1^2, \mathcal{U}_2^2, \cdots, \mathcal{U}_{N1}^2\}$ denotes all the candidate deployment regions of $D_2$. For each $\mathcal{U}_{j1}^2 \in U^2$, let set $U_{j1}^2$ denotes the locations of all the users within $\mathcal{U}_{j1}^2$. Visit each $U_{j1,j2}^2 \in \mathbb{U}_{j1}^2$, let $D_2 = U_{j1,j2}^2$, find the number of users $n_{j1,j2}$ that can be covered by $D_2$ and the corresponding deployment regions $\mathcal{U}_{j1,j2,j3}^3$ for the third drone-cell $D_3$. Then, apply the solution from [141] to find the maximum number of users $n_{j1,j2,j3}$ can be covered by $D_3$ and its corresponding optimal deployment $U_{j1,j2,j3}^3$. Finally, compare and find the combination $\{j1^*, j2^*, j3^*\}$ that maximizes $(n_{j1} + n_{j1,j2} + n_{j1,j2,j3})$ and the corresponding optimal deployment $[D_1, D_2, D_3] = [U^1_{j1*}, U^2_{j1*,j2*}, U^3_{j1*,j2*,j3*}]$. Note that the solution generated by proposed Algorithm 2 is only an approximation of the global optimum of the considered deployment problem, i.e., quasi-optimal. This is because the deployments of the first $(m - 1)$ drone-cells are limited to some locations of the users, not any points on the candidate deployment regions.

**Lemma 1.** The worst-case running time of the exhaustive search algorithm is $O(n^{2m})$.

---

**Algorithm 2** Exhaustive Search Algorithm for Optimal Backhaul-aware Deployment of Drone-cells.

---

**Input:**

$G = \{G_1, G_2, \cdots, G_k\}, U = \{U_1, U_2, \cdots, U_n\}, m,$
$R_d, R_g$

**Output:**

$D = \{D_1, D_2, \cdots, D_m\}$

1: $t = 1$;
2: **while** $t < m - 1$ **do**
3: Find the set $U^t \in \mathbb{U}$ contains the coordinates of users within all the candidate deployment regions of the $t$th drone-cell $D_t$;
4: Visit all users in $U^t$, for each of the user $U^t_{j1,\ldots,jt} \in \mathbb{U}^t$, let $D_t = U^t_{j1,\ldots,jt}$, calculate the number of users $n_{j1,\ldots,jt}$ that $D_t$ can cover, record $n_{j1,\ldots,jt}$ and its corresponding $U^t_{j1,\ldots,jt}$ as the candidate of $D_t$. Then, find the set $U^{t+1}$ contains the coordinates of users within all the candidate deployment regions of the $(t+1)$th drone-cell $D_{t+1}$;
5: Visit all users in $U^{t+1}$, for each of the user $U^{t+1}_{j1,\ldots,j(t+1)} \in \mathbb{U}^{t+1}$, let $D_{t+1} = U^{t+1}_{j1,\ldots,j(t+1)}$, calculate the number of users $n_{j1,\ldots,j(t+1)}$ that $D_{t+1}$ can cover and its corresponding $U^{t+1}_{j1,\ldots,j(t+1)}$ as the candidate of $D_{t+1}$;
6: $t = t + 1$;
7: **end while**
8: Find all the candidate deployment regions $\mathcal{U}_m$ for the $m$th drone-cell $D_m$. For each candidate deployment region $\mathcal{U}^m_{j1,\ldots,jm} \in U_m$ , apply the solution from [141] to find the maximum number of users $n_{j1,\ldots,jm}$ can be covered by $D_m$ and its corresponding optimal deployment $U^m_{j1,\ldots,jm}$;
9: Compare and find the combination $\{j1*, j2*, \ldots, jm*\}$ that gives $\max(n_{j1} + n_{j1,j2} + \ldots + n_{j1,\ldots,jm})$, and the corresponding optimal deployment candidates set $U* = \{U^1_{j1*}, U^2_{j1*,j2*}, \ldots, U^m_{j1*,\ldots,jm*}\}$;
10: **return** $D = U*$; =0

---

## 4.3.2 Genetic Algorithm

Metaheuristic algorithms such as GA, simulated annealing, and particle swarm optimization are often used in solving NP-Hard problems. Here we use GA to find the optimal backhaul-aware deployment of the drone-cells. The GA is a heuristic search algorithm developed by John Holland in the 1960s and first published in 1975 [142]. Inspired by Darwin's theory of evolution, GA has long been a widely used non-deterministic optimization method. Specifically, GA simulates the evolution of a population of chromosomes to optimize a problem. Similarly to organisms adapting to their environment over the generations, the problems to be solved by GA are simulated as a process of biological evolution. The chromosomes in GA adapt to a fitness function over an iterative process. The next generation of chromosomes is generated by using biology-like operators such as the crossovers, the mutations, and the inversions. The chromosomes with low fitness function value will be eliminated gradually, while the percentage of those with high fitness function value is increased. In this way, after a certain number of iterations, it is possible to evolve solutions with high fitness function values.

In this chapter, we use GA to simulate the evolution of a population of drone-cells' deployment adapting to the fitness function defined simply by the total number of users covered by the drone-cells. In detail, the GA begins by encoding and creating a random initial population. Since the coordinates of the deployed drone-cells are real numbers, direct value encoding is used for our problem. Each chromosome is a 2D array with a size of $m$ by 2, which stores the 2D coordinates of the group of $m$ drone-cells. The initial Let $\mathcal{P}_{size}$ denotes the number of chromosomes in the population, i.e., the population size. The initial population is generated randomly, with each drone-cell coordinates in the chromosomes bounded in the operating area. Then we calculate the fitness function $f(I)$ for each of the chromosomes $I$. Where $f(I)$ equals the number of users covered by the drone-cells with their coordinates satisfy the backhaul constraints (4.7) (4.8). Afterward, we start the iterations of evaluation. Let $iter\_max$ denotes the maximum number of iterations. In each of the next generations, the children chromosomes from the parent ones are produced following three steps:

### 4.3.2.1 Mutation

In this step, the following adaptive mutation operator is applied with a mutation probability $P_m$ to some genes $g$ (coordinates of a drone-cell in the X-axis or Y-axis) of the chromosomes.

$$g' = g \left[ 1 \pm rnd \left( 1 - \frac{f}{f_{\text{best}}} \right)^2 \right] \tag{4.10}$$

Where $rnd \in [0,1]$ is a random value, $f$ is the fitness of the current chromosome, $f_{\text{best}}$ is the current highest fitness achieved by the population. It is noticeable that the influence of the mutation operator becomes tiny if $f$ is close to $f_{\text{best}}$, this is used to limit adverse moves and protect the best chromosome of every population. Moreover, to improve the efficiency of the GA and the consistency of the final results. We adopt Simulated-Annealing Mutation (SAM) [143] operator that uses the Simulated-Annealing stochastic acceptance function internally to limit adverse moves. Specifically, we calculate the difference between the fitness before and after the mutation $\delta_f$. If the fitness after mutation decreases, the annealing probability $\exp(\delta_f/T)$ is given to determine whether to accept the mutation or not. Where $T$ is an adjustable constant. Let $T = \alpha T$ after each iteration, where $\alpha \in [0,1]$ is a constant to let T gradually decreases after each iteration.

### 4.3.2.2 Crossover

To protect the high-fitness chromosomes and improve the speed for the GA converges to the global optimum, we adopt the following adaptive crossover operator proposed in [144] with a crossover probabilities $P_{c1}$ and $P_{c2}$.

$$\left\{ \begin{array}{l} [x_n, y_n] \\ [x_m, y_m] \end{array} \right\} \xrightarrow{crossover} \left\{ \begin{array}{l} [x_n, y_m] \\ [x_m, y_n] \end{array} \right\} \tag{4.11}$$

$$P_c = \begin{cases} P_{c1} - \frac{(P_{c1}-P_{c2})(f'-f_{avg})}{f_{max}-f_{avg}}, & f' \geq f_{avg} \\ P_{c1}, & f' < f_{avg} \end{cases} \quad (4.12)$$

Where $f_{\text{avg}}$ is the current average fitness of the population, $f'$ is the larger fitness of two chromosomes selected to perform the crossover.

### 4.3.2.3 Select

In this step, we first calculate the fitness function $f(I)$ for each of the chromosomes $I$ again. Find the worst 20% chromosomes in terms of fitness. Then replace them with the chromosome that has the highest fitness.

Once reach the maximum number of iterations *iter_max*, return the chromosome $I^*$ that has the highest fitness in the population as the final solution $D = I^*$ for the backhaul-aware deployment of drone-cells. Currently, there is no theory on algorithm parameters that applies to all GA applications. The parameters selected by the GA should be based on the specific problem being solved. After a bunch of trails and tests, we find that the combination of the algorithm parameters in Table 4.1 can deliver the best performance in terms of probability of finding the global optimum by the GA.

Table 4.1: GA Parameter Values

| Parameters | Values |
|:---:|:---:|
| $T$ | 500 |
| $\alpha$ | 0.99 |
| $P_m$ | 0.2 |
| $P_{c1}$ | 0.4 |
| $P_{c2}$ | 0.2 |

However, the backhaul-aware deployment of drone-cells could be a multi-peak problem with lots of local optima. As shown in Figure 4.3(a), an example shows that the probability of finding the global optimum by the GA runs 50 times is only 12%, for the backhaul-aware deployment problem of two drone-cells with 1 GBS, $iter\_max = 2000$ and the population size $\mathcal{P}_{size} = 100$. Figure 4.3(a) also shows that the deployment problem is a multi-peak problem with many local optima. To this situation, we now introduce the 'restart strategy', which is very simple but surprisingly effective for improving the probability of finding the global optimum by the GA to solve the backhaul-aware deployment problem, without increasing the computation complexity.

**Restart strategy**: instead of running the GA with a large number of iterations, we now run the GA with a much smaller number of iterations, but restart the GA for a number of times $N_{start}$. For instance, instead of running the GA with 2000 iterations for the example in Figure 4.3(a), we now let $iter\_max = 50$, and restart the GA for 40 times, while all the other parameters such as $\mathcal{P}_{size}$ remain same. Note that, the total number of iterations calculated is $50 \times 40 = 2000$. Thus the total computation complexity remains unchanged.

Reference [145] indicates that a restart strategy is a very economical approach for hard computational problems. Expressly, the restart strategy acknowledges that we do not have a particularly effective solution to multi-peak problems. Compare with other complex jump-out strategies, a simple restart strategy may be a more practical solution for avoiding local optima, especially when the problem is difficult, and the success probability for finding the global optimum is minimal. Another advantage of the restart strategy is that it does not need to make any changes to the algorithm and reinitialize it. As shown in Figure 4.3(b), the probability of finding the global optimum has been increased to 92% by restarting the GA with $iter\_max = 50$ for 40 times, for solving the same deployment problem in Figure 4.3(a). In the following part of this chapter, all the GA we mentioned are GA with restart strategy. Notably, applying the proposed GA to solve the backhaul-aware deployment problem for more drone-cells requires a larger $\mathcal{P}_{size}$ and/or $N_{start}$, to guarantee finding the global optimum.

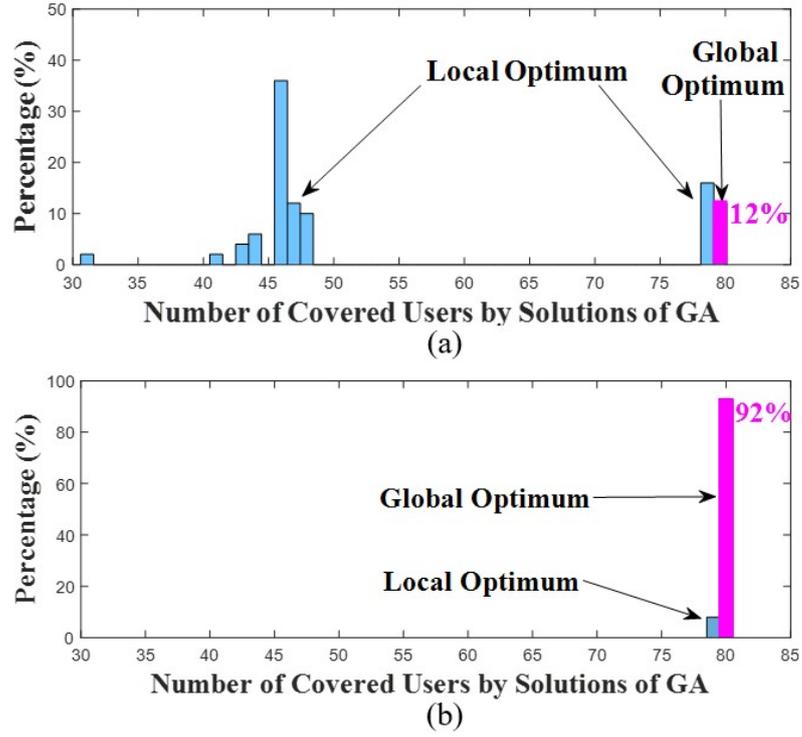

Figure 4.3: Distribution and percentage of global optimum versus local optima finds by the GA without restart strategy (a) and with restart strategy (b), each runs for 50 times.

## 4.4 Simulation Results

In this section, the performance of the proposed exhaustive search algorithm and the GA with the restart strategy is evaluated using Matlab. We simulated the deployment of a team of drone-cells in a 10km × 10km quadrangle area with only one operational GBS with its coordinates $G$. The locations of the users are generated by the fore-mentioned doubly Poisson *matérn* cluster process, with the clustered user density of $\lambda_c$ and the non-clustered user density of $\lambda_{nc}$. The number of user clusters is $N_c$. For simplicity, all user clusters are set to be disk-like, with random radii no larger than $R_c$. The coverage radius of the GBS and the drone-cell are $R_g$ and $R_d$, respectively. The limited backhaul link distances between a GBS and a drone-cells and between two drone-cells are $r_1$ and $r_2$, respectively. The parameters of the GA are set the same as Table 4.1. The detailed simulation parameters are as shown in Table 4.2.

Table 4.2: Simulation Parameter Values

| Parameters | Values | Parameters | Values |
|---|---|---|---|
| $G$ | (0,0) | $\lambda_c$ | 9e-5 users/$m^2$ |
| $\lambda_{nc}$ | 3e-6 users/$m^2$ | $N_c$ | 15 |
| $R_c$ | 250 m | $R_d$ | 500 m |
| $R_g$ | 1000 m | $r_1$ | 3500 m |
| $r_2$ | 3000 m | | |

We first apply our proposed exhaustive search (ES) algorithm and the GA to find the optimal backhaul-aware deployment in a 10km × 10km quadrangle area with the same user distribution for 1,2,3 and 4 drone-cells, respectively. The computing times for both algorithms are compared in Table 4.3. Where $m$ is the number of drone-cells, $T_{ES}$ is the computing time of ES, $T_{GA}$ is the computing time of GA, $P_{save}$ is the percentage of the computing time saved by GA, $\mathcal{P}_{size}$ is the population size of GA, $N_{start}$ is the number of restart of GA, $P_{Global}$ is the success rate for the GA to find the global optimal deployment over 20 runs. $iter\_max$=50 for all the examples. The simulations were run by MATLAB 2019b installed on an HP desk computer with the 7th generation Intel Core i7 processor and 16GB RAM.

As shown in Table 4.3, the computing time of the ES increased dramatically from 0.78 seconds to 276970 seconds, with the number of drone-cells increases from 1 to 4. Meanwhile, the computing time of the GA increased with much less speed. Consequently, the percentage of computing time saved by GA rises significantly. In particular, the proposed GA can save the computing time up to 99.927%, with the success rate of $P_{Global} = 95\%$ over 20 runs for the example of 4 drone-cells. Moreover, the trend of $P_{save}$ in Table 4.3 suggests that a higher percentage of the computing time saved by GA could be achieved when solving the deployment problem with more drone-cells. These results verified the computational efficiency of the proposed GA. The corresponding optimal backhaul-aware deployment of the four examples finds by the GA are as shown in Figure 4.4, where the black circles are the coverage of the drone-cells. The GBS is located at the origin. The red dash circles denote the limited backhaul link range. Specifically, A drone can only have backhaul links if it is within the red dash circle of the GBS or another drone-cell. As indicated in Figure 4.4, some of the drone-cells have to be placed at the edge of the red dash circles to cover a maximum number of users.

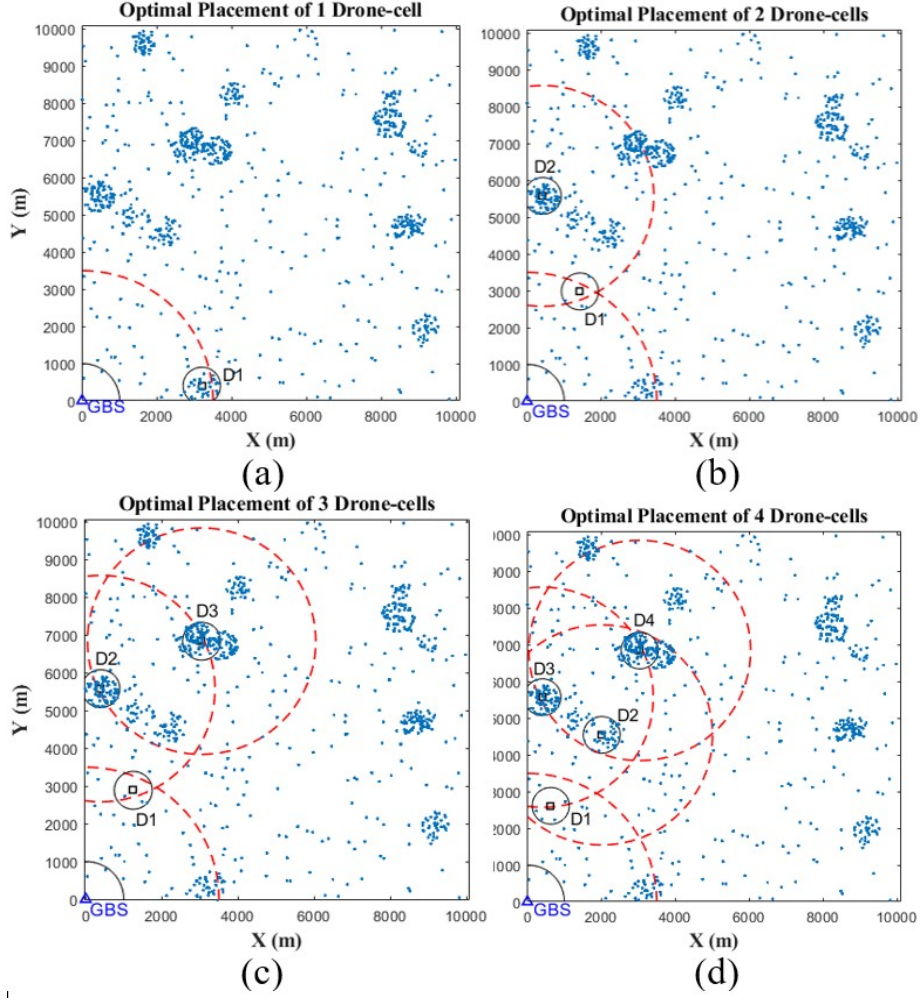

Figure 4.4: Optimal backhaul-aware deployment finds by the GA for 1 drone-cell (a), 2 drone-cells (b), 3 drone-cells (c), and 4 drone-cells (d).

Figure 4.5 indicates the success rate $P_{Global}$ for the GA to find the global optimal deployment of 3 drone-cells with different population size $\mathcal{P}_{size}$ and number of restart $N_{start}$. $iter\_max = 50$ for all the examples. According to Figure 4.5, for different values of $N_{start}$, $P_{Global}$ does not increase with the population size $\mathcal{P}_{size}$ increases. However, when $N_{start}$ enhances, $P_{Global}$ does increase accordingly, for different $\mathcal{P}_{size}$. Therefore, in terms of improving $P_{Global}$, increasing $N_{start}$ would be a more efficient choice than increasing $\mathcal{P}_{size}$.

Table 4.3: Computing Time Comparison

| $m$ | $T_{ES}$ (s) | $T_{GA}$ (s) | $P_{save}$ | $\mathcal{P}_{size}$ | $N_{start}$ | $P_{Global}$ |
|---|---|---|---|---|---|---|
| 1 | 0.78 | 0.82 | -5.13% | 20 | 30 | 100% |
| 2 | 15.4 | 8.1 | 47.4% | 50 | 50 | 100% |
| 3 | 1449 | 23.3 | 98.39% | 100 | 60 | 100% |
| 4 | 276970 | 202 | 99.927% | 200 | 200 | 95% |

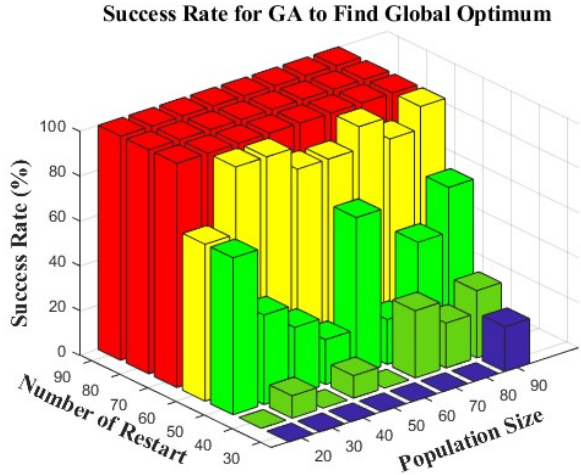

Figure 4.5: The success rate for finding the global optimum by the GA with different population size $\mathcal{P}_{size}$ and number of restart $N_{start}$.
.

Finally, we investigate the influence of different user distribution models to the optimal backhaul-aware deployment results. Specifically, by changing the clustered user density of $\lambda_c$ and the non-clustered user density of $\lambda_{nc}$, we simulated two user distribution scenarios: clustered and non-clustered. The total number of users and all the other parameters are set to be the same for both scenarios. Then we run the ES and GA to find the optimal backhaul-aware deployment for a different number of drone-cells. We then compared the total number of users that can be covered $N_{total}$ under the optimal deployment, as shown in Figure 4.6. It can be seen from Figure 4.6 that a certain number of drone-cells can always cover significantly more users for the scenario with clustered user distribution than the one with non-clustered user distribution. Moreover, the increasing speed of the $N_{total}$ decreased when adding the number of drone-cells $m$ for the clustered scenario, while $N_{total}$ increased with a constant speed as $m$ increases for the non-clustered scenario. Since we have verified the high efficiency and accuracy for the proposed ES and GA to find the optimal deployment. The slow down of $N_{total}$'s increasing speed could be contributed to that the user clusters with more users were effectively identified by the proposed algorithms, so that they have the priority to be covered by the first several drone-cells. The user clusters covered lately tend to have fewer users. Consequently, the $N_{total}$'s increasing speed reduced when adding more drone-cells to the network. For non-clustered scenario, the effectiveness of proposed algorithms tend to be less significant, because $N_{total}$ is mainly decided by $m$, not the optimal deployment, as long as constraints (4.7) (4.8) are satisfied. Thus, our proposed algorithms are more suitable for solving the drone-cell's optimal backhaul-aware deployment problem for the scenario with clustered users.

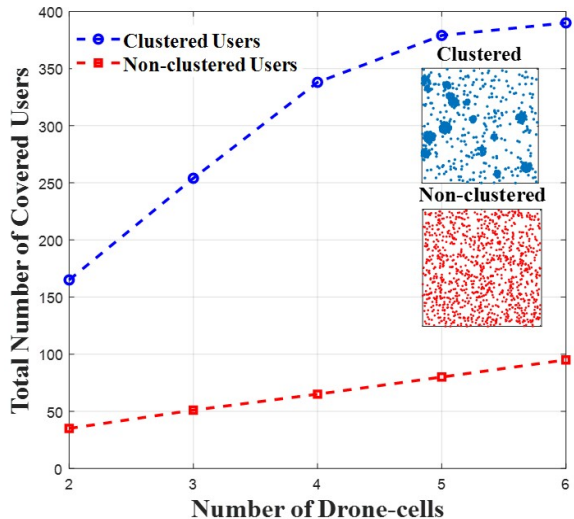

Figure 4.6: The maximum number of users can be covered with a different number of drone-cells, for clustered and non-clustered user distribution.
.

## 4.5 Conclusion

In this chapter, we considered the optimal deployment problem of a team of drone-cells with limited backhaul communication distance, aiming at maximizing the total number of users covered by the drone-cells. We proposed two approaches, exhaustive search algorithm and a computationally efficient GA to find the optimal 2D backhaul-aware deployment of multiple drone-cells. We also introduced a restart-strategy that helps proposed GA to avoid local optima. Simulations show that the proposed GA can significantly save the computing time compare with the exhaustive search algorithm and the restart-strategy is verified to be a simple but very effective technique that significantly increases the success rate for the GA to find the global optimum. The proposed method relies on accurate location information of users. One future research direction is to extend the current method to the case where more precise location estimation can be acquired by some estimation tools such as robust Kalman filter [146–149]. The applications of the proposed method is not limited to drone-cells, another future research direction is to apply the current method to drones for surveillance purposes [7, 47, 108].

# Chapter 5

# A Novel Method for Protecting Swimmers and Surfers from Shark Attacks using Communicating Autonomous Drones

Shark attacks can make beach tourists anxious about sharing the ocean with apex predators. Although the raw number of shark attacks is deficient, the absolute terror caused by sharks is genuine. This chapter introduces a novel method named as 'drone shark shield system', which uses communicating autonomous drones to intervene and prevent shark attacks for protecting swimmers and surfers. We detail the design of the drone shark shield system and the strategy for repelling sharks through multiple intersections. A shark interception algorithm is developed to guide drones to predicted intersection points for deterring sharks. Computer simulations are conducted to illustrate our method.

## 5.1 Motivation

Despite its rare occurrence, shark attacks have long been a problem threatening
beach visitors such as swimmers and surfers. According to Shark Research Institute,
126 fatal shark attacks were reported globally from 2000 to 2013 [28]. Though this
is not a horrifying number, as soon as a fatal shark attack occurs, it will continue
to scare off residents and tourists from the entire area, causing a devastating blow
to the local tourism industry.

To reduce shark incidents, shark hazard-mitigation strategies have been adopted
by the United States, South Africa, Australia, Brazil and Reunion Island, which are
regarded as shark bite "hotspots" [150]. Such strategies generally rely on culling pro-
grams aiming at reducing the local abundance of hazardous species, which have typi-
cally been implemented by deploying shark nets, longlines and drumlines [151–153].
The shark nets, longlines or drumlines, however, result in not only the incidence
of bycatch, including threatened and endangered species like dolphins, sea turtles,
whales, and dugongs, but also the death of a huge number of sharks. According to
the non-profit organization (NGO) "The Shark Angels" [154], shark nets have been
responsible for the death of over 33,000 sharks in the last thirty years, with 25,000
being harmless to humans. In addition, the bycatches of the shark nets during the
same period were 2,211 turtles, 2,310 dolphins, and 8,448 rays. The death of these
marine animals has had significant impacts on the health of the aquatic ecosystem.
Nevertheless, shark nets do not always work. Many cases of shark attacks happened
in the beach area having shark nets [155], and some video footages show sharks are
close to the swimmers or surfers [156, 157].

Once sharks were spotted near the beach area, the lifeguards will go out on Jet
Skis to investigate and drive the shark away. At the same time, the entire beach will
be closed. This method could prevent some potential shark attacks. However, the
effectiveness depends on the timeliness of spotting sharks. Swimmers or surfers near
the shark may still be attacked before lifeguards arrive. The high-speed rotating
propeller on the Jet Skis may hurt the shark while driving it away. Moreover, shut-ting the entire beach down will damage the tourist economy, especially for famous tourist attractions.

When sharks are approaching the beach area, how to spot them is always a problem, and the earliest spotting is crucial in saving human lives. For this purpose, unmanned aerial vehicles, also known as drones, that have been widely applied in a variety of surveillance [3, 4, 8, 158] and target tracking applications [159], bring with a promising solution, thanks to the agility, easy-to-deploy, and low cost. For example, the Westpac Little Ripper Lifesaver drone fitted with a SharkSpotter artificial intelligence (AI) algorithm [160] to detect sharks from live video feeds, can efficiently differentiate sharks from 16 other species of marine animals [161]. With 90 percent accuracy at shark detecting, this system distinguishes between marine life such as whales, dolphins, and rays, and can identify human beings such as swimmers, surfers, boats, and other objects on the sea surface.

Beyond shark detection, repelling sharks is more important to prevent attacks. One possible solution is to use a group of communicating autonomous drones, with some of them being able to detect sharks [160] and others being able to repel sharks with attached electric shark repellent. Specifically, sharks have poor eyesight but have a special sensory system. In addition to the hearing, vision, touch, smell, and taste shared with humans, sharks also have a highly sensitive electric receptor called the *ampullae of Lorenzini*, which is found throughout the shark's snout and head that can feel the weak electrical signal from the movement of muscles of the prey. Sharks use electric receptors to sense preys at very close distances, typically less than one meter. In detail, sharks use senses such as audition and olfaction as the primary drivers track animate objects over long distances, and use *ampullae of Lorenzini* as short-range sensors when feeding or searching for food. Particularly, electromagnetic pulses produced by prey animals will not attract sharks if they are not close enough [162, 163].

Considering this feature, an electric shark repellent technology was proposed in South Africa in the 1990s, based on the creation of an electric field that overwhelms the shark's highly sensitive electrical receptors and forces the shark to retreat [164].
The electric shark repellent uses two or more electrodes to generate an artificial
electric field that exceeds the sensory load of the shark's highly sensitive electrical
system and makes the shark feel uncomfortable. If the shark is over-stimulated by
artificial electric fields, it will experience an unbearable embarrassment and take
evasive actions. The electrodes must be immersed in seawater so that the seawater
can be used as the conductor to generate a three-dimensional (3D) electric field
for repelling the shark. As a substantially non-invasive, non-lethal shark deterring
technology, electric shark repellent has got its effectiveness proved by independent
scientific studies [165, 166]. The reference [167] presented the quantitative evidence
of the effectiveness of the electric shark deterrent. Specifically, the article shows
that sharks would not cross an electric field when field gradients reached 7–10 V/m.
Moreover, real field tests in [167] suggest that the strength of the electric field for
sharks' first encounter is about 9.7 V/m, and the average proximity of all encounters
is 15.7 V/m.

The purpose of this chapter is to introduce a novel shark defence system that
uses communicating autonomous drones to intervene and prevent shark attacks for
protecting swimmers and surfers, and we name the system as 'drone shark shield'.
We also propose the shark repelling strategy that can eventually drive the shark
to leave the beach area by multiple interceptions. Furthermore, we propose an
efficient interception algorithm to guide drones to predicted intersection points for
deterring sharks. The main contribution of this chapter is that we propose the
first intelligent system that can proactively prevent shark attacks while protecting
sharks and other marine creatures. We detail not only the design of the proposed
system but also its working mechanism and the control algorithm. The proposed
system is promising to make it more comfortable for people to share the ocean with
sharks. In addition to protecting swimmers and surfers, the proposed system can
also save the life of sharks and other marine creatures from current shark defence
measures. Thus, the system can promote the harmonious coexistence of humans,
sharks, and other marine animals. Moreover, during the execution of surveillance and shark deterring missions, drones in the proposed system can form a wireless mobile sensors network [168, 169], which can continuously collect data such as the population and activities of sharks and other marine creatures. The collected data can be used by marine biologists for further study and protection of marine life.

The remainder of this chapter is structured as follows. Section 5.2 introduces the design of the drone shark shield system. Section 5.3 presents the strategies for deterring the shark and the interception algorithm. In Section 5.4, we demonstrate the theoretical effectiveness of the drone shark shield by computer simulations. Finally, Section 5.5 concludes this chapter.

## 5.2 Design of the Drone Shark Shield

We now detail the design of the proposed drone shark shield system. It consists of a fleet of two types of drones to continuously patrol beach areas [170]. Similar to the Westpac Little Ripper Lifesaver drone mentioned in Section 5.1, the first type is shark-detecting drones that are equipped with cameras and fitted with an AI algorithm that can detect sharks from live video feeds with high accuracy [160]. From now on, we call this type of drone the 'observer'. The observer can also identify human beings such as swimmers and surfers. Moreover, the observers are fitted with speakers for warning swimmers and surfers when detecting any shark.

A drone of the second type is attached with an electric shark repellent and a miniature sonar, and we call it the 'operator'. There are two reasons for using drones as operators, not autonomous underwater vehicles (AUVs). The first one is that drones are more maneuverable and faster than AUVs. The second is that during the execution of surveillance and shark deterring missions, the obstacles that drones need to avoid tend to be fewer than AUVs, which need to avoid reefs, sharks, and other marine life all the time. The observer and operator drones should have the embedded ability to avoid collisions with any obstacles e.g. kite-surfers and between each other, during the entire mission [171, 172]. The considered system of communicating autonomous drones can be viewed as an example of networked control system [173–184].

The electric shark repellent fitted on the operator is designed to be an inflatable tube with several electrodes attached to it. In this chapter, we consider the case with three electrodes [1]. A miniature sonar is attached to the tail of the inflatable tube. The sonar is to detect the positions and velocities of sharks and humans in the water. The material of the inflatable tube should be carefully chosen so that the tube can be coiled and stored in a relatively small box when it is not inflated. Furthermore, the inflatable tube should be hard enough to act like a stick attached to the operator after it been inflated, so that it will not act as a pendulum when the operator is moving. Carbon fiber fabric can be a possible solution for the material of the inflatable tubes, because it has strong tensile resistance [185]. The inflation process should be very fast and can be completed in seconds. For example, the implementation of the inflation process can be realized by puncturing a small carbon dioxide cylinder with a steel needle. The carbon dioxide cylinder should be replaceable and the inflation process should able be triggered automatically.

The observers and operators have the communication ability. The communication between them can be realized by 2.4 gigahertz radio waves, which is commonly used by different drone products. The communication between the observer and the operator is mainly unidirectional from the observer to the operator. Specifically, the observer monitors the horizontal locations of sharks and humans and sends them to the operator in the real-time. During the surveillance, once sharks are detected by any observer, the operators will be immediately sent to repel the sharks, with the observer staying at a certain altitude to identify sharks and humans, and keep sending the information to the operators. The information will then be combined at the operators with the signals provided by the sonar to generate the real-time positions and velocities of the sharks and humans. We assume that each operator can only repel one shark at the same time, but one observer can provide the detected information to multiple operators. Figure 5.1 shows a schematic diagram of a basic unit of the drone shark shield system, with one operator and one observer. Any drone in the system should have a switchable battery and can autonomously fly back to the ground drone base station to replace the battery or recharge it with automatic charging devices [186]. The system should also consist of backup operator and observer drones that can be deployed alternately to achieve persistent operation. Besides, the advancement of solar-harvesting technology enables the drones to prolong the battery lifetime [187]. The predetermined drone base station should be located near the house of the lifeguards on the beach for the drones to take off and land. The lifeguards should be able to maintain the drone shark shield system after training.

---

[1]Other cases with different number of electrodes can be analysed similarly.

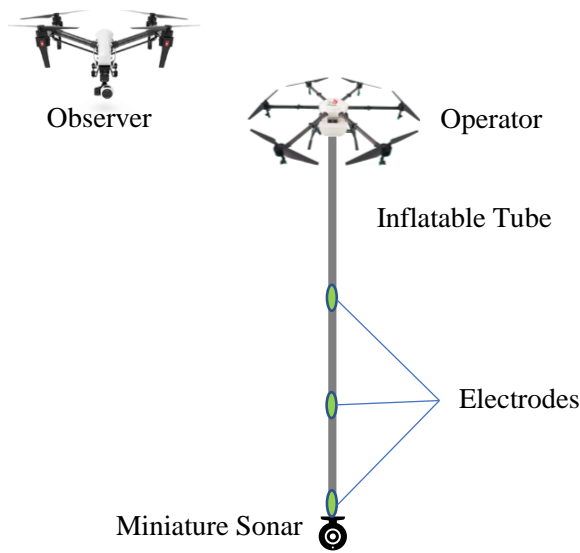

Figure 5.1: Design of a basic unit of the drone shark shield system.

The electrodes trailed by the operator generate a 3D electric field for deterring the shark once they are immersed in the seawater, as shown in Figure 5.1. The generated electric field is set to be vertically symmetrical. The dotted line depicts the peripheral electric field that the shark will turn way after the encounter (∼9.7 V/m), and the solid line depicts the core electric field that is strong enough so that shark will never touch (∼15.7 V/m). We define *optimal planes* as where the generated electric field has the largest radius at these planes. There are two optimal planes located between each pair of electrodes. Let $l_e$ be the height difference between the operator and the top of the peripheral electric field. To keep the shape of the electric field underwater, we assume that the distance between of operator the sea surface cannot be larger than $l_e$. Let $l_1$ be the height difference between the operator and another optimal plane that is closer to the operator. Let $l_2$ be the height difference between the operator and the optimal plane that is farther to the operator. Let $l_{min}$ be the minimum distance between the operator and the sea surface to keep the drone safe. Let $l_{max}$ be the height difference between the operator and the bottom of the peripheral electric field, as shown in Figure 5.2. Their relationship is given by:

$$l_{min} < l_e < l_1 < l_2 < l_{max}. \tag{5.1}$$

Let $\varepsilon(h)$ be the radius of the peripheral electric field on a plane, where $h$ is the height difference between the plane and the operator. Let $\varepsilon_o$ be the radius of the peripheral electric field on the optimal planes. We have

$$\varepsilon_o = \varepsilon(l_1) = \varepsilon(l_2), \tag{5.2}$$

where $\varepsilon_o$ and $\varepsilon(h)$ are assumed to be known.

We assume that the shark will immediately turn away after its snout touches the peripheral electric field, and the shark would not cross the core electric field. Figure 5.3 shows the top view of the turning trajectories of a shark encountered the electric field, where $D$ is the operator, $L$ is the *ampullae of Lorenzini* located on the shark's snout. $\lambda_1$ and $\lambda_2$ are two tangent lines between $L$ and the core electric field. $\mu_1$, $\mu_2$, $\mu_3$, $\mu_4$, $\mu_5$, and $\mu_6$ are examples of shark's turning trajectories in top view. Notably, $\mu_1$ and $\mu_6$ are trajectories that the shark is still approaching humans, and the others show that the shark has (temporarily) stopped approaching any human after it encountered with the electric field. We assume that sharks can only dive when moving forward (will still hit the electric field), and cannot dive directly to avoid the electric field.

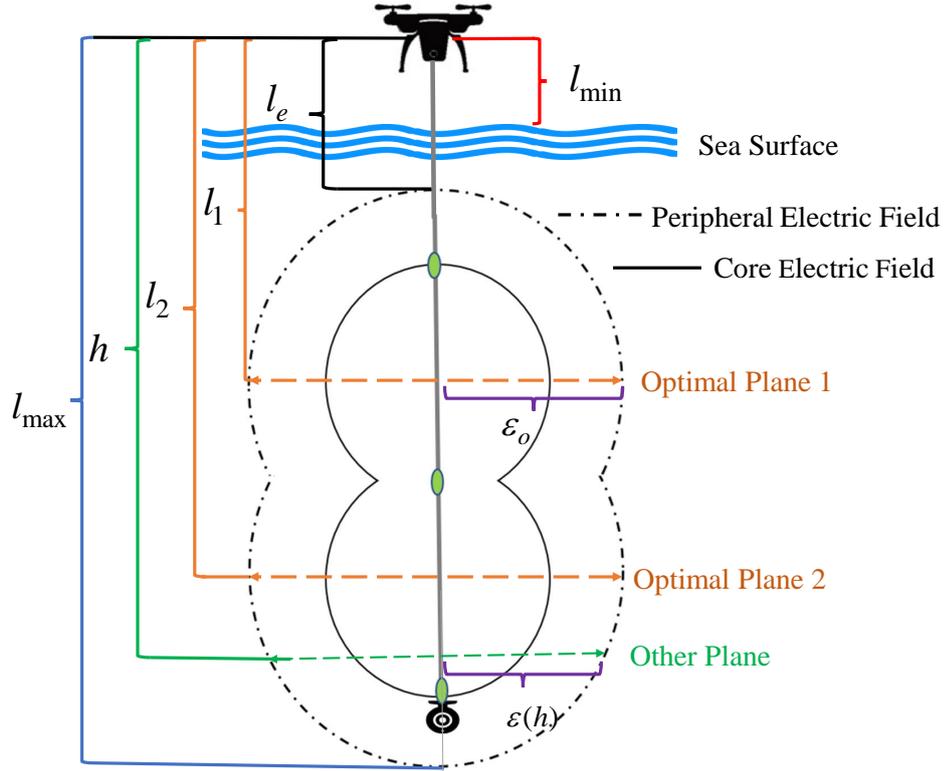

Figure 5.2: Schematic representation of the electric field generated by the operator.

# 5.3 Repelling Strategy and Interception Algorithm

In this section, we present the repelling strategy of the drone shark shield system to repel sharks for protecting swimmers and surfers, and then propose an algorithm for an operator to pursues and intercepts a moving shark in a limited time.

## 5.3.1 Shark Repelling Strategy

The drone shark shield system consists of several observers and operators in a certain formation that regularly patrol the sea area near the beach. Let $n$ be the number of operators in the current formation. Once a number of $1 \leq s \leq n$ sharks are spotted by any observer during the aerial surveillance, $s$ operators immediately leave the surveillance formation to repel the sharks following certain procedures that will be introduced later. For each operator, we allocate one shark for it. At the same time, one observer immediately leaves the surveillance formation, stays at a certain altitude to monitor the sharks, identifies humans and provides the relative information to the operators. Meanwhile, it uses the attached speakers to alarm swimmers and surfers to evacuate the water immediately. The speakers should be loud enough so that the nearby swimmers and surfers can hear the alarm immediately once a shark is spotted. If no human is detected near to the shark, it provides early warning to swimmers or surfers through the lifeguards. Moreover, the rest of the operators and observers in the formation continue the surveillance mission to spot and repel the sharks that have not been spotted. If $s > n$ sharks are spotted at the same time, the observer sends the highest alert to the lifeguards to close the beach immediately and deploy all operators to repel the $n$ sharks that are closer to humans in the water. A drone shark shield system with a sufficient number of operators and observers can effectively avoid this worst-case scenario.

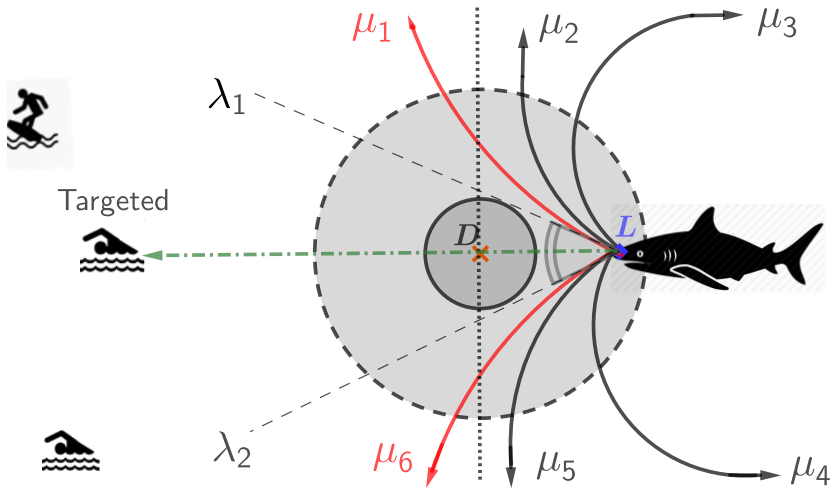

Figure 5.3: Top view of turning trajectories of the shark encountered with the electric field.

We now introduce the procedures for a single operator to repel its allocated shark. The key idea is to use the generated electric field to intercepts the shark whenever it is approaching any human. The flow diagram of the shark repelling procedures can be seen in Figure 5.4, and the detailed procedures are explained as follows:

1. The operator leaves the surveillance formation after shark spotted, triggers the inflation process. It inserts the inflated tube into the seawater, immerses the attached miniature sonar an all the electrodes in the seawater while keeps a safe distance above the sea surface. If the observer finds that the shark is very close to any human, the operator covers the targeted human with the electric field immediately before the shark attacks him/her. Otherwise, the operator stays close to the shark but keeps a certain distance between the shark and the electric field.

2. The system detects the positions and velocities of the shark and nearby humans based on the combined information from the observer and the sonar. Once finding the shark is approaching any human, the operator immediately intercepts the shark with the electric field following a certain algorithm that will be introduced later. If no human is detected near the shark, intercepts the shark if it attempts to approach the beach.

3. To avoid the decrease of shark's sensitivity to the electric field caused by continuous stimulation, the operator waits for a certain time $T_w$ to see shark's response after each interception. If the shark stops approaching any human and is leaving the beach area, the operator follows the shark but keeps a certain distance from the shark. If the shark dives and keeps diving to avoid touching the electric field, the operator simply follows the shark, stays at a certain distance ahead of the shark's moving direction, and covers the targeted human with the electric field before the shark attacks him/her.

4. Repeat step 2) and step 3) until the shark leaves the beach area for a certain safety distance (e.g. 3 km).

5. The operator returns to the base station for replacing the carbon dioxide cylinder and replacing or recharging its battery.

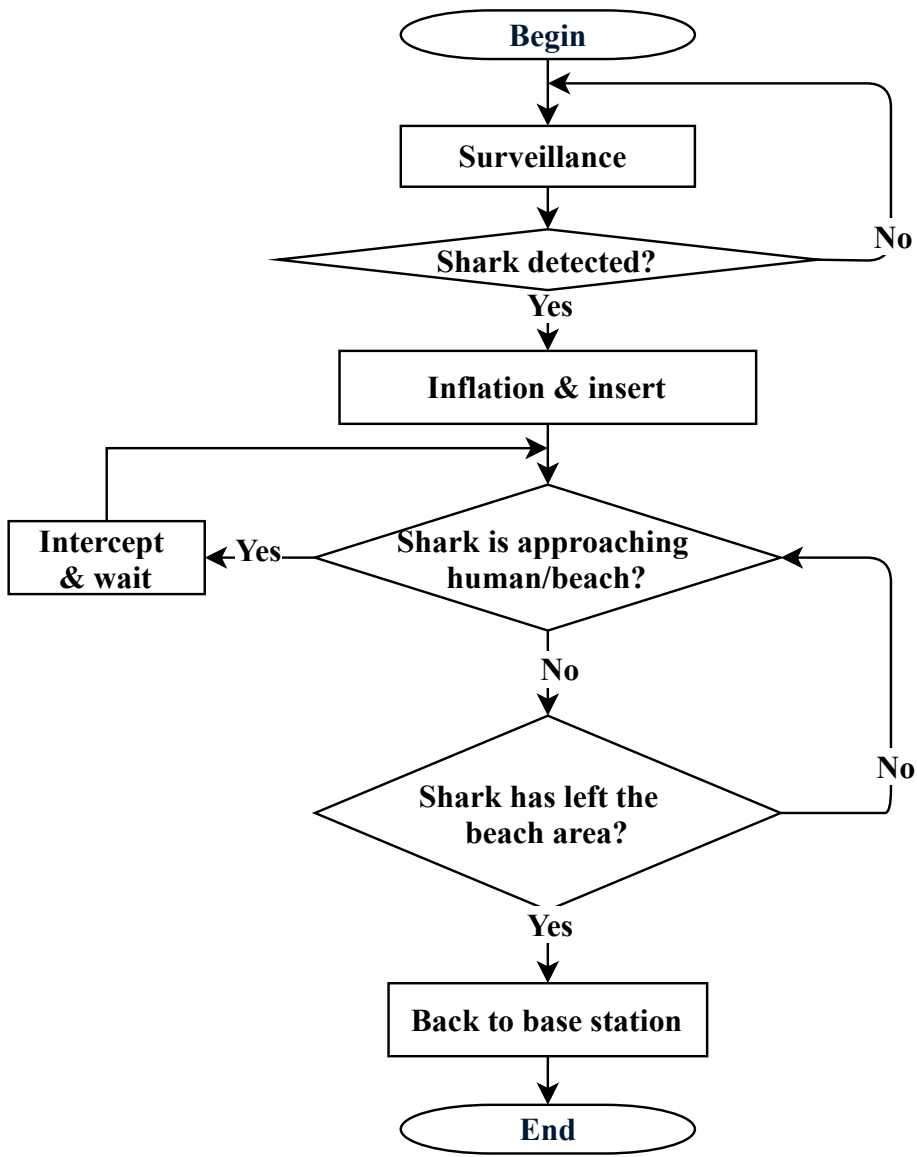

Figure 5.4: Flow diagram of the shark repelling procedures.

## 5.3.2 Interception Algorithm

We now introduce the motion control algorithm designed for an operator to intercepts a moving shark. The interception problem is a continuous-time pursuit-evasion problem involving a single pursuer and a single evader in a 3D environment [188–190]. Instead of considering the entire shark as the evader, we only care about the movement of the shark's snout, where the highly sensitive electric receptor *ampullae of Lorenzini* is located. For simplicity, in the following part of this section, every 'shark' we mention is actually the 'shark's snout'.

Let $O(t) := [x_o(t), y_o(t), z_o(t)]$ and $S(t) := [x_s(t), y_s(t), z_s(t)]$ be the Cartesian coordinates of the operator and the shark in the 3D space, respectively. The z-axis of the global coordinate system is set to be the opposite direction of gravity. Let the altitude of the sea surface be zero, i.e. $z = 0$. To better illustrate the system and our method, we decouple the movements of the operator and the shark in the vertical dimension from the horizontal dimension. We define $O_{xy}(t) := [x_o(t), y_o(t)]$ and $S_{xy}(t) := [x_s(t), y_s(t)]$ as the horizontal coordinates of the operator and the shark, respectively. $z_o(t)$ and $z_s(t)$ are the vertical coordinates of the operator and the shark, respectively.

The horizontal velocity of the operator is given by $\vec{v}_{oxy}(t) := [v_{ox}(t), v_{oy}(t)]$ with $\|\vec{v}_{oxy}(t)\| \leq V_{oMxy}$, where $V_{oMxy} > 0$ is the maximum speed of the operator in the horizontal dimension. Similarly, the horizontal velocity of the shark is given by $\vec{v}_{sxy}(t) := [v_{sx}(t), v_{sy}(t)]$ with $\|\vec{v}_{sxy}(t)\| \leq V_{sMxy}$, where $V_{sMxy} > 0$ is the maximum speed of the shark in the horizontal dimension. Let $\vec{v}_{oz}(t)$ and $\vec{v}_{sz}(t)$ be the vertical velocities of the operator and the shark, respectively, and $V_{oMz}$ and $V_{sMz}$ be the maximum vertical speeds of the operator and the shark, respectively. The operator is set to be faster than the shark, and it satisfies that $V_{oMxy} > V_{sMxy}$, $V_{oMz} > V_{sMz}$. Let $V_{oM}$ and $V_{sM}$ be the maximum speeds of the operator and the shark, respectively. Naturally, we have

$$V_{oM} = \sqrt{V_{oMxy}^2 + V_{oMz}^2} > V_{sM} = \sqrt{V_{sMxy}^2 + V_{sMz}^2}. \tag{5.3}$$

The observer and operator are rotary-wing UAVs with their mobility modelled as an integral process, which is commonly used to model the mobility of different drone products such as the DJI Inspire series. The considered system evolves as

$$\begin{aligned}
\dot{O}_{xy} = \vec{v}_{oxy}; \quad & \dot{O}_z = \vec{v}_{oz}; \quad \vec{v}_o = \vec{v}_{oxy} + \vec{v}_{oz}; \\
\dot{S}_{xy} = \vec{v}_{sxy}; \quad & \dot{S}_z = \vec{v}_{sz}; \quad \vec{v}_s = \vec{v}_{sxy} + \vec{v}_{sz}.
\end{aligned} \tag{5.4}$$

To effectively dispel the shark's offensive intentions, the operator should intercept the shark right in front of it. We now define *interception plane* as the cross-section of the electric field at the shark's altitude. We also define *interception* as
the instance when $\vec{v}_{sxy}$ is pointing from $S_{xy}$ to $O_{xy}$, which is also the center of the
electric field on the interception plane. Also, the horizontal distance between the
operator and the shark should be $\varepsilon(h)$, where $h = z_o - z_s$. An example of the top
view of the interception can be seen in Figure 5.5, where $t_i$ is the time when the
interception occurs.

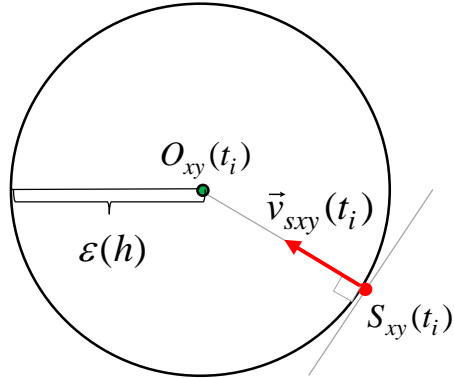

Figure 5.5: An example of the top view of the interception.

The time-optimal strategy for the operator to pursues and intercepts the shark
is to move with the maximum speed in the direction of the shark. Such a strategy,
known as classical pursuit [191], is given by the control law:

$$\vec{v}_o(t) = V_{oM} \frac{S(t) - O(t)}{\|S(t) - O(t)\|}. \tag{5.5}$$

Since we assume that the shark will turn away whenever it touched the electric filed,
in order to efficiently intercept the shark and dispel the shark's offensive intentions,
it is required that the electric field should not touch the shark during the pursuit.
However, direct pursuit following the control law (5.5) could lead to the operator
intercepts the shark from behind, which will obviously violate this requirement.

To solve this problem, we decouple the operator movement in the vertical di-
mension from the horizontal dimension. In the vertical dimension, recall that the
generated electric field has a special shape, as shown in Figure 5.2. $\varepsilon(h)$ is the radius
of the electric field on the interception plane. We assume $z_s$ is known. Thus, by adjusting $z_o$, we can always try to intercept the shark with the optimal planes (i.e. with the largest electric field radius) for different $z_s$.

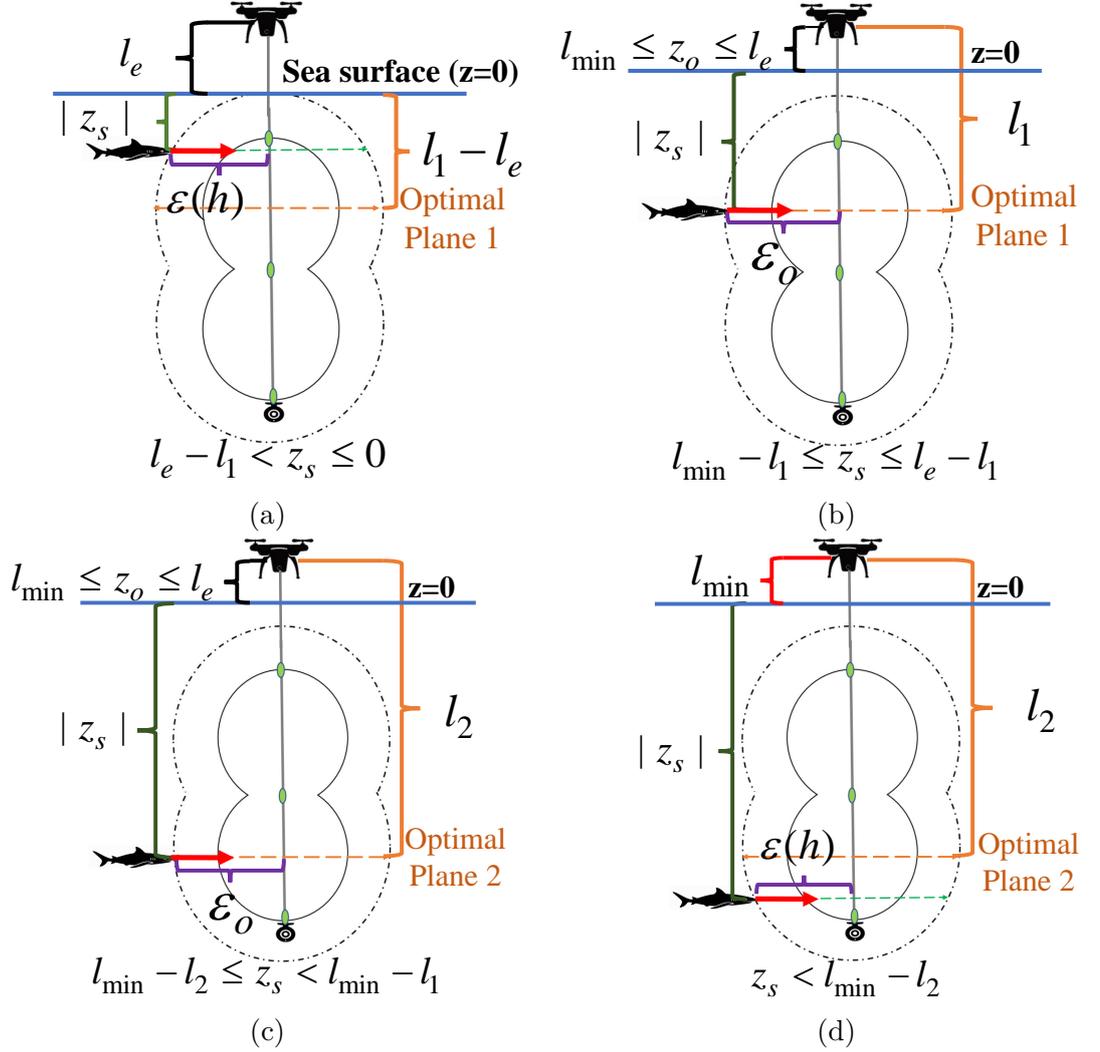

Figure 5.6: Maximize $\varepsilon^*$ for different values of $z_s$ in 4 cases. (a) When the shark is too close to the sea surface, $l_e - l_1 < z_s \le 0$. (b) The shark stays slightly deeper, $l_{min} - l_1 \le z_s \le l_e - l_1$. (c) The shark stays deeper, $l_{min} - l_2 \le z_s < l_{min} - l_1$. (d) The shark stays too deep, $z_s < l_{min} - l_2$.

**Proposition 5.3.1.** Let $z_o^*$ be the optimal altitude of the operator that gives the largest possible interception radius $\varepsilon^*$ of the electric field. Then $z_o^*$ and $\varepsilon^*$ are piece-wise functions depending on the value of $z_s$, as shown following:

$$z_o{}^* = \begin{cases} l_e, & \text{if} \quad l_e - l_1 < z_s \leq 0 \\ z_s + l_1, & \text{if} \quad l_{min} - l_1 \leq z_s \leq l_e - l_1 \\ z_s + l_2, & \text{if} \quad l_{min} - l_2 \leq z_s < l_{min} - l_1 \\ l_{min}, & \text{if} \quad z_s < l_{min} - l_2 \end{cases} ; \tag{5.6}$$

$$\varepsilon^* = \begin{cases} \varepsilon(l_e - z_s), & \text{if} \quad l_e - l_1 < z_s \leq 0 \\ \varepsilon_o, & \text{if} \quad l_{min} - l_2 \leq z_s \leq l_e - l_1 \\ \varepsilon(l_{min} - z_s), & \text{if} \quad l_{min} - l_{max} \leq z_s < l_{min} - l_2 \\ 0, & \text{if} \quad z_s < l_{min} - l_{max} \end{cases} . \tag{5.7}$$

*Proof.* We aim to maximize $\varepsilon^*$ for different values of $z_s$. We categorize the values of $z_s$ into 4 cases with $z_s$ gradually decreases from zero, as shown in Figure 5.6. In case (a), if the shark is too close to the sea surface, i.e., $l_e - l_1 < z_s \leq 0$, the operator cannot intercept the shark with the optimal planes, because $z_o \leq l_e$ always holds to keep the shape of the electric field underwater. In this case, the operator should keep at its highest altitude $z_o{}^* = l_e$ to maximize $h$ since $\varepsilon(h)$ increases with the growth of $h$. Thus, $\varepsilon^* = \varepsilon(l_e - z_s)$ since $h = l_e - z_s$. Following a similar pattern, in case (b), if the shark stays slightly deeper and $l_{min} - l_1 \leq z_s \leq l_e - l_1$, the operator can intercept the shark with optimal plane 1. In this case, the operator should keep at the altitude $z_o{}^* = z_s + l_1$ to maintain $h = z_o{}^* - z_s = l_1$, so that $\varepsilon^* = \varepsilon_o$. Specifically, if $z_s = l_{min} - l_1$, the shark has equal distance to the two optimal planes, and the operator can use either optimal plane 1 or optimal plane 2 to intercepts the shark. For simplicity, we let $z_o{}^* = z_s + l_1$ (i.e. use optimal plane 1) when $z_s = l_{min} - l_1$. In case (c), if the shark stays deeper and $l_{min} - l_2 \leq z_s < l_{min} - l_1$, the operator can intercept the shark with optimal plane 2. Similar to case (b), the operator should keep at the altitude $z_o{}^* = z_s + l_2$ to maintain $h = z_o{}^* - z_s = l_2$, and $\varepsilon^* = \varepsilon_o$. In case (d), if the shark stays too deep and $z_s < l_{min} - l_2$, the operator cannot intercept the shark with the optimal planes, because $z_o \geq l_{min}$ always holds to keep the safety of the operator. In this case, the operator should keep at its lowest altitude $z_o{}^* = l_{min}$ to minimize $h$ since $\varepsilon(h)$ decreases with $h$ increases. Therefore, $\varepsilon^* = \varepsilon(l_{min} - z_s)$ since $h = l_{min} - z_s$. Note that if $z_s < l_{min} - l_{max}$, the electric field cannot reach the shark, thus $\varepsilon^* = 0$. In such circumstances, the operator still keeps at its lowest altitude $z_o^* = l_{min}$, and waits for the shark rises. This completes the proof. □

Next, we discuss the operator movement in the horizontal dimension. We define a vector $\vec{p}$ that is generated by rotating $\vec{v}_{sxy}$ 90 degrees clockwise around the z-axis if $y_o \leq y_s$, and by rotating $\vec{v}_{sxy}$ 90 degrees anticlockwise around the z-axis if $y_o < y_s$. We imagine there are two virtual sharks $S_1$ and $S_2$ at the same altitude $z_o^*$, with $S_{1xy}$ and $S_{2xy}$ as their horizontal coordinates, respectively. $S_{2xy}$ is generated by moving $S_{xy}$ a distance of $\varepsilon'$ on the direction of $\vec{v}_{sxy}$, where

$$\varepsilon' := \varepsilon^*(1 + \gamma). \tag{5.8}$$

Here, $\gamma > 0$ is a pre-defined small constant. $S_{1xy}$ is generated by moving $S_{2xy}$ the same distance on the direction of $\vec{p}$, as shown in Figure 5.7. The horizontal velocities of both $S_1$ and $S_2$ are $\vec{v}_{sxy}$, i.e. same as $S$, which means $S$, $S_1$, and $S_2$ are relatively static in the horizontal dimension.

$S_1$ and $S_2$ can be calculated by:

$$S_{2xy} = \frac{\vec{v}_{sxy}}{|\vec{v}_{sxy}|}\varepsilon' + S_{xy}; \tag{5.9}$$

$$S_{1xy} = \frac{\vec{p}}{|\vec{p}|}\varepsilon' + S_{2xy}. \tag{5.10}$$

Let $\vec{p} = (x_p, y_p)$. Since $\vec{p}$ is generated by rotating $\vec{v}_{sxy}$ around the z-axis, $\vec{p}$ can be calculated by:

$$\begin{bmatrix} x_p \\ y_p \end{bmatrix} = \begin{bmatrix} \cos\theta & -\sin\theta \\ \sin\theta & \cos\theta \end{bmatrix} * \begin{bmatrix} v_{sx} \\ v_{sy} \end{bmatrix}; \tag{5.11}$$

$$\theta = \begin{cases} -90^o, & \text{if} \quad y_o < y_s \\ 90^o, & \text{if} \quad y_o \geq y_s. \end{cases} \tag{5.12}$$

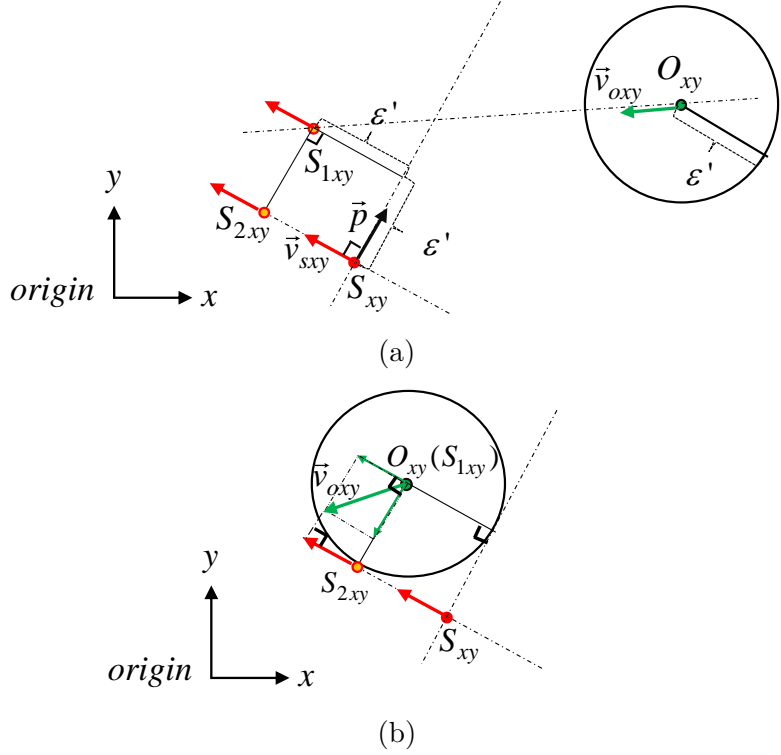

Figure 5.7: Top view of (a) the operator that is pursuing the virtual shark $S_1$ and (b) pursuing the virtual shark $S_2$ after captured $S_1$.

Substituting (5.12) into (5.11), we can get

$$\vec{p} = \begin{cases} (v_{sy}, -v_{sx}), & \text{if} \quad y_o < y_s \\ (-v_{sy}, v_{sx}), & \text{if} \quad y_o \geq y_s \end{cases}. \tag{5.13}$$

Known $O_{xy}$, $S_{xy}$ and $\vec{v}_{sxy}$, $S_{1xy}$ and $S_{2xy}$ can be calculated from (5.6), (5.7), (5.8), (5.9), (5.10), (5.13), and $S_1 = (S_{1xy}, z_o{}^*)$, $S_2 = (S_{2xy}, z_o{}^*)$.

Now, we are ready to introduce the algorithm for the operator to efficiently intercept a single shark:

1. Once the operator starts the interception mission, it first pursues the virtual shark $S_1$ following the control law:

$$\vec{v}_o(t) = V_{oM} \frac{S_1(t) - O(t)}{\|S_1(t) - O(t)\|} \tag{5.14}$$

until

$$\|S_1(t) - O(t)\| \leq \gamma. \tag{5.15}$$

We also call the pre-defined positive small constant $\gamma$ the 'capture' radius.

2. Then, the operator immediately starts to pursue the virtual shark $S_2$ following the control laws:

$$\vec{v}_{oxy}(t) = (-\vec{p'})\sqrt{\frac{V_{oMxy}^2}{\|\vec{v}_{sxy}(t)\|^2} - 1} + \vec{v}_{sxy}(t); \tag{5.16}$$

$$\vec{v}_{oz}(t) = V_{oMz}\frac{z_o^* - z_o(t)}{\|z_o^* - z_o(t)\|}; \tag{5.17}$$

$$\vec{v}_o(t) = \vec{v}_{oxy}(t) + \vec{v}_{oz}(t); \tag{5.18}$$

until

$$\|S_2(t) - O(t)\| \leq \gamma. \tag{5.19}$$

where $\vec{p'}$ is generated by rotating $\vec{v}_{sxy}$ 90 degrees clockwise around the z-axis, if $y_o \leq y_2$; and by rotating $\vec{v}_{sxy}$ 90 degrees anticlockwise around the z-axis, if $y_o < y_2$. The only difference between $\vec{p'}$ and $\vec{p}$ is their piecewise conditions. Similar to (5.11), (5.12) and (5.13), $\vec{p'}$ can be calculated as

$$\vec{p'} = \begin{cases} (v_{sy}, -v_{sx}), & \text{if} \quad y_o < y_2 \\ (-v_{sy}, v_{sx}), & \text{if} \quad y_o \geq y_2 \end{cases}. \tag{5.20}$$

3. After step 2), the operator immediately slows down. The shark will directly hit the electric field with the largest possible radius, and $\vec{v}_{sxy}$ will be pointing from $S_{xy}$ to $O_{xy}$.

**Proposition 5.3.2.** Let $D^0$ be the initial separation between the operator and the shark. Given that $V_{oMxy} > V_{sMxy}$, $V_{oMz} > V_{sMz}$, $D^0 > \varepsilon_o$ and $\gamma < D_0$, it is guaranteed that the proposed algorithm navigates the operator to intercepts the shark in a finite time.

*Proof.* We first prove that the operator can 'capture' $S_1$ in a finite time. Let $t_0 = 0$

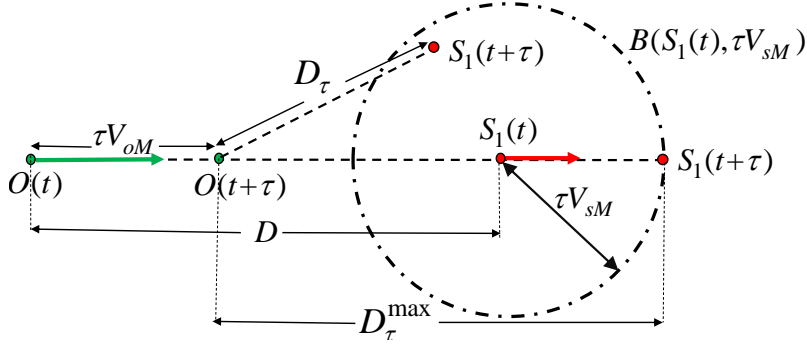

Figure 5.8: The operator (●) is initially at $O(t)$ and $S_1$ (●) is initially at $S_1(t)$.
After the duration $\tau$, they have moved to $O(t + \tau)$ and $S_1(t + \tau)$, respectively.
$B\left(S_1(t), \tau V_{sM}\right)$ outlines the boundary of the reachable set of $S_1$ after $\tau$. $D_\tau^{\max}$
is the maximum possible separation between the operator and $S_1$ at $t + \tau$. The
green arrow denotes $\vec{v}_o$, the red arrow denotes $\vec{v}_s$.

be the time when the operator starts pursuing $S_1$. Let $t > 0$ be an arbitrary time
before the operator 'captures' $S_1$. Let $D$ be the separation between the operator and
$S_1$ at time $t$. Suppose that the new separation between them, after a duration of $\tau$,
is $D_\tau$. $\tau$ is small enough so that the velocities of the operator and the shark can be
seen as unchanged, during the duration of $\tau$. If the operator pursues $S_1$ following
the control law (5.14) and (5.15), the maximum distance between the operator and
$S_1$, at time $T + \tau$, is

$$D_\tau^{max} = D + (V_{sM} - V_{oM})\tau. \tag{5.21}$$

Let $\vec{q} := \frac{S_1(t) - O(t)}{\|S_1(t) - O(t)\|}$, (5.16) becomes:

$$\vec{v}_o(t) = V_{oM}\vec{q}. \tag{5.22}$$

$\vec{q}$ is a unit vector that is pointing from $S_1(t)$ to $O(t)$. After the duration $\tau$, the
operator moves a distance of $\tau V_{oM}$ units:

$$\|\vec{v}_o(t)\| = \|V_{oM}\vec{q}\| = V_{oM}. \tag{5.23}$$

$S_1$ can be anywhere inside a ball of radius $\tau V_{sM}$ centered at $S_1(t)$ (as it moves
with speed $\|\vec{v}_s(t)\| < V_{sM}$). The maximum separation between the operator and $S_1$

is denoted by $D_\tau^{max}$ and is given by $D + (V_{sM} - V_{oM})\tau$. As shown in Figure 5.8, when $S_1(t + \tau)$ locates on the farthest point to $O(t + \tau)$, on the reachable set boundary. Since $V_{oM} > V_{sM}$, we have

$$D_\tau < D_\tau^{max} = D + (V_{sM} - V_{oM})\tau < D. \qquad (5.24)$$

This means the separation between the operator and $S_1$ is strictly decreasing. Therefore, given some pre-defined positive 'capture' radius $\gamma < D_0$, the control law (5.14) and (5.15) ensure that $S_1$ can be 'captured' in a finite time. In addition, (5.23) shows that the operator is always moving with its maximum speed in the direction of the $S_{1xy}$, thus step 1) can be finished in the shortest time. Specifically, [192] has proved that the time-to-capture is bounded by

$$T_{\text{cap}} \leq -\frac{D_0 - \gamma}{V_{oM} - V_{sM}}, \qquad (5.25)$$

and the worst-case scenario is that $S_1$ is moving directly away from the operator with its maximum speed $V_{sM}$, at all times.

In step 2), the operator starts pursuing $S_2$ from $S_1$. The initial separations between the operator and $S_2$ in the horizontal and vertical dimensions are $D_{xy}^0 = \varepsilon'$ and $D_z^0 = 0$, respectively, as shown in Figure 5.7(b). Since $\vec{p'}$ is generated by rotating $\vec{v}_{sxy}$ 90 degree or -90 degree, we have $\left\|\vec{p'}\right\| = \|\vec{v}_{sxy}\|$ and $\vec{p'} \cdot \vec{v}_{sxy} = 0$. Thus, it always holds that

$$
\begin{aligned}
\|\vec{v}_{oxy}(t)\|^2 &= \left\|\vec{p'}\right\|^2 \left(\frac{V_{oMxy}^2}{\|\vec{v}_{sxy}(t)\|^2} - 1\right) + \|\vec{v}_{sxy}(t)\|^2 \\
&\quad - 2\vec{v}_{sxy}(t)\vec{p'}\sqrt{\frac{V_{oMxy}^2}{\|\vec{v}_{sxy}(t)\|^2} - 1} \\
&= \left\|\vec{p'}\right\|^2 \left(\frac{V_{oMxy}^2}{\|\vec{v}_{sxy}(t)\|^2} - 1\right) + \|\vec{v}_{sxy}(t)\|^2 \\
&= \|\vec{v}_{sxy}(t)\|^2 \left(\frac{V_{oMxy}^2}{\|\vec{v}_{sxy}(t)\|^2} - 1\right) + \|\vec{v}_{sxy}(t)\|^2 \\
&= V_{oMxy}^2;
\end{aligned}
\qquad (5.26)
$$

$$\|\vec{v}_{oz}(t)\|^2 = V_{oMz}^2 \frac{\|z_o^* - z_o(t)\|}{\|z_o^* - z_o(t)\|} = V_{oMz}^2 \qquad (5.27)$$

which means the operator is always pursuing $S_2$ with its maximum speed.

In the horizontal dimension, let $\vec{v}_{os\_xy}$ denote the relative velocity between the operator and $S_2$. It can be calculated from (5.16) that

$$\vec{v}_{so\_xy}(t) = \vec{v}_{oxy}(t) - \vec{v}_{sxy}(t) = (-\vec{p'})\sqrt{\frac{V_{oMxy}^2}{\|\vec{v}_{sxy}(t)\|^2} - 1}. \qquad (5.28)$$

Since $\vec{p'}$ and $\vec{p}$ have the same direction that is pointing from $S_2$ to $S_1$, the direction of $\vec{v}_{so\_xy}(t)$ will be always pointing from $S_1$ to $S_2$. Moreover,

$$\begin{aligned}
\|\vec{v}_{so\_xy}(t)\|^2 &= \|\vec{p'}\|^2 \left(\frac{V_{oMxy}^2}{\|\vec{v}_{sxy}(t)\|^2} - 1\right) \\
&= \|\vec{v}_{sxy}\|^2 \left(\frac{V_{oMxy}^2}{\|\vec{v}_{sxy}(t)\|^2} - 1\right) \\
&= V_{oMxy}^2 - \|\vec{v}_{sxy}\|^2 > V_{oMxy}^2 - V_{sMxy}^2 > 0.
\end{aligned} \qquad (5.29)$$

This means the separation between the operator and $S_2$ in the horizontal dimension is strictly decreasing.

Note that control law (5.17) and control law (5.14) have a similar form. Thus, in the vertical dimension, $z_o(t)$ will 'capture' $z_o^*$ in a finite time. This can be proved following the similar pattern of step 1)'s proof, except that it is simpler with just one dimension. The detailed proof is omitted. Combining the conclusions in the horizontal dimension and the vertical dimension, the control laws (5.16), (5.17), (5.18) and (5.19) will navigate the operator from $S_1$ to $S_2$, and step 2) can be finished in a finite time. Particularly, at the instant when the operator 'captures' $S_2$, the horizontal separation between the operator and the shark will be $D' = \varepsilon' = \varepsilon^*(1+\gamma)$. Since $\gamma$ is a small positive constant, the shark will directly hit the electric field if the operator immediately slows down after 'captured' $S_2$. This completes the proof. $\square$

**Proposition 5.3.3.** The proposed algorithm ensures that the generated electric field will not touch the shark before the interception point.

*Proof.* We first prove that the generated electric field will not touch the shark during step 1). Let $D_{xy}$ be the separation between the operator and the shark in the horizontal dimension at time $t$. $D_{xy\tau}$ denotes the separation between the operator and the shark in the horizontal dimension at time $t + \tau$. Let $D_{xy\tau}^{\min}$ be the minimum possible separation between the operator and the shark at $t + \tau$ in the horizontal dimension. As shown in Figure 5.9, when $S_{1xy}(t + \tau)$ is locating on the closest point to $O_{xy}(t + \tau)$, on the reachable set boundary. As indicated in Figure 5.9, $D_{xy\tau}^{\min}$ is the hypotenuse (the yellow dash line) of a right triangle with one leg (the blue dash line) equals to $\varepsilon'$. Thus, we have

$$D_{xy\tau}^{\min} > \varepsilon' = \varepsilon^*(1 + \gamma) > \varepsilon^* \tag{5.30}$$

at the time $t+\tau$, since $\gamma > 0$. Particularly, at the instant that the operator 'captures' $S_1$, the horizontal separation between the operator and the shark will be

$$D_{xy}^{cap} = \sqrt{2}\varepsilon' > \varepsilon' > \varepsilon^*. \tag{5.31}$$

Since initially $D_{xy}^0 > \varepsilon_o \geq \varepsilon^*$, $t + \tau$ is an arbitrary time before the operator 'captures' $S_1$, and $\varepsilon^*$ is the largest electric field radius at the shark's altitude, it can be concluded that the generated electric field will not touch the shark in step 1).

Then, we prove that the generated electric field will not touch the shark during step 2). Recall that $S_2$ and the shark are relatively static in the horizontal dimension. Thus, during step 2), the relative velocity between the operator and the shark is the same as $\vec{v}_{so\_xy}(t)$, which is parallel to $\vec{p'}$ and perpendicular to $\vec{v}_{sxy}(t)$. Let $D'$ be the horizontal separation between the operator and the shark during step 2). It can be seen from Figure 5.7(b) that $D'$ satisfies

$$\varepsilon' \leq D' \leq \sqrt{2}\varepsilon'. \tag{5.32}$$

Thus,

$$D' \geq \varepsilon' = \varepsilon^*(1 + \gamma) > \varepsilon^*. \tag{5.33}$$

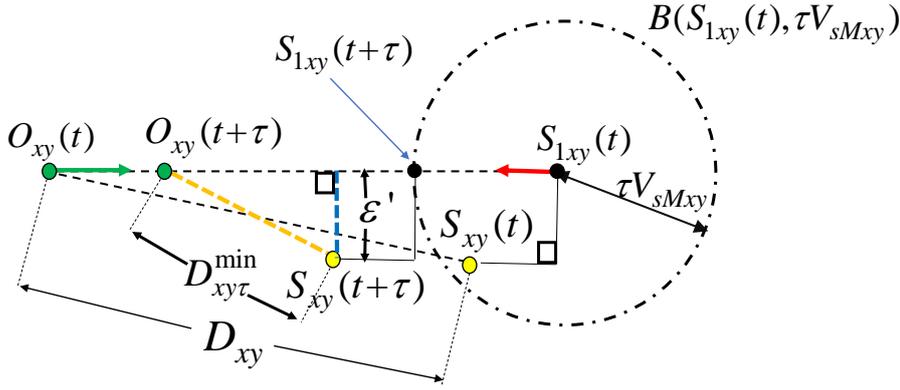

Figure 5.9: In the horizontal dimension, the operator (●) is initially at $O_{xy}(t)$, $S_1$
(●) is initially at $S_{1xy}(t)$, and the shark (●) is initially at $S_{xy}(t)$. After the duration $\tau$, they have moved to $O_{xy}(t+\tau)$, $S_{1xy}(t+\tau)$ and $S_{xy}(t+\tau)$, respectively.
$B\left(S_{1xy}(t), \tau V_{sMxy}\right)$ outlines the boundary of the reachable set of $S_{1xy}$ after $\tau$. $D_{xy\tau}^{\min}$
is the minimum possible horizontal separation between the operator and the shark
at $t+\tau$.

Therefore, the generated electric field will not touch the shark during step 2). This
completes the proof.                                                                    □

**Proposition 5.3.4.** *The proposed algorithm guarantees the electric field has the
largest possible radius on the interception plane, and $\vec{v}_{sxy}$ will be pointing from $S_{xy}$
to $O_{xy}$ when the interception occurs.*

*Proof.* According to Proposition , $z_o{}^*$ is the optimal altitude of the operator that
gives the largest possible interception radius $\varepsilon^*$ of the electric field. Since the altitude
of $S_2$ is $z_o{}^*$, the electric field on the interception plane will have the largest possible
radius if the operator 'captures' $S_2$. It can be inferred from Figure 5.7(b) that $\vec{v}_{sxy}$
will be pointing from $S_{xy}$ to $O_{xy}$ when the interception occurs, because $\vec{v}_{sxy}$ is always
pointing from $S_{xy}$ to $S_{2xy}$. This completes the proof.                                □

## 5.4   Simulation Results

In this section, we demonstrate the performance of the proposed interception
algorithm via computer simulations. All the simulations are carried out with MAT-
LAB. The simulation parameters are as shown in Table 5.1, where $D_i$ is the ini-tial separations between the operator and the shark, and $Ts$ is the sampling time. The generated electric field is simplified as two partly overlapped balls centred at $[x_o, y_o, z_o - l_1]$ and $[x_o, y_o, z_o - l_1]$, respectively, of radius of $\varepsilon_o$. As indicated in Figure 5.10, $z_o^*$ and $\varepsilon^*$ are all piece-wisely changing when $z_s$ is increasing, and $z_o^*$ may drop certainly to keep $\varepsilon^*$ being the largest, i.e. $\varepsilon^* = \varepsilon_0$. It can be inferred from Proposition that $z_o^*$ will drop from $z_s + l_2$ to $z_s + l_1$ when $z_s = l_{min} - l_1 = 3.5$m to keep $\varepsilon^* = \varepsilon_0$. Therefore, in practice, $V_{oMz}$ is required to be sufficiently large to catch this sudden change in operator's altitude.

Table 5.1: Simulation Parameter Values

| Parameters | Values | Parameters | Values |
|------------|--------|------------|--------|
| $l_{min}$ | 1 m | $l_e$ | 2 m |
| $l_1$ | 4.5 m | $l_2$ | 7.5 m |
| $l_{max}$ | 11 m | $\varepsilon_o$ | 3.5 m |
| $V_{oM}$ | 8 m/s | $r_1$ | 3500 m |
| $D_i$ | 13 m | $Ts$ | 0.1 s |

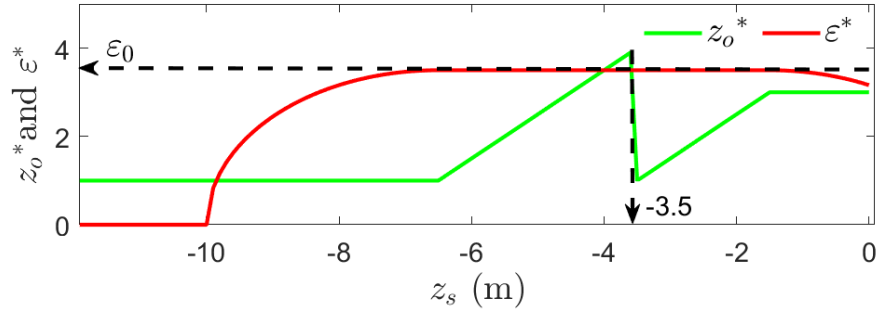

Figure 5.10: The operator's optimal altitude $z_o^*$ and the largest electric field radius on the interception plane $\varepsilon^*$ versus the shark's altitude $z_s$.

Following given trajectories, the shark moves from left to right with varying speeds and altitudes. To show the effectiveness of our proposed interception algorithm, we first simulate the case of pursuing the shark following the classical pursuit control law (5.5). Figure 5.11 illustrates the trajectories of the shark and the operator (top view) with the classical pursuit algorithm. An illustration of the corresponding electric field radius on the interception plane $\varepsilon(h)$ with the shark to operator horizontal distance $D_h$ can also be seen in Figure 5.11. The black circle is the range of the electrical field on the interception plane. As shown in Figure 5.11, the operator catches the shark from behind when the electrical field first touches the shark. In this case, the shark will not be stopped and may even accelerate to swim to the targeted human. For comparison, we simulate pursuing the shark with the proposed interception algorithm. We demonstrate four examples of the trajectories of the shark and the operator in Figure 5.11. Figure 5.12 also shows the comparison of $\varepsilon(h)$ and $D_h$, as well as the comparison of the altitudes and speeds of the shark and the operator, corresponding to each trajectory. It can be found that the proposed shark interception algorithm works for both flat (Figure 5.12(a),(b)) and curvier (Figure 5.12(c),(d)) trajectories of the shark. Moreover, comparing Figure 5.12(a), Figure 5.12(b), Figure 5.12(c) and Figure 5.12(d), we can find that the varying shark speed or altitude does not influence the performance of the proposed algorithm. In addition, Figure 5.12 suggests that $\varepsilon(h)$ is always larger than $D_h$, until the interception point, which means the generated electrical field hits the shark only at the interception point. Moreover, the heading of the shark's velocity $\vec{v}_s$ is pointing from the shark to the operator (in the horizontal dimension) at the interception point, as what we require.

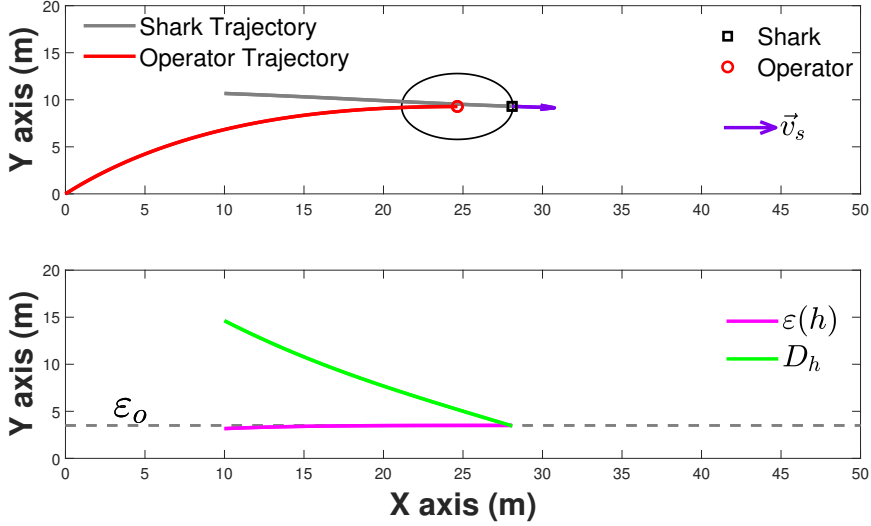

Figure 5.11: Trajectories of the shark and the operator (top view) with classical pursuit; the corresponding electric field radius on the interception plane $\varepsilon(h)$ compares with the shark to operator horizontal distance $D_h$.

Table 5.2: Average shark speeds versus interception distances.

| Trajectory | Average shark speed (m/s) | Interception distance (m) |
|:---:|:---:|:---:|
| (a) | 5.34 | 29.70 |
| (b) | 5.48 | 30.50 |
| (c) | 5.55 | 19.79 |
| (d) | 5.78 | 21.26 |

Table 5.2 compares the interception distances and the average shark speeds $||\vec{v}_s||$ for the given trajectories, where the interception distance stands for the distance between the shark's initial location and its intercepted location. As seen from Table 5.2, the interception distance of trajectories (a) and (b) are very close, with similar average $||\vec{v}_s||$. Since the major difference between trajectories (a) and (b) is that (b) has a varying $z_s$, it can be concluded that the varying shark's altitude does not significantly influence the interception distance. Similar fact can be found through the comparison of trajectories (c) and (d). However, the interception distance of trajectory (c) is considerably smaller than that of trajectory (a), with the average $||\vec{v}_s||$ of trajectory (c) slight larger than trajectory (a). Thus, a shark with curvier trajectory is likely to be intercepted in a shorter distance, because the major difference between trajectories (a) and (c) is that trajectory (c) is curvier. Similar facts can also be seen in the comparison of trajectories (b) and (d). Notably, for better presenting the operator trajectories, the maximum speed $V_{oM}$ of the operator is set relatively low, i.e. not significantly larger than the maximum speed $V_{sM}$ of the shark. The pursuit distance could be reasonably lower with sufficiently larger $V_{oM}$.

## 5.5 Conclusion

In this chapter, a novel shark defence method based on communicating autonomous drones for protecting swimmers and surfers was proposed. The goal is using autonomous drones to protect swimmers and surfers from shark attacks, and eventually, drive the shark to leave the beach area. We detailed the design of the proposed drone shark shield system and its working mechanism. We also proposed a shark repelling strategy and an interception algorithm for drones to efficiently intercept sharks. Computer simulations were conducted to demonstrate the performance of the proposed shark interception algorithm. In the future study of this chapter, tests in real-world environments need to be conducted to verify the effectiveness of the proposed method.

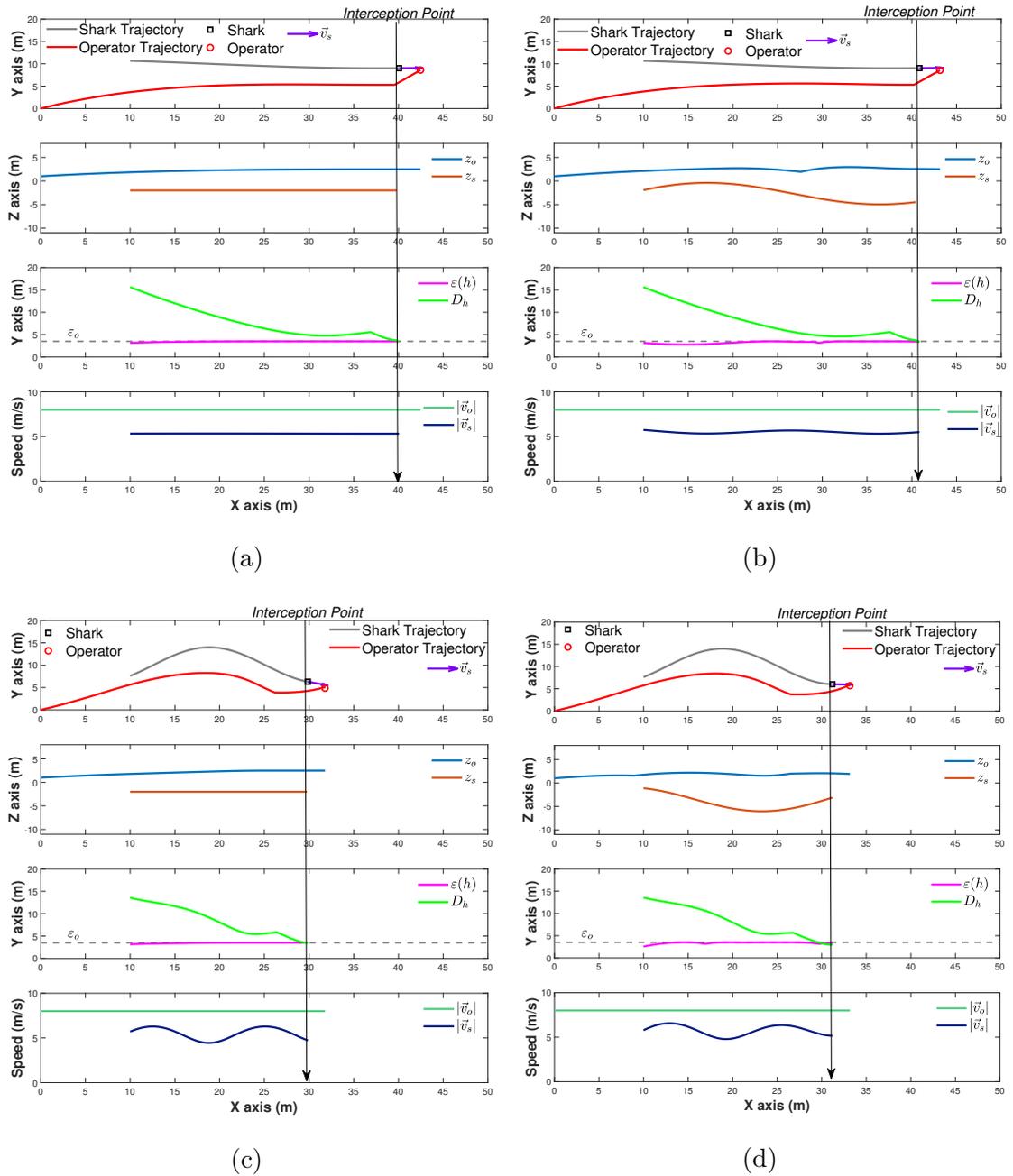

Figure 5.12: Four examples of trajectories of the shark and the operator (top view); the corresponding shark's altitude $z_s$ and operator's altitude $z_o$; the corresponding electric field radius on the interception plane $\varepsilon(h)$ compares with the shark to operator horizontal distance $D_h$; the corresponding shark speed $|\vec{v}_s|$ and operator speed $|\vec{v}_o|$. (a) shark with flat trajectory, constant altitude and speed; (b) shark with flat trajectory, varying altitude and constant speed; (c) shark with curvier trajectory, constant altitude and varying speed; (d) shark with curvier trajectory, varying altitude and speed. Videos recording the movements are available at: https://youtu.be/crY29iZ2gyU, https://youtu.be/5gwgIHlnOMA, https://youtu.be/5HOLtfpQNAA, and https://youtu.be/FopFsVFP64g.

# Chapter 6

# Autonomous Navigation of a Network of Barking Drones for Herding Farm Animals

This chapter proposes the concept of a novel automated animal herding system based on a network of autonomous barking drones. The objective of such a system is to replace traditional herding methods (e.g., dogs) so that a large number (e.g., thousands) of farm animals such as sheep can be quickly collected from a sparse status and then driven to a designated location. In this chapter, we particularly focus on the motion control of the barking drones. To this end, a computationally efficient sliding mode based control algorithm is developed, which navigates the drones to track the moving boundary of the animal herd and drive the animals to the herd center with barks. The developed algorithm also enables the drones to avoid collisions with others by a dynamic allocation of the steering points. Computer simulations are conducted to demonstrate the performance of the proposed method.

# 6.1 Motivation

Automated farming plays a critical role in preventing food crisis caused by future population growth [193]. The past decades have seen the rapid development of automated crop farming [194, 195], such as automated crop monitoring, harvesting, weed control, etc. In contrast, research and implementations of automated livestock farming have been mostly restricted in the fields of virtual fencing [196], animal monitoring and pasture surveying [197, 198]. These applications can improve livestock production yields to a certain extent. However, animal herding, as the vital step of livestock farming, has long been the least automated. Herding dogs that have been used for centuries are still the leading tools of animal herding all over the world, and the research on automated animal herding is still in its infancy. Two main obstacles of automated animal herding systems are: 1) the lack of practical robot-to-animal interactions and the suitable automated herding platforms; and 2) the lack of efficient automated herding algorithms for abundant animals.

The applications of robots to animal herding started from the Robot Sheepdog project in the 1990s [29, 30]. These groundbreaking studies achieved maneuvering a flock of ducks to a specified goal position by wheeled robots. The last three decades have only seen very few studies on automated herding with real-world animals. Recent implementations of automated animal herding mainly employ ground robots that drive animals through bright colors [31] or collisions [32–34]. But the disadvantages of the current animal herding platforms are quite obvious. The chapter [31] shows that the robot initially repulsed the sheep at a distance of 60 m; however, after only two further trials, the repulsion distance drops to 10 m. Besides, such ground legged or wheeled robots might not be agile enough to deal with various terrains during herding. Moreover, the current animal herding solutions can only herd tens of farm animals, while a modern farm can have tens of thousands of cattle or sheep.

In addition to the platforms, efficient algorithms are also critical to the study of automated animal herding. Bio-inspired swarming-based control algorithms for herding swarm robots are receiving much attention in robotics due to the effectiveness of solutions found in nature (e.g., interactions between sheep and dogs). Such algorithms can also be applied to herd animals. A considerable amount of literature has been published on this topic. For example, authors of [199] design a simple heuristic algorithm for a single shepherd to solve the shepherding problem, based on adaptive switching between collecting the swarm robots when they are too dispersed and driving them once they are aggregated. One unique contribution of [199] is that it conducted field tests with a group of real sheep and reproduces key features of empirical data collected from sheep–dog interactions. Its further study [200] extends the shepherd and swarm robots' motion and influential force vectors to the third dimension. The references [201, 202] propose the multi-shepherd control strategies for guiding swarm robots in 2D and 3D environments based on a single continuous control law. The implementation of such strategies requires more shepherds than swarm robots. Authors of [203] design a force modulation function for the shepherd agent and adopt a genetic algorithm to optimize the energy used by the agent subject to a threshold of success rate. These algorithms and most of the studies in automated herding, however, have only been carried out in the tasks with tens of swarm robots. The algorithm for efficiently herding abundant swarm robots has not been investigated.

Compared with ground robots, autonomous drones have superior maneuverability and are finding increasing use in different areas, including agriculture [204], surveillance [205], communications [92], and disaster relief [9]. Particularly, reference [204] demonstrates the feasibility of counting and tracking farm animals using drone cameras. With the ability of rapidly crossing miles of rugged terrain, drones are potentially the ideal platforms for automated animal herding, if they can efficiently interact with animals like herding dogs. Herding dogs usually herd animals by barking, glaring, or nibbling the heels of animals. For example, the New Zealand Huntaway (a species of herding dog) uses its loud, deep bark to muster sheep [206]. Drones can act like herding dogs by playing a pre-recorded dog bark loudly through a speaker, referred to as the barking drones. Recently, some successful attempts have been made using human-piloted barking drones to herd farm animals [207].

### 6.1.1 Objectives and Contributions

This chapter's primary objective is to design an automated herding system that can efficiently herd a large number of farm animals without human input. The system should be able to collect a herd of farm animals when they are too dispersed and drive them to a designated location once they are aggregated. We propose a novel automated herding system by improving the design of the human-piloted barking drones. Compared with the existing approaches of ground herding robots that drive animals through collisions or bright colors, the proposed autonomous barking drones provide more effective robot-to-animal interactions. We also develop a collision-free motion control algorithm for a network of barking drones to efficiently herd a large group of farm animals by tracking the moving boundary of the animal herd. The proposed algorithm is a computationally simple sliding mode based algorithm that can also be applied to herd swarm robots.

Practices show that compared with herding dogs, using drones to herd livestock is faster and causes less stress on animals [207]. Besides, the emerging Internet of Things (IoT) platforms for precision livestock farming [208,209] bring the possibility of closely monitoring the behaviour, welfare status, and other parameters of individual animals. With their functions being limited on non-essential applications to farmers, the return on investment of these platforms, however, can be relatively low. In addition to solving the rigid demand (i.e., herding) for farmers, the proposed system can also serve as the IoT platform to achieve the same functions. Thus, it has the potential to promote the IoT implementations in precision livestock farming, so that more animals' welfare status can be monitored and necessary health remedies can be delivered on time.

### 6.1.2 Organization

The remainder of the chapter is organized as follows. In Section 6.2, we introduce
the design of the drone herding system. Section 6.3 presents the system model and
problem statement. Drones motion control is proposed in Section 6.4. Simulation
results are presented in Sections 6.5. Finally, we give our conclusions in Section 6.6.

## 6.2 Design of the Drone Herding System

We now introduce the proposed drone herding system. It consists of a fleet
of two types of drones. The duty of the first type of drones is to detect and track
animals, called the observer. Each observer is equipped with cameras and fitted with
some Artificial Intelligence (AI) algorithms that can detect and track animals from
live video feeds with sufficient accuracy. The observer shares some similarities with
the goat tracking drones in [204]. But different from it, our system only requires the
tracking information of the animals on the boundary of the herd. This definitely
relaxes the drones' workload, and many existing image processing techniques such
as edge detection can be adopted in real-time.

A drone of the second type is attached with a speaker that plays herding dogs'
barking. The speaker should have a clear voice, abundant volume, relatively small
size, and low weight. Moreover, the speaker is designed to be mounted on a stabilizer
attached to the drone, so that it can stably broadcast to the desired direction, no
matter which direction the barking drone is moving towards. It is worth mentioning
that the speaker on the current human-piloted barking drone is not mounted on the
stabilizer, so we improved the design of the current barking drones.

Different from the observer, the barking drones need to fly at a relatively low
altitude to herd animals. The observer also acts as the controller of the barking
drones. The communications between them can be realized by radio waves. A typical
application scenario of the proposed system is herding a large group of animals with one observer and multiple barking drones. Figure 6.1 shows a schematic diagram of the drone herding system, with one operator and two barking drones. Limited battery-life is always a problem of drones applications. Later we will show that the proposed herding system can usually accomplish the herding task in less than 30 minutes, which is the common endurance of some commercialised industrial drone products such as the DJI M300.

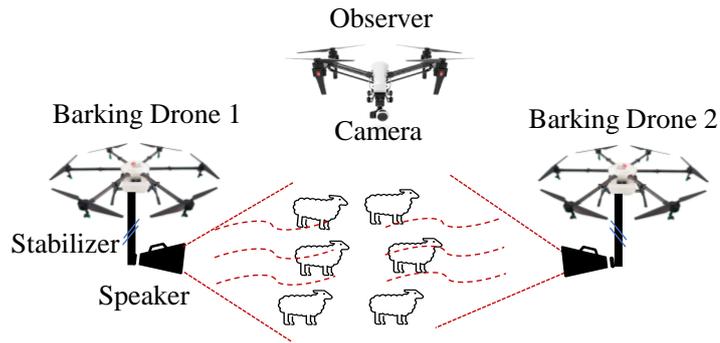

Figure 6.1: Illustration of the proposed drone herding system.

## 6.3 System Model and Problem Statement

In this section, we introduce the dynamics models of animal motion and drone motion. Then, we present the herding problem statement and the preliminaries for designing drones motion controller.

### 6.3.1 Animal Motion Dynamics

Same as most studies on automated herding [20, 203, 210], we describe the dynamic of animal flocking by using the boids model based on Reynolds' rules [211]. Field tests in [199] also show that the boids model matches the behavior of the real sheep. Specifically, let unit vector $\vec{\aleph}_d$ be the repulsive force from the barking drone to the animal. Let $\vec{\aleph}_L$ be the attractive force to the center of mass of the animal's neighbours. Let $\vec{\aleph}_r$ be the repulsive force from other animals. Let $\vec{\aleph}_i$ be the inertial force to remain at the current location. Let $\vec{\aleph}_e$ be the noise element of the animal's movement. Then, the animal's moving direction vector $\vec{\aleph}$ is obtained by

$$\vec{\aleph} = \eta_d \vec{\aleph}_d + \eta_L \vec{\aleph}_L + \eta_r \vec{\aleph}_r + \eta_i \vec{\aleph}_i + \eta_e \vec{\aleph}_e, \tag{6.1}$$

where $\eta_d$, $\eta_L$, $\eta_r$, $\eta_i$, and $\eta_e$ are the weighting constants.

## 6.3.2 Drone Motion Dynamics

In this work, we assume the barking drones maintain at a fixed altitude. The altitude should be higher than the animals' height to avoid collisions with them. Meanwhile, the altitude should be relatively low to keep the barking drones close to the herding animals. With the fixed altitude, we study the 2D motion of a barking drone described by the following mathematical model. Let

$$\boldsymbol{d}(t) := [x(t), y(t)] \tag{6.2}$$

be the 2D vector of the barking drone's Cartesian coordinates. Then, the motion of the barking drone is described by the equations:

$$\dot{\boldsymbol{d}}(t) = v(t)\boldsymbol{a}(t), \tag{6.3}$$

$$\dot{\boldsymbol{a}}(t) = \boldsymbol{u}(t), \tag{6.4}$$

where $\boldsymbol{a}(t) \in \mathbb{R}^2$, $|\boldsymbol{a}(t)| = 1$ for all $t$, $\boldsymbol{u}(t) \in \mathbb{R}^2$, $v(t) \in \mathbb{R}$, and the following constraints hold:

$$|\boldsymbol{u}(t)| \leq U_{max}, \quad v(t) \in [0, V_{max}], \tag{6.5}$$

$$(\boldsymbol{a}(t), \boldsymbol{u}(t)) = 0, \tag{6.6}$$

for all $t$. Here $|\cdot|$ denotes the standard Euclidean vector norm, and $(\cdot, \cdot)$ denotes the scalar product of two vectors. The scalar variable $v(t)$ is the speed or linear velocity of the barking drone, and the scalar $\boldsymbol{u}(t)$ is applied to change the direction of the drone's motion, given by $\boldsymbol{a}(t)$. $v(t)$ and $\boldsymbol{u}(t)$ are two control inputs in this model. $U_{max}$ and $V_{max}$ are constants depending on the manufacturing of the drone. The condition (6.6) guarantees that the vectors $\boldsymbol{a}(t)$ and $\boldsymbol{u}(t)$ are always orthogonal. Furthermore, $\dot{\boldsymbol{d}}(t)$ is the velocity vector of the barking drone. The kinematics of many unmanned aerial vehicles can be described by the non-holonomic model (6.3)-(6.6); see, e.g. [127] and references therein.

### 6.3.3 Problem Statement

This chapter concerns the problem of navigating a network of barking drones to herd a group of farm animals. A typical herding task consists of gathering and driving. In detail, we aim to navigate the barking drones to collect a group of farm animals when they are too dispersed, namely gathering, and drive them to a designated location once they are aggregated, namely driving.

### 6.3.4 Preliminaries

We now introduce the preliminaries for presenting the drones motion control algorithms, including the system's available measurement and the drones' motion restriction. During the gathering, we use the convex hull of all the herding animals to describe the animal flock. Let $\mathcal{D} = \{\boldsymbol{d}_j\}, j = 1, ..., n_d$ be the sets of the 2D positions of $n_d$ barking drones. Let $n_s$ be the number of herding animals. Let $C_o \in \mathbb{R}^2$ denote the position of the herding animals' centroid.

We assume that at any time $t$, the observer has the measurements of the positions of the vertices of the convex hull of all the herding animals, described by $\mathcal{P} = \{P_i\}, i = 1, ..., n_p$, and $n_p$ is the number of vertices of the convex hull. Besides, we assume the observer can estimate $C_o$ by image processing techniques. The accurate real-time locations of the barking drones should also be available. In practice, the real-time drone locations can be provided by embedded GPS chips since the pastures are often open-air.

**Definition 6.3.1.** The extended hull is a unique polygon that surrounds the convex hull. The edges of the extended hull and the convex hull are in one-to-one correspondence, with each pair of the corresponding edges parallel to each other and maintaining the same fixed distance $d_s$.

Let $\mathcal{E} = \{E_i\}, i = 1, ..., n_e$ be the set of 2D positions of all the extended hull's vertices in a counterclockwise manner, $n_e = n_p$. $\mathcal{E}$ can be calculated from $\mathcal{P}$ by simple geometry method.

**Motion Restriction:** The animals can be dispersed if a drone is too close to them (e.g. within animals' convex hull). Besides, any drones' trajectory outside the extended hull will be longer than the trajectory on the extended hull with the same start, end and direction. Thus, for efficient gathering and avoiding dispersing, all the barking drones are restricted to move only on the extended hull during the gathering.

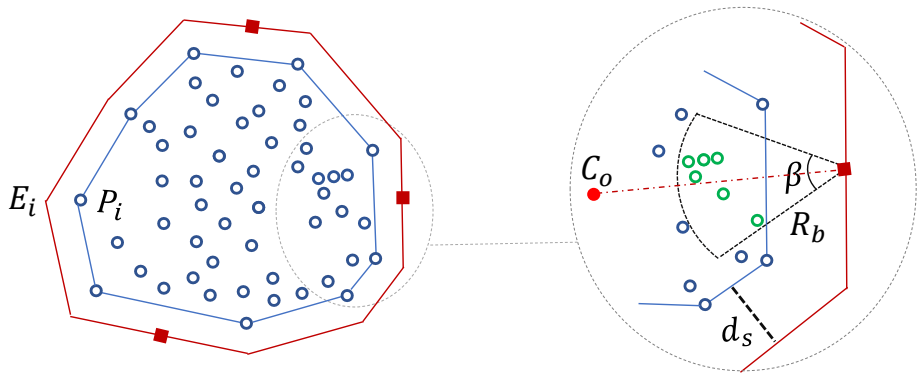

Figure 6.2: Illustration of the extended hull with the drone-to-animal distance $d_s$; the barking cone with effective broadcasting angle $\beta$ and distance $R_b$. Here (○) stands for the animal repulsing by the barking drone (■); (●) stands for $C_o$.

We assume that the spread range of the barking from the drone is fan-shaped, and only animals within this range will be affected by the repulsion of the barking drone. We call this fan-shaped range as the barking cone, with the effective broadcasting angle $\beta$ and distance $R_b$. With the help of the stabilizer, the speaker should always face to $C_o$, as illustrated in Figure 6.2.

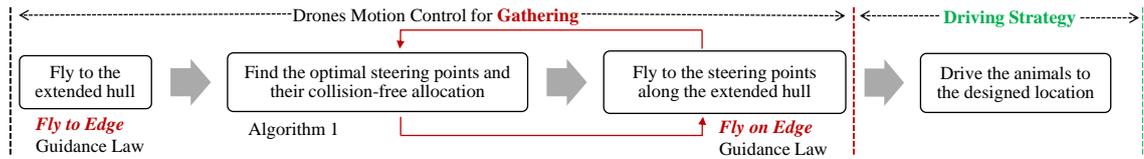

Figure 6.3: Overview of the proposed method.

## 6.4 Drones Motion Control

This section introduces the motion control algorithms for barking drones to accomplish the gathering and driving tasks. We first introduce the algorithm for navigating the barking drones to fly to the extended hull and to fly on the extended hull in Section 6.4.A and Section 6.4.B, respectively. Then, Section 6.4.C presents the optimal positions (steering points) and their collision-free allocation for the barking drones to gather animals efficiently. Finally, Section 6.4.D briefly introduces the driving strategy. A flowchart of the proposed method is shown in Figure 6.3.

**Remark 6.4.1.** The barking drones should only bark at outside of the convex hull to avoid dispersing any herding animals.

Let $A$ be any point on the plane of the extended hull. Let $B$ be a vertex of the extended hull. We now introduce two guidance laws for navigating a barking drone from $A$ to $B$ in the shortest time:

1. **Fly to edge**: Navigate the barking drone from an initial position $A$ to the extended hull. Note that the vertices of the extended hull can be moving. Let $O$ denote the barking drone's reaching point on the extended hull.

2. **Fly on edge**: Navigate the barking drone from $O$ to $B$ following a given direction, e.g., clockwise or counterclockwise, while keeping the barking drone on the extended hull.

### 6.4.1 Fly to Edge Guidance Law

Let $\boldsymbol{w}_1$ and $\boldsymbol{w}_2$ be non-zero 2D vectors, and $|\boldsymbol{w}_1| = 1$. Now introduce the following function $F(\cdot, \cdot)$ mapping from $\mathbb{R}^2 \times \mathbb{R}^2$ to $\mathbb{R}^2$ as

$$\text{F}(\boldsymbol{w}_1, \boldsymbol{w}_2) := \begin{cases} 0, & f(\boldsymbol{w}_1, \boldsymbol{w}_2) = 0, \\ |f(\boldsymbol{w}_1, \boldsymbol{w}_2)|^{-1} f(\boldsymbol{w}_1, \boldsymbol{w}_2), & f(\boldsymbol{w}_1, \boldsymbol{w}_2) \neq 0, \end{cases} \tag{6.7}$$

where $f(\boldsymbol{w}_1, \boldsymbol{w}_2) := \boldsymbol{w}_2 - (\boldsymbol{w}_1, \boldsymbol{w}_2)\boldsymbol{w}_1$. In other words, the rule (6.7) defined in the plane of vectors $\boldsymbol{w}_1$ and $\boldsymbol{w}_2$. The resulted vector $F(\boldsymbol{w}_1, \boldsymbol{w}_2)$ is orthogonal to $\boldsymbol{w}_1$ and directed "towards" $\boldsymbol{w}_2$. Moreover, introduce the function $g(\boldsymbol{w}_1, \boldsymbol{w}_2)$ as follows

$$g(\boldsymbol{w}_1, \boldsymbol{w}_2) := \begin{cases} 1, (\boldsymbol{w}_1, \boldsymbol{w}_2) > 0, \\ -1, (\boldsymbol{w}_1, \boldsymbol{w}_2) \leq 0, \end{cases} \tag{6.8}$$

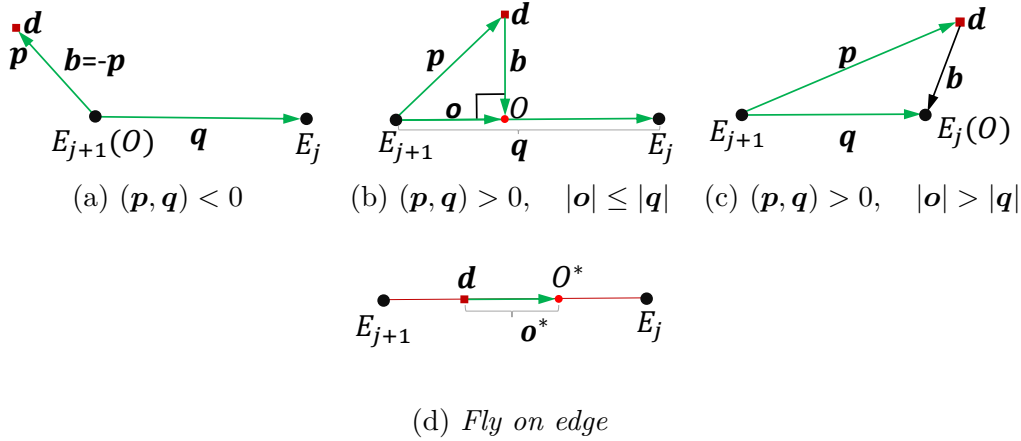

(a) $(\boldsymbol{p}, \boldsymbol{q}) < 0$     (b) $(\boldsymbol{p}, \boldsymbol{q}) > 0, \quad |\boldsymbol{o}| \leq |\boldsymbol{q}|$     (c) $(\boldsymbol{p}, \boldsymbol{q}) > 0, \quad |\boldsymbol{o}| > |\boldsymbol{q}|$

(d) *Fly on edge*

Figure 6.4: Illustration of (a-c) *Fly to edge* guidance with $\boldsymbol{b}$ in different cases; (d) *Fly on edge* guidance navigates the drone from $\boldsymbol{d}$ to $O^*$.

We will also need the following notations to present the *Fly to edge* guidance law. At time $t$, let $E_{j+1}E_j$ be the extended hull edge that is the closest to the drone. Let $\boldsymbol{q}(t) \in \mathbb{R}^2$ denote the vector from vertex $E_{j+1}$ to $E_j$. Let $\boldsymbol{p}(t) \in \mathbb{R}^2$ denote the vector from the vertex $E_{j+1}$ to the drone. Let $O$ be the point on $E_{j+1}E_j$ that is the closest to the drone. Let $\boldsymbol{b}(t)$ be the vector from the drone to $O$. If $(\boldsymbol{p}, \boldsymbol{q}) < 0$, we have $O = E_{j+1}$ and $\boldsymbol{b}(t) = -\boldsymbol{p}(t)$, see Figure 6.4a. Let $\boldsymbol{o}(t)$ be the vector from $E_{j+1}$ to $O$. If $(\boldsymbol{p}, \boldsymbol{q}) > 0$ and $|\boldsymbol{o}| \leq |\boldsymbol{q}|$, $\boldsymbol{b}(t)$ is always orthogonal to $\boldsymbol{q}(t)$, see Figure 6.4b.

$\boldsymbol{o}(t)$ can be obtained by the following equations:

$$|\boldsymbol{o}(t)| = |\boldsymbol{q}(t)|^{-1}(\boldsymbol{p}(t), \boldsymbol{q}(t)), \tag{6.9}$$

$$\boldsymbol{o}(t) = |\boldsymbol{o}(t)| \cdot |\boldsymbol{q}(t)|^{-1}\boldsymbol{q}(t), \tag{6.10}$$

Otherwise, we have $O = E_j$ and $\boldsymbol{b}(t) = \boldsymbol{q}(t) - \boldsymbol{p}(t)$, see Figure 6.4c. Given $\boldsymbol{p}(t)$ and $\boldsymbol{q}(t)$, we present the following *Fly to edge* guidance law:

$$\boldsymbol{b}(t) = \begin{cases} -\boldsymbol{p}(t), & \text{if } (\boldsymbol{p}, \boldsymbol{q}) < 0, \\ \boldsymbol{o}(t) - \boldsymbol{p}(t), & \text{if } (\boldsymbol{p}, \boldsymbol{q}) > 0 \text{ and } |\boldsymbol{o}| \leq |\boldsymbol{q}|, \\ \boldsymbol{q}(t) - \boldsymbol{p}(t), & \text{otherwise}, \end{cases} \tag{6.11}$$

$$\boldsymbol{u}(t) = U_{max}g(\boldsymbol{a}(t), \boldsymbol{b}(t))F(\boldsymbol{a}(t), \boldsymbol{b}(t)), \tag{6.12}$$

$$v(t) = V_{max}g(\boldsymbol{a}(t), \boldsymbol{b}(t)). \tag{6.13}$$

Obviously, equation (6.11) gives that $\boldsymbol{b}(t)$ is pointing from the drone to its closest point on $E_{j+1}E_j$, see Figure 6.4 (a-c). Moreover, equations (6.12) and (6.13) adjust the heading of the drone towards the target location pointed to by $\boldsymbol{b}(t)$ with the maximum angular speed and guides the drone towards such a location with maximum linear speed. The proposed *Fly to edge* guidance law belongs to the class of sliding-mode control laws (see, e.g., [139, 212, 213] as well as the class of switched control laws (see e.g. [214, 215]).

**Remark 6.4.2.** At time $t$, given $\boldsymbol{d}(t)$ and $E$, calculate $\boldsymbol{b}(t)$ for the barking drone to each edge of the extended hull. Then, the edge with the minimum $|\boldsymbol{b}(t)|$ is the closest edge of the extended hull to the drone.

## 6.4.2 Fly on Edge Guidance Law

We now introduce the *Fly on edge* guidance law for a drone flying along an edge of the extended hull, with possibly moving vertices. At time $t$, let $E_{j+1}E_j$ be the edge that we want to keep the drone on. Let $O^* \in E_j E_{j+1}$ be the target position of the barking drone. Let $\boldsymbol{o}^*(t) \in \mathbb{R}^2$ denote the vector from the drone to $O^*$, as shown in Figure 6.4d. We introduce $\boldsymbol{b}^*(t)$ that is given by:

$$\boldsymbol{b}^*(t) = \boldsymbol{b}(t) + \boldsymbol{o}^*(t). \tag{6.14}$$

Then, the *Fly on edge* guidance law is as follows:

$$\boldsymbol{u}(t) = U_{max} g(\boldsymbol{a}(t), \boldsymbol{b}^*(t)) F(\boldsymbol{a}(t), \boldsymbol{b}^*(t)), \tag{6.15}$$

$$v(t) = V_{max} g(\boldsymbol{a}(t), \boldsymbol{b}^*(t)). \tag{6.16}$$

Note that, $\boldsymbol{b}^*$(t) consists of two vector components: $\boldsymbol{b}$(t) and $\boldsymbol{o}^*$(t). Where $\boldsymbol{b}$(t) is for keeping the drone on $E_{j+1}E_j$ and $\boldsymbol{o}^*$(t) is for navigating the drone to $O^*$. Thus, the guidance law (6.14) (6.15) and (6.16) navigates the barking drone from $\boldsymbol{d}(t)$ to $O^*$ along $E_{j+1}E_j$, and enables the drone to stay at $O^*$. The presented guidance law is designed to navigate the barking drone from any point on the extended hull to a selected vertex following a given direction, and stop the drone at the selected vertex. To this end, the drone may fly to the vertex of the adjacent edge in the given direction multiple times following the *Fly on edge* guidance law, until reaching the selected vertex.

## 6.4.3   Collision-free Allocation of Steering Points

We now find the optimal positions for the barking drones to effectively gather animals. Aiming to minimize the maximum animal-to-centroid distance in the shortest time, at any time $t$, we choose the animals with the largest animal-to-centroid distance as the target animals. These animals are also the convex hull vertices that are farthest to $C_o$. Since the barking drones have their motions restricted on the extended hull, we select the extended hull vertices corresponding to the target animals as the optimal drone positions for steering the target animals to approach $C_o$.

From now on, we call these corresponding extended hull vertices the steering points, denoted by the set $\mathcal{S} = \{s_j\}, j = 1, ..., n_d, \mathcal{S} \subseteq \mathcal{E}$.

**Definition 6.4.1.** The allocation of steering points specifies which drone goes to which steering point through which direction, i.e., clockwise or counterclockwise.

The steering points are computed and allocated by the observer drone. The optimal allocation of steering points should meet the following two requirements:

1. No collision happens when each drone is flying to its allocated steering point along the extended hull.

2. With requirement 1) met, the maximum travel distance of the drones is minimized.

Suppose that all the drones have arrived at the extended hull at time $t = t_1$. We relabel the drones so that the index of the drones increases in the counterclockwise direction. Let $M$ be the perimeter of the extended hull. Imagine that we disconnect the extended hull from the position of the first drone, i.e. $d_1(t)$. Then, 'straighten' the extended hull into a straight line segment with a length of $M$, so that $\mathcal{D}$, $\mathcal{E}$ and $\mathcal{S}$ become the points on the line segment. Based on this line segment, we build a one-dimensional (1D) coordinate axis denoted as the $z$ axis. Let $\mathcal{Z} = \{z_j\}, j = 1, ..., n_d$ be the 1D coordinates of the drones' positions on the $z$ axis. Let $z_1 = 0$ be the origin. We have $z_j < z_{j+1}, j = 1, ..., n_d - 1$, as shown in Figure 6.5a. It can be seen that the left and right flying on the $z$ axis corresponding to counterclockwise and clockwise flying on the extended hull, respectively.

We will also need the following notations to present our algorithm. Let $\mathcal{S}' = \left\{s'_j\right\}, j = 1, ..., n_d$ be a set of *allocated* steering points with a corresponding $z$ axis coordinates set $\mathcal{Z}' = \left\{z'_j\right\}, j = 1, ..., n_d$, as shown in Figure 6.5b. $\mathcal{Z}'$ is the destination of the drones on the $z$ axis. Note that, $z'_j < z'_{j+1}, j = 1, ..., n_d - 1$ may not hold. Let $\Gamma = \{\gamma_j\}, j = 1, ..., n_d$ be the set of the drones' travel distances for reaching their allocated steering points. We now define three variables $\sigma_j$, $\lambda^R_j$ and $\lambda^L_j \in \{0, 1\}$ to indicate the flying direction and extent of drone $j$. Specifically, let $\sigma_j = 1$ if drone $j$ reaches $z_j'(t)$ by right flying on the $z$ axis, and $\sigma_j = 0$ if drone $j$ reaches $z_j'(t)$ by left flying on the $z$ axis. Furthermore, let $\lambda_j^R = 1$ if drone $j$ will pass $z = 0$ by right flying to reach $z_j'(t)$, and $\lambda_j^R = 0$ otherwise. Similarly, let $\lambda_j^L = 1$ if drone $j$ will pass $z = 0$ by left flying to reach $z_j'(t)$, and $\lambda_j^L = 0$ otherwise. Let $\Sigma = \{\sigma_j\}$, $\Lambda^L = \left\{\lambda_j^L\right\}$ and $\Lambda^R = \left\{\lambda_j^R\right\}$, $j = 1, ..., n_d$ be the sets of $\sigma_j$, $\lambda_j^L$ and $\lambda_j^R$, respectively. Given $z_j$, $z_j'$ and $\sigma_j$, $\lambda_j^L$ and $\lambda_j^R$ can be computed by:

$$\lambda_j^L = \begin{cases} 1, & \text{if } z_j' > z_j \text{ and } \sigma_j = 0, \\ 0, & \text{otherwise}, \end{cases} \tag{6.17}$$

$$\lambda_j^R = \begin{cases} 1, & \text{if } z_j' < z_j \text{ and } \sigma_j = 1, \\ 0, & \text{otherwise}. \end{cases} \tag{6.18}$$

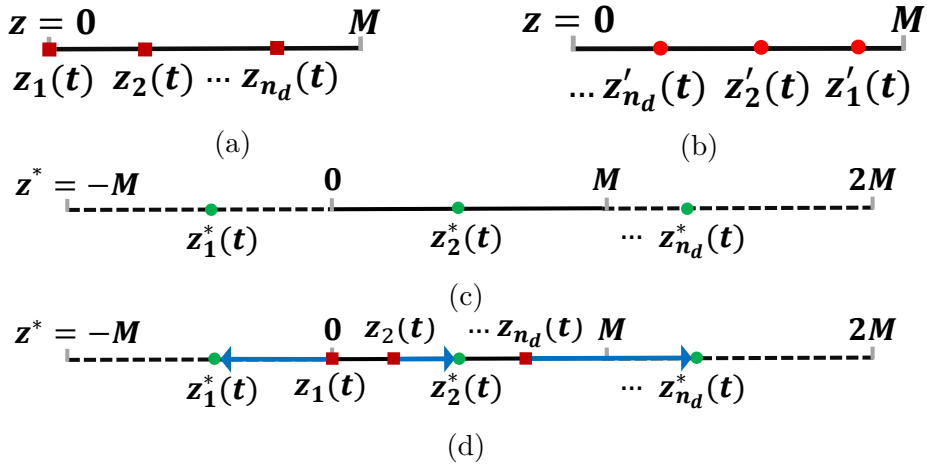

Figure 6.5: The examples of (a) drones' positions on the $z$ axis, (b) steering points' positions on the $z$ axis, (c) steering points on the $z^*$ axis, and (d) drone's positions, steering points' positions and travel routes on the $z^*$ axis. Here (■) stands for $z_j$, (●) stands for $z_j'$, (●) stands for $z_j^*$, and the blue arrows stand for the drones' travel routes.

The main notations are listed in Table 6.1. Since the line segment $z = [0, M]$ is generated by straightening the enclosed extended hull, the drones passed $z = 0$ by left flying will appear on the right side of the line segment, and the drones passed $z = M$ by right flying will appear on the left side of the line segment. We now imagine extending the line segment $z = [0, M]$ to $z = [-M, 2M]$ and build another 1D coordinate $z^*$ axis, as shown in Figure 6.5c. On the $z^*$ axis, the drones passed $z^* = 0$ by left flying will appear on $z^* = [-M, 0]$, and the drones passed $z^* = M$ by right flying will appear on $z^* = [M, 2M]$. Let $\mathcal{Z}^* = \left\{ z^*_j \right\}, j = 1, ..., n_d$ be the 1D coordinates set on the $z^*$ axis corresponding to $\mathcal{Z}'$. Then, the mapping between $\mathcal{Z}^*$ and $\mathcal{Z}'$ is obtained by:

$$z^*_j = \begin{cases} z'_j - \lambda^L_j M, & \text{if } \sigma_j = 0, \\ z'_j + \lambda^R_j M, & \text{if } \sigma_j = 1, \end{cases} \qquad (6.19)$$

Table 6.1: Notations and Descriptions

| Notation | Description |
|---|---|
| $M$ | Perimeter of the extended hull. |
| $\mathcal{D} = \{\boldsymbol{d}_j\}$ | Set of the drones' 2D positions. |
| $\mathcal{Z} = \{z_j\}$ | Set of the $z$ coordinates of $\mathcal{D}$. |
| $\mathcal{S} = \{\boldsymbol{s}_j\}$ | Set of the steering points' 2D positions. |
| $\mathcal{S}' = \{\boldsymbol{s}'_j\}$ | Set of the *allocated* steering points' 2D positions. |
| $\mathcal{Z}' = \{z'_j\}$ | Set of the $z$ coordinates of $\mathcal{S}'$. |
| $\mathcal{Z}^* = \{z^*_j\}$ | Set of the $z^*$ coordinates of $\mathcal{Z}'$. |
| $\Gamma = \{\gamma_j\}$ | Set of drones' travel distances . |
| $\Sigma = \{\sigma_j\}$ | Set of drones' flying directions. |
| $\Lambda^L = \{\lambda^L_j\}$ | Set of the indicators of passing $z = 0$ by right flying. |
| $\Lambda^R = \{\lambda^R_j\}$ | Set of the indicators of passing $z = 0$ by left flying. |

If we place $\mathcal{Z}$ on the $z^*$ axis, as shown in Figure 6.5c, the travel route of any drone $j$ will be $\overrightarrow{z_j z^*_j}$. We obtain the expression for the travel distances $\gamma_j$ as follows:

$$\gamma_j = \begin{cases} z_j - z^*_j, & \text{if } \sigma_j = 0, \\ z^*_j - z_j, & \text{if } \sigma_j = 1, \end{cases} \qquad (6.20)$$

Then, the steering points allocation optimization problem is formulated as follows:

$$\min_{\mathcal{S}', \Sigma} \; \max_{j=1,...,n_d} \; \gamma_j, \qquad (6.21)$$

s.t.

$$0 < z_{n_d}^* - z_1^* < M, \tag{6.22}$$

$$z_j^* < z_{j+1}^*, \; j = 1, ..., n_d - 1, \tag{6.23}$$

where (6.21) minimizes the travel distance of the drone farthest to its allocated
steering point.

**Assumption 6.4.1.** All the drones start flying to their allocated steering points at
the same time, follow the proposed *Fly on edge* guidance law.

**Theorem 6.4.1.** Suppose that Assumptions 6.4.1 holds. Then, (6.22) and (6.23)
guarantees that no collision happens when the drones are flying to their allocated
steering points.

**Proof 4.** Suppose that all the drones start flying to their steering points at time
$t = t_0$. Let $t = t_f^j$ be the time of drone $j$ arrives at its steering point, i.e. $z_j^*(t)$.
From (6.8) and (6.16), at any time $t = t_s \in [t_0, t_f^j]$, drone $j \in \{2, ..., n_d - 1\}$ has:

$$|v(t_s)| = V_{max}. \tag{6.24}$$

As mentioned in Section 6.4, $\boldsymbol{b}(t)$ is always being minimized after drone $j$ arrived
at the extended hull. Since drone $j$ is moving from $z_j(t_0)$ to $z_j^*$ along the $z$ axis, it
can be obtained from (6.3), (6.14) and (6.16) that

$$\boldsymbol{b}^*(t_s) = z_j^* - z_j(t_s), \tag{6.25}$$

$$z_j(t_s) = z_j(t_0) + (t_s - t_0)V_{max}, \tag{6.26}$$

$$z_j^* = z_j(t_0) + (t_f^j - t_0)V_{max}. \tag{6.27}$$

Then, the distance between drone $j$ and drone $j+1$ at time $t = t_s \in [t_0, \max{(t_f^j, t_f^{j+1})}]$

can be computed by

$$z_{j+1}(t_s) - z_j(t_s) =$$

$$\begin{cases} z_{j+1}(t_0) - z_j(t_0), & \text{if } t_s \leq \min(t_f^j, t_f^{j+1}), \\ z_{j+1}(t_s) - z_j^*, & \text{if } t_f^j \leq t_f^{j+1} \text{ and } t_f^j < t_s \leq t_f^{j+1}, \quad (6.28) \\ z_{j+1}^* - z_j(t_s), & \text{if } t_f^j > t_f^{j+1} \text{ and } t_f^{j+1} < t_s \leq t_f^j. \end{cases}$$

Since $z_j < z_{j+1}, j = 1, ..., n_d - 1$, it can be concluded from (6.23), (6.26), (6.27) and (6.28) that

$$z_{j+1}(t_s) - z_j(t_s) > 0, \quad t_s \in [t_0, \max(t_f^j, t_f^{j+1})] \quad (6.29)$$

Which means drone $j \in \{2, ..., n_d - 1\}$ will not collide with drone $j + 1$ before they arrived at their steering points. Moreover, the actual distance $|z_1 z_{n_d}|$ between drone 1 and drone $n_d$ is given by

$$|z_1 z_{n_d}| =$$

$$\begin{cases} z_{n_d}(t_s) - z_1(t_s), & \text{if } z_{n_d}(t_s) - z_1(t_s) \leq M/2, \\ M - (z_{n_d}(t_s) - z_1(t_s)), & \text{if } z_{n_d}(t_s) - z_1(t_s) > M/2. \end{cases} \quad (6.30)$$

Given (6.22), $|z_1 z_{n_d}| > 0, t_s \in [t_0, \max(t_f^1, t_f^{n_d})]$ can be proved similarly. Therefore, (6.22) guarantees that drone 1 will not collide with drone $n_d$, and (6.23) guarantees that each drone will not collide with their neighbors. This completes the proof of Theorem 6.4.1.

For $n_d$ drones, $n_d$ steering points and two possible directions for each drone, the number of possible allocations is $N = n_d! 2^{n_d}$. Since $n_d$ is often a limited number, $N$ will be limited as well. Therefore, the optimal allocation can be found by generating and searching all the possible allocations. We are now in a position to present the algorithm to find the optimal steering points allocation, as shown in Algorithm 3:

Suppose that the gathering task starts at $t = 0$. The proposed herding system first navigates all the barking drones to the extended hull by *Fly to edge* guidance law. Then, the system calculates the optimal steering points allocation after every

---

**Algorithm 3** Optimal Steering Points Allocation.

---

**Input:** $n_d$, $\mathcal{D}$, $\mathcal{E}$
 1: Find $\mathcal{S} \in \mathcal{E}$;
 2: Calculate $\mathcal{Z}$ from $\mathcal{D}$ and $\mathcal{E}$;
 3: Generate possible allocations $(\mathcal{S}', \Sigma)^k$, $k = 1, ..., N$;
 4: For each $(\mathcal{S}', \Sigma)^k$, calculate $(\mathcal{Z}^*, \Gamma)^k$;
 5: Solve (6.21)-(6.23) by searching $(\mathcal{Z}^*, \Gamma)^k$, $k = 1, ..., N$. =0

---

sampling time $\Delta$ and navigates the barking drones to their allocated steering points by *Fly on edge* guidance law, until the distance between $C_o$ and any animal reaches a predefined constant $R_c$. It is worth mentioning that, due to the movement of animals, the optimal allocation may change before some drones reach their assigned steering points. The gathering task, however, will not be interrupted. Because as long as the barking drones are flying on the extended hull, the animals inside the drones' barking cone will be repulsed to move towards $C_o$.

## 6.4.4 Driving Strategy

After gathering, the goal is then transferred to drive the gathered animals to the desired location, e.g., a sheep yard. During driving, we use the smallest enclosing circle to describe the footprint of the gathered animals. Similar to the extended hull definition, we define the *extended circle* as a circle with a larger radius and share the same center of the smallest enclosing circle. We adopt a side-to-side movement for the baking drones, which is a common animal driving strategy that can also be seen in [199], etc. Let $\mathcal{L}$ be the semicircle of the extended circle that is farther to $G$. Let $\mathcal{Q} = \{Q_j\}$, $j = 1, ..., n_d + 1$ be the set of points that *evenly* distributed on $\mathcal{L}$. Each drone $j$ is then deployed to fly on $\mathcal{L}$ between $Q_j$ and $Q_{j+1}$. Besides, $\mathcal{L}$ is set to moving towards $G$ with a constant speed $V_{driving} \leq V_{animal}$ as the driving speed, as shown in Figure 6.6.

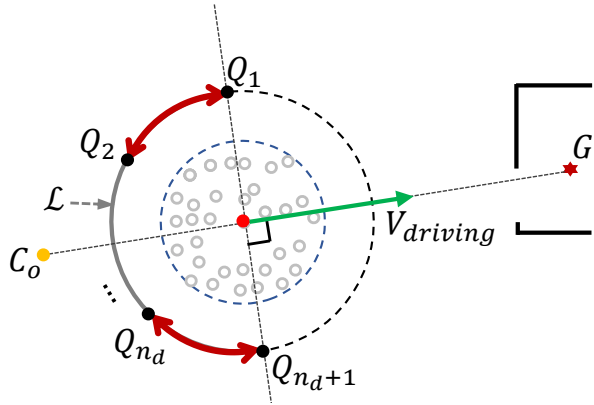

Figure 6.6: Barking drones deployment for animal driving, where (✱) stands for the designated location $G$; The dark red arrows stand for the drones' side-to-side trajectories.

## 6.5 Simulation Results

In this section, the performance of the proposed method is evaluated using MATLAB. Each simulation runs for 20 times. The animal motion dynamic parameters are chosen based on the field tests with real sheep conducted by [199], as shown in Table 6.2. Table 6.2 also shows some parameters of the barking drones, if not specified in the following part.

Table 6.2: Simulation Parameter Values

| Parameters | Values | Parameters | Values |
|------------|--------|------------|--------|
| $\eta_d$ | 1 | $\eta_L$ | 1.05 |
| $\eta_r$ | 2 | $\eta_i$ | 0.5 |
| $\eta_e$ | 0.3 | $\Delta$ | 0.2 s |
| $V_{max}$ | 25 m/s | $U_{max}$ | 5 m/s$^2$ |
| $V_{animal}$ | 4 m/s | $\beta$ | $\frac{2\pi}{3}$ |
| $R_b$ | 100 m | $d_s$ | 30 m |
| $n_s$ | 200, 1000 | $R_c$ | 60, 110 m |
| $V_{driving}$ | 3.8 m/s, 1.9 m/s | $n_d$ | 4 |

**Benchmark for Comparison:** Previously published studies on automated herding have not dealt with a large number of animals. For comparison, we introduce an intuitional collision-free method as the benchmark method. Specifically, the benchmark method divides the extended hull into $n_d$ segments with the same length at any time during the gathering. Each drone is allocated to a segment and does the aforementioned side-to-side movement on the extended hull, until finished gathering. The benchmark method adopts the same driving strategy as the proposed method.

We consider that the animals are randomly distributed in an area with a size of 1200 m by 600 m as the initial field. The drones are initially lined up at 80 m right to the initial field with a 50 m mutual distance, as showing in Figure 6.7a. We first present some illustrative results showing 4 barking drones on two cases herding 200 and 1000 animals, respectively; see https://youtu.be/KMWxrlkU6t0 and https://youtu.be/KPGrAcgPH8Q. We can observe that the proposed method completes the gathering task in 11.1 minutes for the instance with 200 animals and 10.1 minutes for the case with 1000 animals. The total time for gathering and driving is 15 and 18.2 minutes for these cases. Note that, here the time refers to the time for finishing the herding tasks in the simulated environment, rather than the execution time of the computer. However, the benchmark method uses about 3.3 and 4.1 more minutes to complete these missions. Figure 6.7a shows how the animal, footprint radius changes with time $t$ for these cases. We also present snapshots of $t = 0$, $t = 5$ and $t = 8$ minutes for the case of 1000 animals in Figures. 6.7(b-f).

Interestingly, Figure 6.7a shows that the time difference between gathering 200 animals and 1000 animals by the proposed method and the benchmark method is not so obvious. The proposed method, however, can always use less time to complete the gathering mission. This is because the proposed method always chases and repulses the animals that are farthest to the center, while the benchmark method is repulsing the animals indiscriminately. Therefore, the animals' footprint with the proposed method becomes increasingly round-like during shrinking, while animals' footprint with the benchmark method becomes long and narrow. This fact can be observed by comparing Figures. 6.7c and 6.7e, or by comparing Figures. 6.7d and 6.7f.

Note that, the time consumption of flying to the edge and the driving task mainly depends on the initial locations of the drones and the animals. From now on, we focus on evaluating the average gathering time after the drones have arrived on the extended hull. The aforementioned minor difference between herding 200 and 1000 animals is very likely because that the gathering time is strongly correlated with the size of the initial field, rather than the number of animals. To confirm this, we change the initial field into a square and investigate the relationship between the gathering time and the length of the initial square field; see Figure 6.8a. It reveals that the average gathering time increases significantly with the initial square field length. This supports the guess that the gathering time is strongly correlated with the size of the initial field. The reason is also that the gathering time mainly depends on the movement of the animals on edge, and particularly the travelling time for them to move to the area close to $C_o$. With fixed maximum animal speed $V_{animal}$ and the same repulsion from the barking drones, the travelling distances of these animals are dominated by the size of the initial field. Moreover, Figure 6.8a shows that the difference between the gathering time of the benchmark method and the proposed method increases with the initial square field length. It means the benchmark method is more sensitive to the varying length of the initial square field. We further investigate the relationship between the gathering time and the number of barking drones $n_d$; see Figure 6.8b. Not surprisingly, the average gathering time decreases significantly with the increase of $n_d$ for both methods. Besides, Figure 6.8b shows that the superiority of the proposed method becomes more apparent, with $n_d$ increases when $n_d \geq 4$.

Next, we investigate the impact of the drone speed and animal speed on the gathering time; see Figure 6.9. Figure 6.9a shows that slower drones will lead to a higher average gathering time, especially when the maximum drone speed $V_{max} < 15m/s$, for both the benchmark method and the proposed method. Moreover, the average gathering time of the benchmark method is more sensitive to $V_{max}$ when $V_{max} < 15m/s$. In addition, in our simulations, drones with $V_{max} \leq 10m/s$ cannot accomplish the gathering task using the benchmark method. In the implementation of the proposed method, drones with $V_{max} > 15m/s$ is preferable. Furthermore, the average gathering time of the proposed method reduces with $V_{max}$ increases, the percentage of the reduction, however, is not significant when $V_{max} > 30m/s$.

Figure 6.9b shows that animals with higher maximum speed $V_{animal}$ can be gathered in a shorter time. Particularly, with the proposed method, the average gathering time reduces around 38% (from 14.7 minutes to 9.1 minutes) when $V_{animal}$ increases 150% (from $2m/s$ to $5m/s$). For the benchmark method, the average gathering time reduces around 39% (from 20.8 minutes to 12.7 minutes). Therefore, the reduction of average gathering time is much slower than the increase of $V_{animal}$ when $2m/s \leq V_{animal} \leq 5m/s$, for both methods.

Lastly, we investigate the impact of the barking cone radius $R_b$ and the drone-to-animal distance $d_s$ on the gathering time; see Figure 6.10. Figure 6.10a presents the relationship between the barking radius $R_b$ and the gathering time with 200 animals and 1000 animals, respectively. We can observe that increasing $R_b$ will accelerate the gathering when $R_b \leq 100m$. But when $R_b > 100m$, the average gathering time increases with $R_b$, which is contradictory to our expectation. One possible reason is that the gathering time mainly depends on the animals on the edges. If $R_b$ is too large, it may cause mutual interference between the repulsive forces inflicted by the barking drones, which may pull down the gathering. Moreover, Figure 6.10a shows that the proposed method is more sensitive to $R_b$ when gathering more animals with $R_b \leq 100m$. This is because the proportion of the repulsed animals near the edges tend to be less when $n_s$ increases, for a fixed $R_b$.

Figure 6.10b suggests that the average gathering time decreases with $d_s$ increases when $d_s \leq 30m$. One possible reason is that more animals will be repulsed to the directions that do not point to the center if $d_s$ is too small, since the repulsive force from the barking drone points to the opposite of it and the barking zone is fan-shaped. This result is also considered as interference. However, increasing $d_s$ will decelerate the gathering when $d_s > 30m$, and this becomes more obvious with more animals. It is reasonable because increasing $d_s$ is almost equivalent to decreasing $R_b$ when $R_b$ is fixed.

In summary, we present computer simulation results in this section to demonstrate the performance of the proposed method. These results confirm that the proposed method can efficiently herd a large number of farm animals and outperform the benchmark method. By investigating the impact of the system parameters, we can obtain that a higher speed of drones leads to shorter gathering time. The barking cone radius $R_b$ and the drone-to-animal distance $d_s$ also significantly affect the gathering time. The optimal values of them can be obtained via experiments on real-world animals.

## 6.6 Summary

In this chapter, we proposed a novel automated herding system based on autonomous barking drones. We developed a collision-free sliding mode based motion control algorithm, which navigates a network of barking drones to efficiently collect a group of animals when they are too dispersed and drive them to a designated location. Simulations using a dynamic model of animal flocking based on Reynolds' rules showed the proposed drone herding system can efficiently herd a thousand of animals with several drones. A unique contribution of this chapter is the proposal of the first prototype of herding a large flock of farm animals by autonomous drones. The future work is to conduct experiments on real farm animals to test the proposed method.

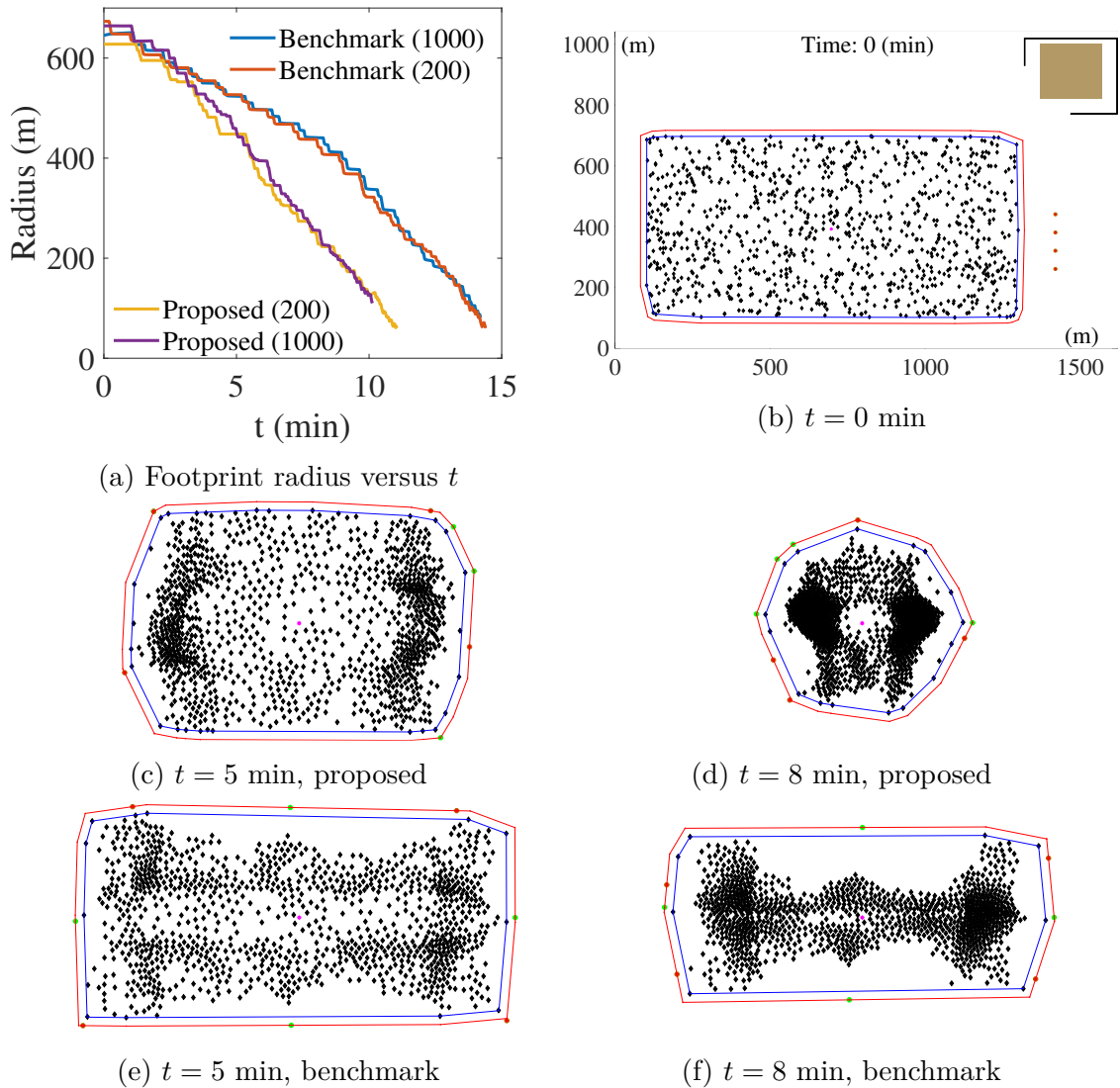

(a) Footprint radius versus $t$

(b) $t = 0$ min

(c) $t = 5$ min, proposed

(d) $t = 8$ min, proposed

(e) $t = 5$ min, benchmark

(f) $t = 8$ min, benchmark

Figure 6.7: (a) Animals' footprint radius versus time $t$ for herding 200 and 1000 animals with 4 barking drones using both methods; videos recording the movements is available online at: https://youtu.be/KMWxrlkU6t0 and https://youtu.be/KPGrAcgPH8Q(b)-(f) snapshots of the 1000 animals at $t = 0, 5, 8$ minutes for both methods

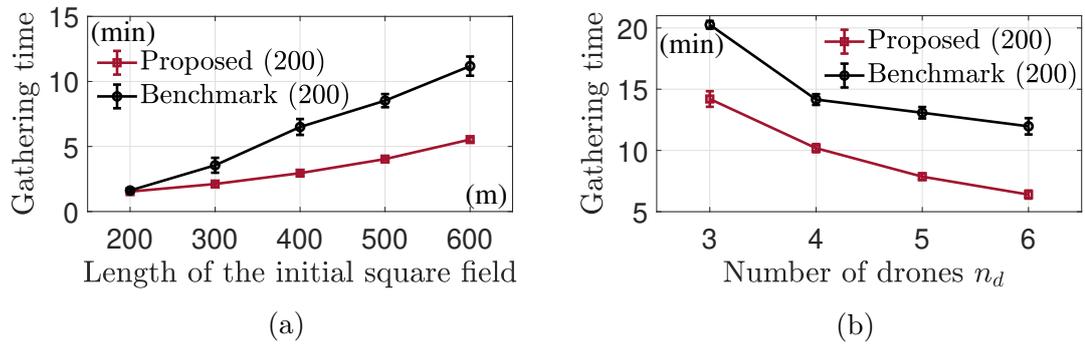

(a)                                    (b)

Figure 6.8: Comparisons of the average gathering time for different values of (a) length of the initial square field; (b) number of barking drones $n_d$.

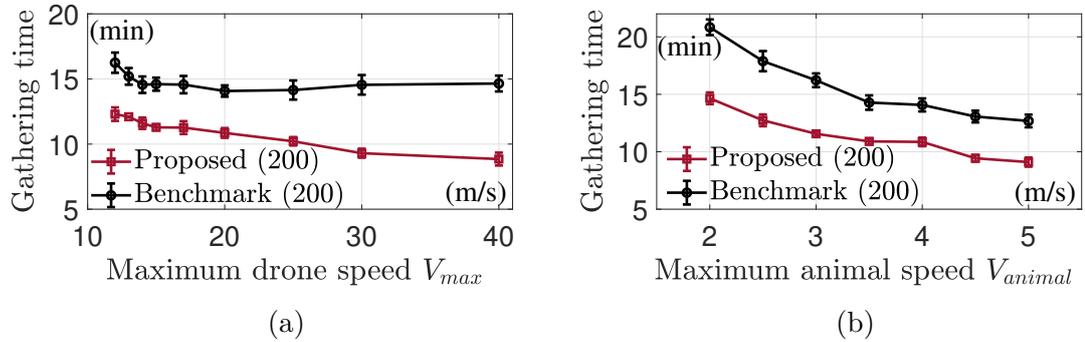

(a)                                    (b)

Figure 6.9: Comparisons of the average gathering time for different values of (a) the maximum drone speed $V_{max}$; (b) the maximum animal speed $V_{animal}$.

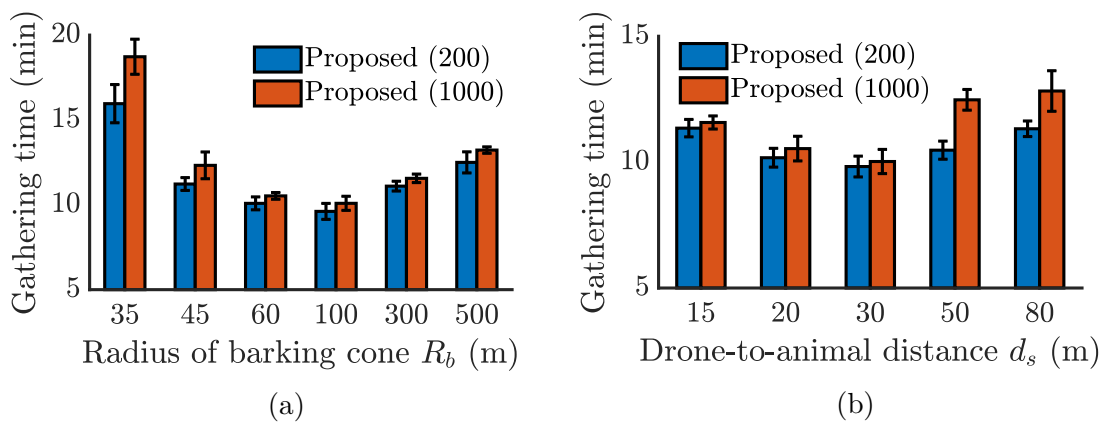

(a)                                    (b)

Figure 6.10: Comparisons of the average gathering time for different values of (a) barking cone radius $R_b$; (b) drone-to-animal distance $d_s$.

# Chapter 7

# Conclusion

This report addressed several challenges of determining aerial drone deployment and navigation to improve system performance for different sensing and interacting applications. We conclude this work by highlighting the contributions.

- **Autonomous Navigation of an Aerial Drone to Observe a Group of Wild Animals with Reduced Visual Disturbance**: The proposal of a navigation method resolves a practical problem of reducing the visual disturbance caused by wildlife observing drones. An optimization problem was formulated with the objective of minimizing the maximum visual disturbance (indicated by bearing changes) of multiple moving targets. This study provides one of the first investigations into reducing the negative impacts of wildlife observing drones by motion control. Theoretical analysis and computer simulations verified the effectiveness of the proposed method.

- **Efficient Optimal Backhaul-aware Deployment of Multiple Drone-Cells Based on Genetic Algorithm**: We formulate the optimal 2D backhaul-aware deployment problem of multiple drone-cells as a mixed-integer nonlinear programming problem (NP-hard). It aims at maximizing the total number of users covered by the drone-cells. Two approaches: an exhaustive search algorithm and a computationally efficient GA-based method are proposed to solve this problem. We also presented a restart-strategy that helps the proposed GA-based method to avoid local optima. Simulations show that the proposed GA-based method can significantly save computing time compare with the exhaustive search algorithm. We also verified that the restart-strategy is a simple but very effective technique that significantly increases the success rate for the GA to find the global optimum.

- **A Novel Method for Protecting Swimmers and Surfers from Shark Attacks using Communicating Autonomous Drones**: The proposal of a novel shark defence method based on communicating autonomous drones for protecting swimmers and surfers. It targets on using autonomous drones to protect swimmers and surfers from shark attacks, and eventually, drive the shark to leave the beach area. We present the design of the proposed drone shark shield system and its working mechanism. We also proposed a shark repelling strategy and an interception algorithm for drones to efficiently intercept sharks. Through various simulations, we demonstrated the performance of the proposed shark interception algorithm.

- **Autonomous Navigation of a Network of Barking Drones for Herding Farm Animals**: The proposal of a collision-free sliding mode based motion control algorithm, which targets on navigating a network of barking drones to efficiently collect a group of animals when they are too dispersed and drive them to a designated location. Simulations using Reynolds' rules based dynamic model of animal motion showed the proposed drone herding method can efficiently herd a thousand animals with several drones. Since the animal motion models used here are experimental verified, the results can serve as suitable guidelines in practice.

# References

[1] CompTIA, "Global Drone Delivery Market - Analysis and Forecast, 2023 to 2030," accessed 9 April. 2021. Online: https://www.comptia.org/content/research/drone-industry-trends-analysis.

[2] J. Jiménez López and M. Mulero-Pázmány, "Drones for conservation in protected areas: present and future," *Drones*, vol. 3, no. 1, p. 10, 2019.

[3] A. V. Savkin and H. Huang, "A method for optimized deployment of a network of surveillance aerial drones," *IEEE Systems Journal*, vol. 13, no. 4, pp. 4474–4477, 2019.

[4] H. Huang and A. V. Savkin, "An algorithm of reactive collision free 3D deployment of networked unmanned aerial vehicles for surveillance and monitoring," *IEEE Transactions on Industrial Informatics*, vol. 16, no. 1, pp. 132–140, 2020.

[5] L. Tang and G. Shao, "Drone remote sensing for forestry research and practices," *Journal of Forestry Research*, vol. 26, no. 4, pp. 791–797, 2015.

[6] H. Huang and A. V. Savkin, "Energy-efficient autonomous navigation of solar-powered UAVs for surveillance of mobile ground targets in urban environments," *Energies*, vol. 13, no. 21, p. 5563, 2020.

[7] A. V. Savkin and H. Huang, "Proactive deployment of aerial drones for coverage over very uneven terrains: A version of the 3D art gallery problem," *Sensors*, vol. 19, no. 6, p. 1438, 2019.

[8] H. Huang and A. V. Savkin, "Reactive 3D deployment of a flying robotic network for surveillance of mobile targets," *Computer Networks*, vol. 161, pp. 172–182, 2019.

[9] A. V. Savkin and H. Huang, "Navigation of a network of aerial drones for monitoring a frontier of a moving environmental disaster area," *IEEE Systems Journal*, vol. 14, no. 4, pp. 4746–4749, 2020.

[10] H. Huang, A. V. Savkin, and C. Huang, "Decentralised autonomous navigation of a UAV network for road traffic monitoring," *IEEE Transactions on Aerospace and Electronic Systems*, 2021.

[11] H. Huang and A. Savkin, "Navigating UAVs for optimal monitoring of groups of moving pedestrians or vehicles," *IEEE Transactions on Vehicular Technology*, vol. 70, no. 4, pp. 3891–3896, 2021.

[12] H. Huang, A. V. Savkin, and X. Li, "Reactive autonomous navigation of UAVs for dynamic sensing coverage of mobile ground targets," *Sensors*, vol. 20, no. 13, p. 3720, 2020.

[13] D. R. McArthur, A. B. Chowdhury, and D. J. Cappelleri, "Autonomous control of the interacting-boomcopter UAV for remote sensor mounting," in *2018 IEEE International Conference on Robotics and Automation (ICRA)*. IEEE, 2018, pp. 5219–5224.

[14] P. Štibinger, G. Broughton, F. Majer, Z. Rozsypálek, A. Wang, K. Jindal, A. Zhou, D. Thakur, G. Loianno, T. Krajník *et al.*, "Mobile manipulator for autonomous localization, grasping and precise placement of construction material in a semi-structured environment," *IEEE Robotics and Automation Letters*, vol. 6, no. 2, pp. 2595–2602, 2021.

[15] D. R. McArthur, Z. An, and D. J. Cappelleri, "Pose-estimate-based target tracking for human-guided remote sensor mounting with a UAV," in *2020 IEEE International Conference on Robotics and Automation (ICRA)*. IEEE, 2020, pp. 10 636–10 642.

[16] H. Huang and A. V. Savkin, "A method of optimized deployment of charging stations for drone delivery," *IEEE Transactions on Transportation Electrification*, vol. 6, no. 2, pp. 510–518, 2020.

[17] H. Huang, A. V. Savkin, and C. Huang, "A new parcel delivery system with drones and a public train," *Journal of Intelligent & Robotic Systems*, vol. 100, no. 3, pp. 1341–1354, 2020.

[18] H. Huang and A. V. Savkin, "A method for optimized deployment of unmanned aerial vehicles for maximum coverage and minimum interference in cellular networks," *IEEE Transactions on Industrial Informatics*, vol. 15, no. 5, pp. 2638–2647, 2018.

[19] X. Li, H. Huang, and A. V. Savkin, "Autonomous drone shark shield: A novel shark repelling system for protecting swimmers and surfers," in *2020 6th International Conference on Control, Automation and Robotics (ICCAR)*, 2020, pp. 455–458.

[20] A. A. Paranjape, S.-J. Chung, K. Kim, and D. H. Shim, "Robotic herding of a flock of birds using an unmanned aerial vehicle," *IEEE Transactions on Robotics*, vol. 34, no. 4, pp. 901–915, 2018.

[21] A. Rodríguez, J. J. Negro, M. Mulero, C. Rodríguez, J. Hernández-Pliego, and J. Bustamante, "The eye in the sky: combined use of unmanned aerial systems and GPS data loggers for ecological research and conservation of small birds," *PLoS One*, vol. 7, no. 12, p. e50336, 2012.

[22] J. R. Barr, M. C. Green, S. J. DeMaso, and T. B. Hardy, "Drone surveys do not increase colony-wide flight behaviour at waterbird nesting sites, but sensitivity varies among species," *Scientific Reports*, vol. 10, no. 1, pp. 1–10, 2020.

[23] M. A. Ditmer, J. B. Vincent, L. K. Werden, J. C. Tanner, T. G. Laske, P. A. Iaizzo, D. L. Garshelis, and J. R. Fieberg, "Bears show a physiological but

limited behavioral response to unmanned aerial vehicles," *Current Biology*, vol. 25, no. 17, pp. 2278–2283, 2015.

[24] E. Bennitt, H. L. Bartlam-Brooks, T. Y. Hubel, and A. M. Wilson, "Terrestrial mammalian wildlife responses to unmanned aerial systems approaches," *Scientific Reports*, vol. 9, no. 1, pp. 1–10, 2019.

[25] H. Huang and A. V. Savkin, "An algorithm of efficient proactive placement of autonomous drones for maximum coverage in cellular networks," *IEEE Wireless Communications Letters*, vol. 7, no. 6, pp. 994–997, 2018.

[26] E. Kalantari, M. Z. Shakir, H. Yanikomeroglu, and A. Yongacoglu, "Backhaul-aware robust 3d drone placement in 5G+ wireless networks," in *2017 IEEE international conference on communications workshops (ICC workshops)*. IEEE, 2017, pp. 109–114.

[27] C. T. Cicek, H. Gultekin, B. Tavli, and H. Yanikomeroglu, "Backhaul-aware optimization of UAV base station location and bandwidth allocation for profit maximization," *IEEE Access*, vol. 8, pp. 154 573–154 588, 2020.

[28] M. Levine, R. S. Collier, E. Ritter, M. Fouda, and V. Canabal, "Shark cognition and a human mediated driver of a spate of shark attacks," *Open Journal of Animal Sciences*, vol. 4, no. 05, p. 263, 2014.

[29] R. Vaughan *et al.*, "Robot sheepdog project achieves automatic flock control," in *Proc. Fifth International Conference on the Simulation of Adaptive Behaviour*, vol. 489. 493, 489, 1998, p. 493.

[30] N. Sumpter *et al.*, "Learning models of animal behaviour for a robotic sheepdog." in *MVA*, 1998, pp. 577–580.

[31] M. Evered *et al.*, "An investigation of predator response in robotic herding of sheep," *International Proceedings of Chemical, Biological and Environmental Engineering*, vol. 63, pp. 49–54, 2014.

[32] BBC, "Robot used to round up cows is a hit with farmers," accessed 28 May. 2020. Online: https://www.bbc.com/news/technology-24955943.

[33] Sciencealert, "Spot the robot sheep dog," accessed 28 May. 2020. Online: https://www.sciencealert.com/spot-the-robot-dog-is-now-herding-sheep-in-new-zealand.

[34] IEEE Spectrum, "Swagbot to herd cattle," accessed 28 May. 2020. Online: https://spectrum.ieee.org/automaton/robotics/industrial-robots/swagbot-to-herd-cattle-on-australian-ranches.

[35] D. R. McArthur, A. B. Chowdhury, and D. J. Cappelleri, "Autonomous control of the interacting-boomcopter UAV for remote sensor mounting," in *2018 IEEE International Conference on Robotics and Automation (ICRA)*. IEEE, 2018, pp. 5219–5224.

[36] D. Xilun, G. Pin, X. Kun, and Y. Yushu, "A review of aerial manipulation of small-scale rotorcraft unmanned robotic systems," *Chinese Journal of Aeronautics*, vol. 32, no. 1, pp. 200–214, 2019.

[37] F. Ruggiero, V. Lippiello, and A. Ollero, "Aerial manipulation: A literature review," *IEEE Robotics and Automation Letters*, vol. 3, no. 3, pp. 1957–1964, 2018.

[38] H. B. Khamseh, F. Janabi-Sharifi, and A. Abdessameud, "Aerial manipulation—a literature survey," *Robotics and Autonomous Systems*, vol. 107, pp. 221–235, 2018.

[39] H. Huang, A. V. Savkin, and C. Huang, "Drone routing in a time-dependent network: Toward low-cost and large-range parcel delivery," *IEEE Transactions on Industrial Informatics*, vol. 17, no. 2, pp. 1526–1534, 2020.

[40] ——, "Reliable path planning for drone delivery using a stochastic time-dependent public transportation network," *IEEE Transactions on Intelligent Transportation Systems*, 2020.

[41] ——, "Round trip routing for energy-efficient drone delivery based on a public transportation network," *IEEE Transactions on Transportation Electrification*, vol. 6, no. 3, pp. 1368–1376, 2020.

[42] ——, "Scheduling of a parcel delivery system consisting of an aerial drone interacting with public transportation vehicles," *Sensors*, vol. 20, no. 7, p. 2045, 2020.

[43] U. S. Panday, A. K. Pratihast, J. Aryal, and R. B. Kayastha, "A review on drone-based data solutions for cereal crops," *Drones*, vol. 4, no. 3, p. 41, 2020.

[44] D. Wu, R. Li, F. Zhang, and J. Liu, "A review on drone-based harmful algae blooms monitoring," *Environmental monitoring and assessment*, vol. 191, no. 4, pp. 1–11, 2019.

[45] B. Mishra, D. Garg, P. Narang, and V. Mishra, "Drone-surveillance for search and rescue in natural disaster," *Computer Communications*, vol. 156, pp. 1–10, 2020.

[46] F. Flammini, R. Naddei, C. Pragliola, and G. Smarra, "Towards automated drone surveillance in railways: State-of-the-art and future directions," in *International conference on advanced concepts for intelligent vision systems*. Springer, 2016, pp. 336–348.

[47] A. V. Savkin and H. Huang, "Asymptotically optimal deployment of drones for surveillance and monitoring," *Sensors*, vol. 19, no. 9, p. 2068, 2019.

[48] H. Huang, A. V. Savkin, and W. Ni, "Online UAV trajectory planning for covert video surveillance of mobile targets," *IEEE Transactions on Automation Science and Engineering*, 2021.

[49] A. V. Savkin and H. Huang, "Bioinspired bearing only motion camouflage UAV guidance for covert video surveillance of a moving target," *IEEE Systems Journal*, pp. 1–4, 2020.

[50] A. V. Savkin and H. Huang, "Navigation of a UAV network for optimal surveillance of a group of ground targets moving along a road," *IEEE Transactions on Intelligent Transportation Systems*, 2021.

[51] W. Chen, J. Liu, and H. Guo, "Achieving robust and efficient consensus for large-scale drone swarm," *IEEE Transactions on Vehicular Technology*, 2020.

[52] C. Huang, C. M. F. T. Lv, P. Hang, and Y. Xing, "Towards safe and personalized autonomous driving: Decision-making and motion control with dpf and cdt techniques," *IEEE/ASME Transactions on Mechatronics*, 2021.

[53] C. Huang, F. Naghdy, and H. Du, "Fault tolerant sliding mode predictive control for uncertain steer-by-wire system," *IEEE Transactions on cybernetics*, vol. 49, no. 1, pp. 261–272, 2017.

[54] C. Huang, H. Huang, P. Hang, H. Gao, J. Wu, Z. Huang, and C. Lv, "Personalized trajectory planning and control of lane-change maneuvers for autonomous driving," *IEEE Transactions on Vehicular Technology*, pp. 1–1, 2021.

[55] H. Huang and A. V. Savkin, "Path planning algorithms for a mobile robot collecting data in a wireless sensor network deployed in a region with obstacles," in *2016 35th Chinese Control Conference (CCC)*. IEEE, 2016, pp. 8464–8467.

[56] C. Huang, F. Naghdy, and H. Du, "Sliding mode predictive tracking control for uncertain steer-by-wire system," *Control Engineering Practice*, vol. 85, pp. 194–205, 2019.

[57] A. V. Savkin and C. Wang, "Seeking a path through the crowd: Robot navigation in unknown dynamic environments with moving obstacles based on an integrated environment representation," *Robotics and Autonomous Systems*, vol. 62, no. 10, pp. 1568–1580, 2014.

[58] H. Li and A. V. Savkin, "Wireless sensor network based navigation of micro flying robots in the industrial internet of things," *IEEE Transactions on industrial informatics*, vol. 14, no. 8, pp. 3524–3533, 2018.

[59] A. S. Matveev, H. Teimoori, and A. V. Savkin, "A method for guidance and control of an autonomous vehicle in problems of border patrolling and obstacle avoidance," *Automatica*, vol. 47, no. 3, pp. 515–524, 2011.

[60] A. S. Matveev, C. Wang, and A. V. Savkin, "Real-time navigation of mobile robots in problems of border patrolling and avoiding collisions with moving and deforming obstacles," *Robotics and Autonomous systems*, vol. 60, no. 6, pp. 769–788, 2012.

[61] C. Huang, F. Naghdy, and H. Du, "Delta operator-based fault estimation and fault-tolerant model predictive control for steer-by-wire systems," *IEEE Transactions on Control Systems Technology*, vol. 26, no. 5, pp. 1810–1817, 2017.

[62] B. Keller and T. Willke, "Snotbot: A whale of a deep-learning project," *IEEE Spectrum*, vol. 56, no. 12, pp. 41–53, 2019.

[63] V. Pirotta, A. Smith, M. Ostrowski, D. Russell, I. D. Jonsen, A. Grech, and R. Harcourt, "An economical custom-built drone for assessing whale health," *Frontiers in Marine Science*, vol. 4, p. 425, 2017.

[64] J. A. Barasona, M. Mulero-Pázmány, P. Acevedo, J. J. Negro, M. J. Torres, C. Gortázar, and J. Vicente, "Unmanned aircraft systems for studying spatial abundance of ungulates: relevance to spatial epidemiology," *PloS one*, vol. 9, no. 12, p. e115608, 2014.

[65] A. Michez, H. Piégay, J. Lisein, H. Claessens, and P. Lejeune, "Classification of riparian forest species and health condition using multi-temporal and hyperspatial imagery from unmanned aerial system," *Environmental monitoring and assessment*, vol. 188, no. 3, p. 146, 2016.

[66] J. Gonçalves, R. Henriques, P. Alves, R. Sousa-Silva, A. T. Monteiro, Â. Lomba, B. Marcos, and J. Honrado, "Evaluating an unmanned aerial vehicle-based approach for assessing habitat extent and condition in fine-scale

early successional mountain mosaics," *Applied Vegetation Science*, vol. 19, no. 1, pp. 132–146, 2016.

[67] J. Paneque-Gálvez, M. K. McCall, B. M. Napoletano, S. A. Wich, and L. P. Koh, "Small drones for community-based forest monitoring: An assessment of their feasibility and potential in tropical areas," *Forests*, vol. 5, no. 6, pp. 1481–1507, 2014.

[68] J. Zhang, J. Hu, J. Lian, Z. Fan, X. Ouyang, and W. Ye, "Seeing the forest from drones: Testing the potential of lightweight drones as a tool for long-term forest monitoring," *Biological Conservation*, vol. 198, pp. 60–69, 2016.

[69] B. C. Lubow and J. I. Ransom, "Practical bias correction in aerial surveys of large mammals: Validation of hybrid double-observer with sightability method against known abundance of feral horse (equus caballus) populations," *PLoS One*, vol. 11, no. 5, p. e0154902, 2016.

[70] Erica Cirino, "Drones help find massive penguin colonies hiding in plain sight," accessed 3 May. 2021. Online: https://deeply.thenewhumanitarian.org/oceans/articles/2018/03/05/drones-help-find-massive-penguin-colonies-hiding-in-plain-sight.

[71] L.-P. Chrétien, J. Théau, and P. Ménard, "Visible and thermal infrared remote sensing for the detection of white-tailed deer using an unmanned aerial system," *Wildlife Society Bulletin*, vol. 40, no. 1, pp. 181–191, 2016.

[72] S. Wich, D. Dellatore, M. Houghton, R. Ardi, and L. P. Koh, "A preliminary assessment of using conservation drones for sumatran orang-utan (pongo abelii) distribution and density," *Journal of Unmanned Vehicle Systems*, vol. 4, no. 1, pp. 45–52, 2015.

[73] K. L. Sweeney, V. T. Helker, W. L. Perryman, D. J. LeRoi, L. W. Fritz, T. S. Gelatt, and R. P. Angliss, "Flying beneath the clouds at the edge of the world: using a hexacopter to supplement abundance surveys of steller sea

lions (eumetopias jubatus) in alaska," *Journal of Unmanned Vehicle Systems*, vol. 4, no. 1, pp. 70–81, 2015.

[74] S. T. Sykora-Bodie, V. Bezy, D. W. Johnston, E. Newton, and K. J. Lohmann, "Quantifying nearshore sea turtle densities: applications of unmanned aerial systems for population assessments," *Scientific reports*, vol. 7, no. 1, pp. 1–7, 2017.

[75] J. J. Kiszka, J. Mourier, K. Gastrich, and M. R. Heithaus, "Using unmanned aerial vehicles (UAVs) to investigate shark and ray densities in a shallow coral lagoon," *Marine Ecology Progress Series*, vol. 560, pp. 237–242, 2016.

[76] C. T. Beranek, A. Roff, B. Denholm, L. G. Howell, and R. R. Witt, "Trialling a real-time drone detection and validation protocol for the koala (phascolarctos cinereus)," *Australian Mammalogy*, 2020.

[77] G. Schofield, K. A. Katselidis, M. K. Lilley, R. D. Reina, and G. C. Hays, "Detecting elusive aspects of wildlife ecology using drones: new insights on the mating dynamics and operational sex ratios of sea turtles," *Functional Ecology*, vol. 31, no. 12, pp. 2310–2319, 2017.

[78] L. G. Torres, S. L. Nieukirk, L. Lemos, and T. E. Chandler, "Drone up! quantifying whale behavior from a new perspective improves observational capacity," *Frontiers in Marine Science*, vol. 5, p. 319, 2018.

[79] I. Evans, T. H. Jones, K. Pang, M. N. Evans, S. Saimin, and B. Goossens, "Use of drone technology as a tool for behavioral research: a case study of crocodilian nesting," *Herpetological Conservation and Biology*, vol. 10, no. 1, pp. 90–98, 2015.

[80] P. A. Groves, B. Alcorn, M. M. Wiest, J. M. Maselko, and W. P. Connor, "Testing unmanned aircraft systems for salmon spawning surveys," *Facets*, vol. 1, no. 1, pp. 187–204, 2016.

[81] D. J. Stark, I. P. Vaughan, L. J. Evans, H. Kler, and B. Goossens, "Combining drones and satellite tracking as an effective tool for informing policy change in

riparian habitats: a proboscis monkey case study," *Remote Sensing in Ecology and Conservation*, vol. 4, no. 1, pp. 44–52, 2018.

[82] J. Junda, E. Greene, and D. M. Bird, "Proper flight technique for using a small rotary-winged drone aircraft to safely, quickly, and accurately survey raptor nests," *Journal of Unmanned Vehicle Systems*, vol. 3, no. 4, pp. 222–236, 2015.

[83] Y.-G. Han, S. H. Yoo, and O. Kwon, "Possibility of applying unmanned aerial vehicle (UAV) and mapping software for the monitoring of waterbirds and their habitats," *Journal of Ecology and Environment*, vol. 41, no. 1, pp. 1–7, 2017.

[84] M. Mulero-Pázmány, R. Stolper, L. Van Essen, J. J. Negro, and T. Sassen, "Remotely piloted aircraft systems as a rhinoceros anti-poaching tool in africa," *PloS one*, vol. 9, no. 1, p. e83873, 2014.

[85] M. J. Shaffer and J. A. Bishop, "Predicting and preventing elephant poaching incidents through statistical analysis, gis-based risk analysis, and aerial surveillance flight path modeling," *Tropical Conservation Science*, vol. 9, no. 1, pp. 525–548, 2016.

[86] E. Bondi, F. Fang, M. Hamilton, D. Kar, D. Dmello, J. Choi, R. Hannaford, A. Iyer, L. Joppa, M. Tambe *et al.*, "Spot poachers in action: Augmenting conservation drones with automatic detection in near real time," in *Proceedings of the AAAI Conference on Artificial Intelligence*, vol. 32, no. 1, 2018.

[87] C. Irigoin-Lovera, D. M. Luna, D. A. Acosta, and C. B. Zavalaga, "Response of colonial peruvian guano birds to flying UAVs: effects and feasibility for implementing new population monitoring methods," *PeerJ*, vol. 7, p. e8129, 2019.

[88] R. I. Bor-Yaliniz, A. El-Keyi, and H. Yanikomeroglu, "Efficient 3-d placement of an aerial base station in next generation cellular networks," in *2016 IEEE international conference on communications (ICC)*.  IEEE, 2016, pp. 1–5.

[89] A. V. Savkin and H. Huang, "Deployment of unmanned aerial vehicle base stations for optimal quality of coverage," *IEEE Wireless Communications Letters*, vol. 8, no. 1, pp. 321–324, 2018.

[90] X. Li, "Deployment of drone base stations for cellular communication without apriori user distribution information," in *2018 37th Chinese Control Conference (CCC)*. IEEE, 2018, pp. 7274–7281.

[91] X. Li and L. Xing, "Optimal deployment of drone base stations for cellular communication by network-based localization," in *2018 37th Chinese Control Conference (CCC)*. IEEE, 2018, pp. 7282–7287.

[92] H. Huang, A. V. Savkin, M. Ding, and M. A. Kaafar, "Optimized deployment of drone base station to improve user experience in cellular networks," *Journal of Network and Computer Applications*, vol. 144, pp. 49–58, 2019.

[93] A. V. Savkin and H. Huang, "Range-based reactive deployment of autonomous drones for optimal coverage in disaster areas," *IEEE Transactions on Systems, Man, and Cybernetics: Systems*, vol. 51, no. 7, pp. 4606–4610, 2021.

[94] M. Coldrey, U. Engström, K. W. Helmersson, M. Hashemi, L. Manholm, and P. Wallentin, "Wireless backhaul in future heterogeneous networks," *Ericsson Review*, vol. 91, pp. 1–11, 2014.

[95] A. Fouda, A. S. Ibrahim, I. Guvenc, and M. Ghosh, "UAV-based in-band integrated access and backhaul for 5G communications," in *2018 IEEE 88th Vehicular Technology Conference (VTC-Fall)*. IEEE, 2018, pp. 1–5.

[96] J. Lyu, Y. Zeng, R. Zhang, and T. J. Lim, "Placement optimization of uav-mounted mobile base stations," *IEEE Communications Letters*, vol. 21, no. 3, pp. 604–607, 2016.

[97] L. Wang, B. Hu, and S. Chen, "Energy efficient placement of a drone base station for minimum required transmit power," *IEEE Wireless Communications Letters*, vol. 9, no. 12, pp. 2010–2014, 2018.

[98] M. Alzenad, A. El-Keyi, F. Lagum, and H. Yanikomeroglu, "3-d placement of an unmanned aerial vehicle base station (uav-bs) for energy-efficient maximal coverage," *IEEE Wireless Communications Letters*, vol. 6, no. 4, pp. 434–437, 2017.

[99] T. Bai, C. Pan, J. Wang, Y. Deng, M. Elkashlan, A. Nallanathan, and L. Hanzo, "Dynamic aerial base station placement for minimum-delay communications," *IEEE Internet of Things Journal*, 2020.

[100] N. Hahn, A. Mwakatobe, J. Konuche, N. de Souza, J. Keyyu, M. Goss, A. Chang'a, S. Palminteri, E. Dinerstein, and D. Olson, "Unmanned aerial vehicles mitigate human–elephant conflict on the borders of tanzanian parks: a case study," *Oryx*, vol. 51, no. 3, pp. 513–516, 2017.

[101] S. Gade, A. A. Paranjape, and S.-J. Chung, "Herding a flock of birds approaching an airport using an unmanned aerial vehicle," in *AIAA Guidance, Navigation, and Control Conference*, 2015, p. 1540.

[102] P. Aliasghari, K. Dautenhahn, and C. L. Nehaniv, "Simulations on herding a flock of birds away from an aircraft using an unmanned aerial vehicle," in *Artificial Life Conference Proceedings*.   MIT Press, 2020, pp. 626–635.

[103] S. G. Penny, R. L. White, D. M. Scott, L. MacTavish, and A. P. Pernetta, "Using drones and sirens to elicit avoidance behaviour in white rhinoceros as an anti-poaching tactic," *Proceedings of the Royal Society B*, vol. 286, no. 1907, p. 20191135, 2019.

[104] J. Knight, "How to chase a monkey: Reforming the oiharai response to crop-feeding macaques in japan," *Society & Animals*, vol. 1, no. aop, pp. 1–19, 2020.

[105] M. V. Ogra, "Human–wildlife conflict and gender in protected area borderlands: a case study of costs, perceptions, and vulnerabilities from uttarakhand (uttaranchal), india," *Geoforum*, vol. 39, no. 3, pp. 1408–1422, 2008.

[106] N. W. Sitati, M. J. Walpole, R. J. Smith, and N. Leader-Williams, "Predicting spatial aspects of human–elephant conflict," *Journal of applied ecology*, vol. 40, no. 4, pp. 667–677, 2003.

[107] H. Huang, A. V. Savkin, M. Ding, and C. Huang, "Mobile robots in wireless sensor networks: A survey on tasks," *Computer Networks*, vol. 148, pp. 1–19, 2019.

[108] H. Huang and A. V. Savkin, "Towards the internet of flying robots: A survey," *Sensors*, vol. 18, no. 11, p. 4038, 2018.

[109] S. Hayat, E. Yanmaz, and R. Muzaffar, "Survey on unmanned aerial vehicle networks for civil applications: A communications viewpoint," *IEEE Communications Surveys & Tutorials*, vol. 18, no. 4, pp. 2624–2661, 2016.

[110] T. M. Cabreira, L. B. Brisolara, and P. R. Ferreira Jr, "Survey on coverage path planning with unmanned aerial vehicles," *Drones*, vol. 3, no. 1, p. 4, 2019.

[111] J. Huang, Y. Chen, Y. Huang, P. Lin, Y. Chen, Y. Lin, S. Yen, P. Huang, and L. Chen, "Rapid prototyping for wildlife and ecological monitoring," *IEEE Systems Journal*, vol. 4, no. 2, pp. 198–209, 2010.

[112] M. Mulero-Pázmány, S. Jenni-Eiermann, N. Strebel, T. Sattler, J. J. Negro, and Z. Tablado, "Unmanned aircraft systems as a new source of disturbance for wildlife: A systematic review," *PLOS One*, vol. 12, no. 6, p. e0178448, 2017.

[113] J. C. Hodgson and L. P. Koh, "Best practice for minimising unmanned aerial vehicle disturbance to wildlife in biological field research," *Current Biology*, vol. 26, no. 10, pp. R404–R405, 2016.

[114] D. Chabot and D. M. Bird, "Wildlife research and management methods in the 21st century: Where do unmanned aircraft fit in?" *Journal of Unmanned Vehicle Systems*, vol. 3, no. 4, pp. 137–155, 2015.

[115] E. Vas, A. Lescroël, O. Duriez, G. Boguszewski, and D. Grémillet, "Approaching birds with drones: first experiments and ethical guidelines," *Biology Letters*, vol. 11, no. 2, p. 20140754, 2015.

[116] A. Barnas, R. Newman, C. J. Felege, M. P. Corcoran, S. D. Hervey, T. J. Stechmann, R. F. Rockwell, and S. N. Ellis-Felege, "Evaluating behavioral responses of nesting lesser snow geese to unmanned aircraft surveys," *Ecology and evolution*, vol. 8, no. 2, pp. 1328–1338, 2018.

[117] N. M. Schroeder, A. Panebianco, R. Gonzalez Musso, and P. Carmanchahi, "An experimental approach to evaluate the potential of drones in terrestrial mammal research: a gregarious ungulate as a study model," *Royal Society Open Science*, vol. 7, no. 1, p. 191482, 2020.

[118] Dronethusiast, "What are the best silent drone choices and what applications are they good for," accessed 9 July. 2020. Online: https://www.dronethusiast.com/what-are-the-best-silent-drone-choices-and-what-applications-are-they-good-for/.

[119] M. V. Srinivasan and M. Davey, "Strategies for active camouflage of motion," *Proceedings of the Royal Society of London. Series B: Biological Sciences*, vol. 259, no. 1354, pp. 19–25, 1995.

[120] A. Mizutani, J. S. Chahl, and M. V. Srinivasan, "Motion camouflage in dragonflies," *Nature*, vol. 423, no. 6940, pp. 604–604, 2003.

[121] K. Ghose, T. K. Horiuchi, P. Krishnaprasad, and C. F. Moss, "Echolocating bats use a nearly time-optimal strategy to intercept prey," *PLOS Biology*, vol. 4, no. 5, p. e108, 2006.

[122] S. A. Kane and M. Zamani, "Falcons pursue prey using visual motion cues: new perspectives from animal-borne cameras," *Journal of Experimental Biology*, vol. 217, no. 2, pp. 225–234, 2014.

[123] A. J. Anderson and P. W. McOwan, "Humans deceived by predatory stealth strategy camouflaging motion," *Proceedings of the Royal Society of London. Series B: Biological Sciences*, vol. 270, no. suppl_1, pp. S18–S20, 2003.

[124] I. Ranó and R. Iglesias, "Application of systems identification to the implementation of motion camouflage in mobile robots," *Autonomous Robots*, vol. 40, no. 2, pp. 229–244, 2016.

[125] R. Strydom and M. V. Srinivasan, "UAS stealth: target pursuit at constant distance using a bio-inspired motion camouflage guidance law," *Bioinspiration & biomimetics*, vol. 12, no. 5, p. 055002, 2017.

[126] A. Prasad, B. Sharma, and J. Vanualailai, "Motion camouflage for point-mass robots using a lyapunov-based control scheme," in *2019 4th International Conference on Control and Robotics Engineering (ICCRE)*. IEEE, 2019, pp. 7–11.

[127] C. Wang, A. V. Savkin, and M. Garratt, "A strategy for safe 3D navigation of non-holonomic robots among moving obstacles," *Robotica*, vol. 36, no. 2, pp. 275–297, 2018.

[128] R. A. Nichols, R. T. Reichert, and W. J. Rugh, "Gain scheduling for H-infinity controllers: A flight control example," *IEEE Transactions on Control Systems Technology*, vol. 1, no. 2, pp. 69–79, 1993.

[129] X. Liu and Q. Zhang, "New approaches to H-infinity controller designs based on fuzzy observers for TS fuzzy systems via LMI," *Automatica*, vol. 39, no. 9, pp. 1571–1582, 2003.

[130] A. Savkin, I. R. Peterson, and V. Ugronovskii, *Robust control design using H-infinity methods*. Springer-Verlag, London, 2000.

[131] A. Bansal and V. Sharma, "Design and analysis of robust H-infinity controller," *Control theory and informatics*, vol. 3, no. 2, pp. 7–14, 2013.

[132] G. He, S. Dong, J. Qi, and Y. Wang, "Robust state estimator based on maximum normal measurement rate," *IEEE Transactions on Power Systems*, vol. 26, no. 4, pp. 2058–2065, 2011.

[133] I. R. Petersen and A. V. Savkin, *Robust Kalman filtering for signals and systems with large uncertainties.* Springer Science & Business Media, 1999.

[134] P. N. Pathirana, N. Bulusu, A. V. Savkin, and S. Jha, "Node localization using mobile robots in delay-tolerant sensor networks," *IEEE Transactions on Mobile Computing*, vol. 4, no. 3, pp. 285–296, 2005.

[135] T. Zhou, "On the convergence and stability of a robust state estimator," *IEEE Transactions on Automatic Control*, vol. 55, no. 3, pp. 708–714, 2010.

[136] A. V. Savkin and I. R. Petersen, "Robust state estimation and model validation for discrete-time uncertain systems with a deterministic description of noise and uncertainty," *Automatica*, vol. 34, no. 2, pp. 271–274, 1998.

[137] P. N. Pathirana, A. V. Savkin, and S. Jha, "Location estimation and trajectory prediction for cellular networks with mobile base stations," *IEEE Transactions on Vehicular Technology*, vol. 53, no. 6, pp. 1903–1913, 2004.

[138] Y. Chen, J. Ma, P. Zhang, F. Liu, and S. Mei, "Robust state estimator based on maximum exponential absolute value," *IEEE Transactions on Smart Grid*, vol. 8, no. 4, pp. 1537–1544, 2015.

[139] V. I. Utkin, *Sliding modes in control and optimization.* Springer Science & Business Media, 2013.

[140] R. K. Ganti and M. Haenggi, "Interference and outage in clustered wireless ad hoc networks," *IEEE Transactions on Information Theory*, vol. 55, no. 9, pp. 4067–4086, 2009.

[141] B. M. Chazelle and D.-T. Lee, "On a circle placement problem," *Computing*, vol. 36, no. 1-2, pp. 1–16, 1986.

[142] H. Holland John, "Adaptation in natural and artificial systems," *Ann Arbor: University of Michigan Press*, 1975.

[143] D. Adler, "Genetic algorithms and simulated annealing: A marriage proposal," in *IEEE International Conference on Neural Networks*. IEEE, 1993, pp. 1104–1109.

[144] M. Srinivas and L. M. Patnaik, "Adaptive probabilities of crossover and mutation in genetic algorithms," *IEEE Transactions on Systems, Man, and Cybernetics*, vol. 24, no. 4, pp. 656–667, 1994.

[145] B. A. Huberman, R. M. Lukose, and T. Hogg, "An economics approach to hard computational problems," *Science*, vol. 275, no. 5296, pp. 51–54, 1997.

[146] X. Kai, C. Wei, and L. Liu, "Robust extended Kalman filtering for nonlinear systems with stochastic uncertainties," *IEEE Transactions on Systems, Man, and Cybernetics-Part A: Systems and Humans*, vol. 40, no. 2, pp. 399–405, 2009.

[147] A. V. Savkin and I. R. Petersen, "Model validation for robust control of uncertain systems with an integral quadratic constraint," *Automatica*, vol. 32, no. 4, pp. 603–606, 1996.

[148] L. El Ghaoui and G. Calafiore, "Robust filtering for discrete-time systems with bounded noise and parametric uncertainty," *IEEE Transactions on Automatic Control*, vol. 46, no. 7, pp. 1084–1089, 2001.

[149] V. Malyavej and A. V. Savkin, "The problem of optimal robust Kalman state estimation via limited capacity digital communication channels," *Systems & Control Letters*, vol. 54, no. 3, pp. 283–292, 2005.

[150] B. K. Chapman and D. McPhee, "Global shark attack hotspots: Identifying underlying factors behind increased unprovoked shark bite incidence," *Ocean & Coastal Management*, vol. 133, pp. 72–84, 2016.

[151] A. S. Afonso, Y. V. Niella, and F. H. Hazin, "Inferring trends and linkages between shark abundance and shark bites on humans for shark-hazard mitigation," *Marine and Freshwater Research*, vol. 68, no. 7, pp. 1354–1365, 2017.

[152] J. A. Bolin, D. S. Schoeman, C. Pizà-Roca, and K. L. Scales, "A current affair: entanglement of humpback whales in coastal shark-control nets," *Remote Sensing in Ecology and Conservation*, 2019.

[153] D. McPhee, "Unprovoked shark bites: are they becoming more prevalent?" *Coastal Management*, vol. 42, no. 5, pp. 478–492, 2014.

[154] Sharkangels, "Remove the nets," accessed 27 Oct. 2019. Online: https://sharkangels.org/index.php/media/news/157-shark-nets.

[155] The Sydney Morning Herald, "Bondi shark attack second strike in two days," accessed 27 Oct. 2019. Online: https://www.smh.com.au/national/bondi-shark-attack-second-strike-in-two-days-20090212-85zf.html.

[156] Youtube, "Drone captures shark lurking close to family," accessed 27 Oct. 2019. Online: https://www.youtube.com/watch?v=UTz5qFteP2U.

[157] CNN, "Drone catches shark stalking a surfer," accessed 27 Oct. 2019. Online: https://edition.cnn.com/videos/world/2019/09/18/drone-warns-surfer-shark-mxp-vpx.hln.

[158] M. Hu, W. Liu, K. Peng, X. Ma, W. Cheng, J. Liu, and B. Li, "Joint routing and scheduling for vehicle-assisted multidrone surveillance," *IEEE Internet of Things Journal*, vol. 6, no. 2, pp. 1781–1790, 2018.

[159] M. Wan, G. Gu, W. Qian, K. Ren, X. Maldague, and Q. Chen, "Unmanned aerial vehicle video-based target tracking algorithm using sparse representation," *IEEE Internet of Things Journal*, vol. 6, no. 6, pp. 9689–9706, 2019.

[160] N. Sharma, P. Scully-Power, and M. Blumenstein, "Shark detection from aerial imagery using region-based CNN, a study," in *Australasian Joint Conference on Artificial Intelligence*. Springer, 2018, pp. 224–236.

[161] Reuters, "Shark-detecting drones to patrol australian beaches," accessed 27 Oct. 2019. Online: https://www.reuters.com/article/us-australia-sharkdrone/shark-detecting-drones-to-patrol-australian-beaches-idUSKCN1B51KB.

[162] S. Collin, "Electroreception in vertebrates and invertebrates," in *Reference module in life sciences*.   Elsevier, 2017, pp. 611–620.

[163] R. Kempster, I. McCarthy, and S. Collin, "Phylogenetic and ecological factors influencing the number and distribution of electroreceptors in elasmobranchs," *Journal of Fish Biology*, vol. 80, no. 5, pp. 2055–2088, 2012.

[164] G. E. Charter, S. H. Ripley, and N. G. Starkey, "Control of sharks," Oct. 22 1996, US Patent 5,566,643.

[165] C. Huveneers, S. Whitmarsh, M. Thiele, L. Meyer, A. Fox, and C. J. Bradshaw, "Effectiveness of five personal shark-bite deterrents for surfers," *PeerJ*, vol. 6, p. e5554, 2018.

[166] C. Smit and V. Peddemors, "Estimating the probability of a shark attack when using an electric repellent: applications," *South African Statistical Journal*, vol. 37, no. 1, pp. 59–78, 2003.

[167] R. M. Kempster, C. A. Egeberg, N. S. Hart, L. Ryan, L. Chapuis, C. C. Kerr, C. Schmidt, C. Huveneers, E. Gennari, K. E. Yopak *et al.*, "How close is too close? The effect of a non-lethal electric shark deterrent on white shark behaviour," *PLoS One*, vol. 11, no. 7, p. e0157717, 2016.

[168] J. Zhu, R. He, C. Yu, and B. Lin, "An improved nearest point based location routing protocol for maritime wireless mesh networks," in *2019 IEEE 9th International Conference on Electronics Information and Emergency Communication (ICEIEC)*.   IEEE, 2019, pp. 1–4.

[169] W. Z. Khan, M. Y. Aalsalem, N. Saad, Y. Xaing, and T. H. Luan, "Detecting replicated nodes in wireless sensor networks using random walks and network

division," in *2014 IEEE Wireless Communications and Networking Conference (WCNC)*.   IEEE, 2014, pp. 2623–2628.

[170] T. M. Cheng, A. V. Savkin, and F. Javed, "Decentralized control of a group of mobile robots for deployment in sweep coverage," *Robotics and Autonomous Systems*, vol. 59, no. 7-8, pp. 497–507, 2011.

[171] A. S. Matveev, A. V. Savkin, M. Hoy, and C. Wang, *Safe Robot Navigation among Moving and Steady Obstacles*.   Elsevier, 2015.

[172] M. Hoy, A. S. Matveev, and A. V. Savkin, "Algorithms for collision-free navigation of mobile robots in complex cluttered environments: a survey," *Robotica*, vol. 33, no. 3, pp. 463–497, 2015.

[173] A. S. Matveev and A. V. Savkin, *Estimation and control over communication networks*.   Springer Science & Business Media, 2009.

[174] W. Zhang, M. S. Branicky, and S. M. Phillips, "Stability of networked control systems," *IEEE Control Systems Magazine*, vol. 21, no. 1, pp. 84–99, 2001.

[175] A. V. Savkin, "Analysis and synthesis of networked control systems: Topological entropy, observability, robustness and optimal control," *Automatica*, vol. 42, no. 1, pp. 51–62, 2006.

[176] Y. Tipsuwan and M.-Y. Chow, "Control methodologies in networked control systems," *Control Engineering Practice*, vol. 11, no. 10, pp. 1099–1111, 2003.

[177] A. S. Matveev and A. V. Savkin, "An analogue of Shannon information theory for detection and stabilization via noisy discrete communication channels," *SIAM Journal on Control and Optimization*, vol. 46, no. 4, pp. 1323–1367, 2007.

[178] G. C. Walsh, H. Ye, and L. G. Bushnell, "Stability analysis of networked control systems," *IEEE Transactions on Control Systems Technology*, vol. 10, no. 3, pp. 438–446, 2002.

[179] A. V. Savkin and I. R. Petersen, "Set-valued state estimation via a limited capacity communication channel," *IEEE Transactions on Automatic Control*, vol. 48, no. 4, pp. 676–680, 2003.

[180] F.-Y. Wang and D. Liu, "Networked control systems," *Theory and Applications*, 2008.

[181] A. S. Matveev and A. V. Savkin, "The problem of state estimation via asynchronous communication channels with irregular transmission times," *IEEE Transactions on Automatic Control*, vol. 48, no. 4, pp. 670–676, 2003.

[182] A. V. Savkin and T. M. Cheng, "Detectability and output feedback stabilizability of nonlinear networked control systems," *IEEE Transactions on Automatic Control*, vol. 52, no. 4, pp. 730–735, 2007.

[183] X.-M. Zhang, Q.-L. Han, X. Ge, D. Ding, L. Ding, D. Yue, and C. Peng, "Networked control systems: a survey of trends and techniques," *IEEE/CAA Journal of Automatica Sinica*, vol. 7, no. 1, pp. 1–17, 2019.

[184] A. S. Matveev and A. V. Savkin, "The problem of LQG optimal control via a limited capacity communication channel," *Systems & Control Letters*, vol. 53, no. 1, pp. 51–64, 2004.

[185] S. Ramakrishna and D. Hull, "Tensile behaviour of knitted carbon-fibre-fabric/epoxy laminates—part II: Prediction of tensile properties," *Composites Science and Technology*, vol. 50, no. 2, pp. 249–258, 1994.

[186] AIROBOTICS, "Automated industrial drones," accessed 22 Feb. 2020. Online: https://www.airoboticsdrones.com/.

[187] Y. Sun, D. Xu, D. W. K. Ng, L. Dai, and R. Schober, "Optimal 3D-trajectory design and resource allocation for solar-powered UAV communication systems," *IEEE Transactions on Communications*, vol. 67, no. 6, pp. 4281–4298, 2019.

[188] A. Zakhar'eva, A. S. Matveev, M. C. Hoy, and A. V. Savkin, "Distributed control of multiple non-holonomic robots with sector vision and range-only measurements for target capturing with collision avoidance," *Robotica*, vol. 33, no. 2, pp. 385–412, 2015.

[189] I. R. Manchester and A. V. Savkin, "Circular-navigation-guidance law for precision missile/target engagements," *Journal of Guidance, Control, and Dynamics*, vol. 29, no. 2, pp. 314–320, 2006.

[190] A. V. Savkin and H. Teimoori, "Bearings-only guidance of a unicycle-like vehicle following a moving target with a smaller minimum turning radius," *IEEE Transactions on Automatic Control*, vol. 55, no. 10, pp. 2390–2395, 2010.

[191] S. A. Aleem, C. Nowzari, and G. J. Pappas, "Self-triggered pursuit of a single evader," in *2015 54th IEEE Conference on Decision and Control*. IEEE, 2015, pp. 1433–1440.

[192] R. Isaacs, *Differential games: a mathematical theory with applications to warfare and pursuit, control and optimization*. Courier Corporation, 1999.

[193] O. Elijah, T. A. Rahman, I. Orikumhi, C. Y. Leow, and M. N. Hindia, "An overview of internet of things (IoT) and data analytics in agriculture: Benefits and challenges," *IEEE Internet of Things Journal*, vol. 5, no. 5, pp. 3758–3773, 2018.

[194] S. Birrell, J. Hughes, J. Y. Cai, and F. Iida, "A field-tested robotic harvesting system for iceberg lettuce," *Journal of Field Robotics*, 2019.

[195] N. Ahmed, D. De, and I. Hussain, "Internet of things (IoT) for smart precision agriculture and farming in rural areas," *IEEE Internet of Things Journal*, vol. 5, no. 6, pp. 4890–4899, 2018.

[196] D. Marini *et al.*, "Controlling within-field sheep movement using virtual fencing," *Animals*, vol. 8, no. 3, p. 31, 2018.

[197] Y. Yao, Y. Sun, C. Phillips, and Y. Cao, "Movement-aware relay selection for delay-tolerant information dissemination in wildlife tracking and monitoring applications," *IEEE Internet of Things Journal*, vol. 5, no. 4, pp. 3079–3090, 2018.

[198] B. Achour, M. Belkadi, R. Aoudjit, and M. Laghrouche, "Unsupervised automated monitoring of dairy cows' behavior based on inertial measurement unit attached to their back," *Computers and Electronics in Agriculture*, vol. 167, p. 105068, 2019.

[199] D. Strömbom *et al.*, "Solving the shepherding problem: heuristics for herding autonomous, interacting agents," *Journal of the Royal Society Interface*, vol. 11, no. 100, p. 20140719, 2014.

[200] H. Hoshi, I. Iimura, S. Nakayama, Y. Moriyama, and K. Ishibashi, "Computer simulation based robustness comparison regarding agents' moving-speeds in two-and three-dimensional herding algorithms," in *2018 Joint 10th International Conference on Soft Computing and Intelligent Systems (SCIS) and 19th International Symposium on Advanced Intelligent Systems (ISIS)*. IEEE, 2018, pp. 1307–1314.

[201] A. Pierson and M. Schwager, "Controlling noncooperative herds with robotic herders," *IEEE Transactions on Robotics*, vol. 34, no. 2, pp. 517–525, 2017.

[202] ——, "Bio-inspired non-cooperative multi-robot herding," in *2015 IEEE International Conference on Robotics and Automation (ICRA)*. IEEE, 2015, pp. 1843–1849.

[203] H. Singh *et al.*, "Modulation of force vectors for effective shepherding of a swarm: A bi-objective approach," in *2019 IEEE Congress on Evolutionary Computation (CEC)*. IEEE, 2019, pp. 2941–2948.

[204] J.-A. Vayssade, R. Arquet, and M. Bonneau, "Automatic activity tracking of goats using drone camera," *Computers and Electronics in Agriculture*, vol. 162, pp. 767–772, 2019.

[205] X. Li, H. Huang, and A. V. Savkin, "A novel method for protecting swimmers and surfers from shark attacks using communicating autonomous drones," *IEEE Internet of Things Journal*, vol. 7, no. 10, pp. 9884–9894, 2020.

[206] RaisingSheep, "Sheep herding dogs," accessed 28 May. 2020. Online: https://http://www.raisingsheep.net/sheep-herding-dogs.html/.

[207] The Washington Post, "New zealand farmers have a new tool for herding sheep: drones that bark like dogs," accessed 28 May. 2020. Online: https://www.washingtonpost.com/technology/2019/03/07/new-zealand-farmers-have-new-tool-herding-sheep-drones-that-bark-like-dogs/.

[208] M. H. Memon, W. Kumar, A. Memon, B. S. Chowdhry, M. Aamir, and P. Kumar, "Internet of things (IoT) enabled smart animal farm," in *2016 3rd International Conference on Computing for Sustainable Global Development (IN-DIACom)*, 2016, pp. 2067–2072.

[209] S. Jo, D. Park, H. Park, and S. Kim, "Smart livestock farms using digital twin: Feasibility study," in *2018 International Conference on Information and Communication Technology Convergence (ICTC)*, 2018, pp. 1461–1463.

[210] K. Fujioka and S. Hayashi, "Effective shepherding behaviours using multi-agent systems," in *2016 IEEE Region 10 Conference (TENCON)*. IEEE, 2016, pp. 3179–3182.

[211] C. W. Reynolds, "Flocks, herds and schools: A distributed behavioral model," in *Proceedings of the 14th annual Conference on Computer Graphics and Interactive Techniques*, 1987, pp. 25–34.

[212] V. Utkin, J. Guldner, and M. Shijun, *Sliding mode control in electro-mechanical systems*. CRC press, 1999, vol. 34.

[213] V. I. Utkin, "Sliding modes and their applications in variable structure systems," *Mir, Moscow*, 1978.

[214] A. V. Savkin and R. J. Evans, *Hybrid dynamical systems: controller and sensor switching problems.* Springer Science & Business Media, 2002.

[215] E. Skafidas, R. J. Evans, A. V. Savkin, and I. R. Petersen, "Stability results for switched controller systems," *Automatica*, vol. 35, no. 4, pp. 553–564, 1999.